\documentclass[preprint, 10pt]{elsarticle}
\usepackage{graphicx}

\makeatletter
\newcommand*\bigcdot{\mathpalette\bigcdot@{.5}}
\newcommand*\bigcdot@[2]{\mathbin{\vcenter{\hbox{\scalebox{#2}{$\m@th#1\bullet$}}}}}
\makeatother
\usepackage[margin=2.5cm]{geometry}
\usepackage{setspace}
\usepackage[percent]{overpic}
\usepackage{lmodern}
\usepackage{amsmath,amssymb}
\usepackage{amsfonts,amsthm}
\usepackage{lineno}
\usepackage{graphicx}
\usepackage{multirow}
\usepackage{bm, bbm}
\usepackage{booktabs}
\usepackage{fancyvrb}
\usepackage{algorithmicx,algorithm}
\usepackage{algpseudocode}
\usepackage{physics}
\usepackage{xcolor}
\usepackage{pxfonts}
\usepackage[footnotesize,bf]{caption}
\usepackage{subcaption}
\usepackage{mathtools}
\usepackage{hhline}
\usepackage[T1]{fontenc}
\usepackage[colorlinks,linkcolor=red,anchorcolor=blue,citecolor=green]{hyperref}
\usepackage{placeins}

\newcommand{\mb}{\mathbb}
\newcommand{\mbe}{\mathbb{E}}

\newcommand{\mc}{\mathcal}

\newtheorem{remark}{Remark}[section]

\DeclareMathOperator{\diag}{diag}

\graphicspath{{figures/}}
\journal{}

\begin{document}
\begin{frontmatter}
	\title{LVM-GP: Uncertainty-Aware PDE Solver via coupling latent variable model and Gaussian process}
	\author[UIC]{Xiaodong Feng}
	\author[SHNU]{Ling Guo}
	\author[lu]{Xiaoliang Wan}
	\author[SJTU]{Hao Wu\corref{cor1}} 
	\cortext[cor1]{Corresponding author.} 
	\ead{hwu81@sjtu.edu.cn}  
	
	\author[LSECL]{Tao Zhou}
	\author[SHNU]{Wenwen Zhou}
	
	\vspace{0.2cm}
	\address[UIC]{Faculty of Science and Technology, Beijing Normal-Hong Kong Baptist University, Zhuhai 519087, China.}

	\vspace{0.2cm}
	\address[SHNU]{Department of Mathematics, Shanghai Normal University, Shanghai, China.}

	\vspace{0.2cm}
	\address[lu]{
		Department of Mathematics and Center for Computation and Technology, Louisiana State University, Baton Rouge 70803, USA}

	\vspace{0.2cm}
	\address[SJTU]{School of Mathematical Sciences, Institute of Natural Sciences, and MOE-LSC, Shanghai Jiaotong University, Shanghai, China.}

	\vspace{0.2cm}
	\address[LSECL]{Institute of Computational Mathematics and Scientific/Engineering
		Computing, Academy of Mathematics and Systems Science, Chinese Academy
		of Sciences, Beijing, China}

		\begin{abstract}
		We propose a novel probabilistic framework, termed LVM-GP, for uncertainty quantification in solving forward and inverse partial differential equations (PDEs) with noisy data. The core idea is to construct a stochastic mapping from the input to a high-dimensional latent representation, 
		enabling uncertainty-aware prediction of the solution. Specifically, the architecture consists of a confidence-aware encoder and a probabilistic decoder. The encoder implements a high-dimensional latent variable model based on a Gaussian process (LVM-GP), where the latent representation is constructed by interpolating between a learnable deterministic feature and a Gaussian process prior,  with the interpolation strength adaptively controlled by a confidence function learned from data. The decoder defines a conditional Gaussian distribution over the solution field, where the mean is predicted by a neural operator applied to the latent representation, allowing the model to learn flexible function-to-function mapping. Moreover, physical laws are enforced  as soft constraints in the loss function to ensure consistency with the underlying PDE structure.
		Compared to existing approaches such as Bayesian physics-informed neural networks (B-PINNs) and deep ensembles, the proposed framework can efficiently capture functional dependencies via merging a latent Gaussian process and neural operator, resulting in competitive predictive accuracy and robust uncertainty quantification.
		Numerical experiments demonstrate the effectiveness and reliability of the method.
	\end{abstract}
\end{frontmatter}
\section{Introduction}\label{sec:introduction}
Partial differential equations (PDEs) are extensively employed to model complex systems in various physical contexts. However, achieving accurate modeling and forecasting of the responses of physical systems necessitates the numerical solution of the corresponding PDEs, which has remained an challenging scientific problem for decades. In recent years, scientific machine learning (SciML), especially physics-informed machine learning \cite{karniadakis2021physics,willard2022integrating}, has achieved remarkable success in various applications. Many machine learning techniques have been used to solve PDEs, see \cite{weinan2020machine} and references therein. Among others we mention physics-informed neural networks (PINNs) \cite{raissi2019physics}, deep Ritz methods \cite{weinan2018deep} and deep Galerkin methods \cite{sirignano2018dgm}. These approaches have found a wide range of successful applications, including modeling fluid mechanics \cite{wang2017physics,raissi2020hidden} and solving high-dimensional PDEs \cite{han2018solving,huang2022augmented,zang2020weak}, to name just a few. Meanwhile, operator learning has recently attracted growing attention for modeling mappings between function spaces arising in PDE systems. Representative methods include DeepONet \cite{lu2021learning}, based on the universal approximation theorem for nonlinear operators \cite{chen1995universal}, and the Fourier Neural Operator (FNO) \cite{li2020fourier}. Other developments include Galerkin transformers \cite{cao2021choose}, kernel-based attention operators \cite{kissas2022learning}, physics-informed variants \cite{wang2021learning,wang2022improved,geneva2020modeling} and so on.

However, due to incomplete or inaccurate knowledge and inherent variability in physical systems, probabilistic approaches are essential for reliable uncertainty quantification (UQ) \cite{psaros2023uncertainty}. 
Among existing methods, two representative methods are Gaussian process regression (GPR) \cite{raissi2017machine} and Bayesian neural networks
(\cite{kendall2017uncertainties}). GPR is one of the most popular data-driven methods, however, vanilla GPR has difficulties
in handling the nonlinearity when applied to solve PDEs, although there are some recent works in \cite{chen2021solving, chen2023sparse} to overcome this issues. The Bayesian neural network, incorporating prior knowledge of physics, has demonstrated strong performance across a range of tasks \cite{yang2021b,linka2022bayesian,lin2023b}. However, the accurate computation of posterior probability distributions comes at a significant cost \cite{lotfi2022bayesian}, such as Bayesian PINN with Hamiltonian Monte Carlo (B-PINN-HMC) \cite{yang2021b}. In addition to these Bayesian approaches, non-Bayesian techniques such as deep ensembles \cite{lakshminarayanan2017simple,fort2019deep}, random priors \cite{malinin2018predictive, yang2022scalable}, evidence theory \cite{amini2020deep} and hybrid methods \cite{akhare2023diffhybrid,du2024conditional} have been explored in the context of scientific machine learning. 
More recently, Guo et al.~\cite{guo2023ib} proposed the information bottleneck based UQ (IB-UQ) framework for uncertainty quantification in function regression and operator learning tasks. Bergna et al. \cite{bergna2025post} introduce the Gaussian Process Activation Function, which propagates neural-level uncertainty through a pre-trained network. Nair et al. \cite{nair2025pinns} propose Epistemic PINNs by incorporating additive noise terms to explicitly model epistemic uncertainty. Psaros et al. provided an in-depth review and proposed novel methods of UQ for scientific machine learning in \cite{psaros2023uncertainty}, alongside the open-source library NeuralUQ \cite{zou2022neuraluq}. 

\paragraph{Our goals and contribution} Despite these advances, many existing methods either lack a structured mechanism for representing latent functional uncertainty or remain computationally burdensome in PDE applications. These limitations motivate the exploration of a new probabilistic
uncertainty quantification framework within the context of physics-informed machine learning. To this end, we propose a novel method termed LVM-GP, which couples latent variable modeling with Gaussian process priors to enable expressive, scalable, and physically consistent UQ in noisy forward and inverse PDE problems. Our contributions are summarized as follows:

\begin{itemize}
\item We develop LVM-GP, a physics-informed probabilistic framework for solving nonlinear partial differential equations (PDEs) under noisy observations.
\item We propose a novel encoder design in which the latent representation is formed by interpolating between a learnable deterministic feature and a Gaussian process prior, guided by a confidence function inferred from training data. This formulation enables input-dependent representation of epistemic uncertainty. In contrast to the commonly used standard normal prior, 
the Gaussian process prior captures spatial correlations across input locations, leading to more informed and structured uncertainty estimates.
\item We propose a decoder that models the conditional distribution of the solution as a Gaussian, where the mean is predicted by a neural operator applied to the latent representation. This design enables flexible function-to-function mappings while naturally capturing uncertainty in the solution.

\item We conduct comparisons with existing methods including B-PINN-HMC and deep ensembles. Our approach achieves comparable performance to B-PINN-HMC in both predictive accuracy and uncertainty quantification. In contrast, deep ensembles may produce spurious oscillations in the solution when the amount of observed data is limited.
\end{itemize}

The rest of this paper is organized as follows. In Section 2, we provide a concise overview of PINNs. Then we review recent advances in uncertainty quantification for physics informed machine learning. In Section 3, we present our LVM-GP model for solving forward and inverse PDEs. Section 4 investigates the performance of the developed techniques with a variety of tests for various PDE systems. We conclude in Section 5 with a summary of this work and extensions to address limitations that have been identified.
\section{Background} 
In this section, we give a brief review on physics informed neural networks (PINNs) \cite{raissi2019physics}, followed by recent advances in uncertainty quantification for physics informed machine learning.
\subsection{Physics-informed neural network}
Consider the following PDE
\begin{equation}
	\begin{aligned}
		 & \mathcal{N}_{\bm{x}} (u;\lambda)=f, & \bm{x}\in \Omega,          \\
		 & \mc{B}_{\bm{x}}(u;\lambda)=b,  & \bm{x}\in \partial \Omega,
	\end{aligned}
	\label{pinn_pde}
\end{equation}
where $\mc{N}_{\bm{x}}$ is the general form of a differential operator that could be nonlinear, $\Omega$ is the $d$-dimensional physical domain in $\mathbb{R}^d$, $u(\bm{x})$ is the solution of PDE and $\lambda$ is the vector parameter or function that appears in the PDE. The boundary conditions are imposed through a generalized boundary operator $\mc{B}_{\bm{x}}$ on $\partial \Omega$. In the forward problem, the parameter $\lambda$ is known and the solution $u(\bm{x})$ is to be computed for any $\bm{x} \in \Omega$. In the inverse problem, the parameter vector or function $\lambda$ is unknown, and both $\lambda$ and $u$ need to be inferred from the available data.

A deep neural network (DNN) is a sequence alternative composition of linear functions and nonlinear activation functions.
We aim to approximate the solution $u(\bm{x})$ and the unknown parameter $\lambda$ using neural networks. Depending on the problem setting, $\lambda$ may be either a constant or a spatially varying function. Accordingly, we denote it by $\lambda(\theta)$ if it is a learnable constant, or by $\lambda(\bm{x};\theta)$ if it is approximated by a neural network as a function of $\bm{x}$.
The PINNs approach uses the outputs of DNN, $u_{\mathrm{NN}}(\bm{x};\theta)$ and $\lambda_{\mathrm{NN}}(\bm{x};\theta)$ (or $\lambda_{\mathrm{NN}}(\theta)$), to approximate the solution of equation $u(\bm{x})$ and the unknown parameter $\lambda$, then calculate the differential operator via automatic differentiation. Here $\theta$ is a collection of all learnable parameters in the DNN. Specifically, we assume that $\lambda$ is an unknown function of $\bm{x}$, define the PDE residual as
\begin{equation}
	r(\bm{x};\theta) = \mathcal{N}_{\bm{x}} \big(u_{\mathrm{NN}}(\bm{x};\theta); \lambda_{\mathrm{NN}}(\bm{x};\theta)\big) - f(\bm{x}),
\end{equation}
assume that we have a limited number of extra $u$-sensors $\{\bm{x}_{d}^{i}, u(\bm{x}_d^i)\}_{i=1}^{N_d}$, then $\bm{\theta}$ can be learned by minimizing the following composite loss function
\begin{equation}\label{loss}
	\mathcal{L}(\theta) = w_r\cdot \mathcal{L}_r(\theta) +w_b\cdot  \mathcal{L}_b(\theta) + w_d \cdot \mathcal{L}_d(\theta),
\end{equation}
where
\begin{equation}
\begin{aligned}
	\mathcal{L}_r(\theta) = \frac{1}{2}\sum_{i=1}^{N_r}\big\vert r(\bm{x}_r^i;\theta)\big\vert ^2,\;\mathcal{L}_b(\theta) = \frac{1}{2}\sum_{i=1}^{N_b}\big\vert \mc{B}_{\bm{x}}(u_{\mathrm{NN}}(\bm{x}_b^i;\theta))-b(\bm{x}_b^i)\big\vert ^2, \; \mc{L}_d(\theta) = \frac{1}{2}\sum_{i=1}^{N_d} \big\vert u(\bm{x}_d^i) - u_{\mathrm{NN}}(\bm{x}_d^i;\theta)\big\vert^2.
\end{aligned}
\end{equation}
Here $\{w_r,w_b, w_d\}$ are prescribed weights and $\{\bm{x}_r^i\}_{i=1}^{N_r}$, $\{\bm{x}_b^i\}_{i=1}^{N_b}$ denote the interior and boundary training data, respectively.

Although PINNs have already led to a series of remarkable successes across of a range of problems in computational science and engineering, they are not equipped with built-in uncertainty quantification. Next, we will introduce some recent advances in uncertainty quantification for PDE problems.

\subsection{Uncertainty Quantification for physics informed machine learning}

While PINNs have achieved significant success in solving PDE-constrained problems, they inherently lack mechanisms for quantifying predictive uncertainty. To address this limitation, a variety of UQ techniques have been developed within the framework of physics-informed machine learning. Among them, Gaussian processes (GPs) provide a flexible non-parametric Bayesian framework to model uncertainty directly over function spaces, while latent variable models (LVMs) leverage latent variables to capture hidden structures and represent complex uncertainties in PDE solutions.
In what follows, we first review the GP framework in physics-informed learning, then discuss latent variable models.
\subsubsection{Gaussian processes}
GPs offer a non-parametric and fully Bayesian framework to modeling distributions over functions. We begin by placing a standard GP prior on unknown PDE solution $u$:\begin{equation*}
	u(\bm{x})\sim \mc{GP}\left(m(\bm{x}), k(\bm{x},\bm{x}')\right),
\end{equation*}
where $m(\bm{x})$ is the prior mean function (often taken as zero), and $k(\bm{x},\bm{x}')$ is a positive-definite kernel encoding smoothness and spatial correlation.
To incorporate physical knowledge in the form of differential equations or boundary conditions - into the GP framework, two main strategies are commonly adopted: physics-informed priors \cite{raissi2017machine,raissi2018numerical,chen2021solving} and physics-informed likelihoods \cite{long2022autoip,fang2023solving}. The first approach modifies the GP prior directly to reflect physical laws. For example, if the differential operator $\mc{N}_{\bm{x}}$ is linear or linearized, one can apply the differential operator $\mc{N}_{\bm{x}}$ to the GP prior, yielding a new GP:
\begin{equation*}
	\mc{N}_{\bm{x}}[u](\bm{x})\sim \mc{GP}\left(
	\mc{N}_{\bm{x}}m(\bm{x}), \mc{N}_{\bm{x}}\mc{N}_{\bm{x'}}k(\bm{x},\bm{x}')\right).
\end{equation*}
One can then condition the GP not only on observed data, but also on synthetic observations enforcing $\mc{N}_{\bm{x}}[u]\approx 0$ at selected collocation points $\{\bm{x}_j\}$. The second approach retains the standard Gaussian process prior while incorporating physical constraints as additional virtual observations by augmenting the likelihood. Specifically, we define the joint distribution as:
\begin{equation*}
	p(r, y, u, f) = p(u) p(f) p(y|u) p(r|u,f),
\end{equation*}
where $p(u)$ and $p(f)$ are Gaussian priors, $y$ denotes the solution observation, $r$ denotes the physical residual, and $p(r|u,f)$ reflects the likelihood of satisfying the governing equations given $u$ and $f$. Since exact inference in this extended model is generally intractable, especially when non-Gaussian terms are present, variational inference is commonly employed to approximate the posterior over 
$u$ and $f$.

Although Gaussian processes provide a principled way to quantify uncertainty, they can struggle with scalability and complex, highly nonlinear PDEs. These challenges motivate the use of alternative methods, such as latent variable models, which we discuss next.

\subsubsection{Latent Variable Models}  
Latent variable models (LVMs) introduce hidden stochastic variables to capture underlying structures and complex dependencies in data, making them powerful tools for representing uncertainty and extracting low-dimensional features in PDE solutions. Broadly speaking, LVMs can be categorized into Bayesian and non-Bayesian approaches.
\paragraph{Bayesian Latent Variable Models (Bayesian LVMs)}
Bayesian LVMs introduce unobserved stochastic variables to capture hidden structures and underlying factors in data. A typical formulation defines the joint distribution:
        \begin{equation*}
            p(\bm{x},\bm{z},u) = p(\bm{x})p(\bm{z})p(u|\bm{x},\bm{z}),
        \end{equation*}
        where $p(\bm{z})$ is a prior distribution over the latent space (commonly standard Gaussian), $p(\bm{x})$  denotes the marginal distribution of inputs, often assumed fixed, and $p(u|\bm{x},\bm{z})$ is a conditional decoder that maps $\bm{z}$ and $\bm{x}$  to the solution $u$. Then the PDE information can be enforced on the decoder output $u\sim p(u|\bm{x},\bm{z})$ by incorporating a physics-informed regularization term in the loss function, such as the expected PDE residual:
        \begin{equation*}
            \mc{L}_{\mathrm{PDE}} = \mbe_{q(\bm{z}|\bm{x},u)}\left[\Vert \mc{N}_{\bm{x}}[u](\bm{x},\bm{z}) - f(\bm{x},\bm{z})\Vert^2_{L^2(\Omega)} \right].
        \end{equation*}
        Again, the posterior distribution $p(\bm{z}|\bm{x},u)$ is typically intractable due to the nonlinear dependence of $u$ on $\bm{z}$, and is thus approximated by a variational distribution $q(\bm{z}|\bm{x},u).$ Several successful applications include PI-VAE \cite{zhong2023pi} and neural process \cite{garnelo2018neural}. These models follow a Bayesian learning paradigm by explicitly defining prior and approximate posterior distributions and optimizing an evidence lower bound (ELBO). 
\paragraph{Non-Bayesian Latent Variable Models (Non-Bayesian LVMs)}
Non-Bayesian LVMs also employ latent variables to capture essential structure in the solution but do not perform 
explicit Bayesian posterior inference. Instead, these methods typically optimize alternative objectives, such as information-theoretic 
criteria or adversarial losses, to learn meaningful latent representations. A prominent example is the 
Information Bottleneck (IB) method \cite{guo2023ib,cheng2024bi}, which seeks to compress input information while preserving predictive relevance by maximizing
\[
I(Z;Y) - \beta \cdot I(Z;X),
\]
where $I(\cdot;\cdot)$ denotes mutual information and $\beta$ controls the trade-off between compression and accuracy. Although IB introduces latent variables and often uses variational approximations, it does not define or infer a probabilistic posterior, distinguishing it from Bayesian approaches.
Other non-Bayesian LVMs include generative adversarial networks \cite{yang2020physics} and adversarial LVM \cite{yang2019adversarial}, which leverage adversarial training to learn latent distributions without explicit posterior distributions. These models are generally more scalable and flexible, allowing them to handle high-dimensional, complex problems. 

Despite the practical advantages of LVM approaches, many of them rely on heuristics and may struggle to capture complex, function-level uncertainty, especially in the context of partial differential equation (PDE) learning. To address this gap, now we will propose a hybrid framework that couples latent variable modeling with Gaussian process priors, coined as LVM-GP model.

\section{Methodology}
 In this section, we present the technical detail of our main method -- LVM-GP. The new method enables the construction of expressive stochastic mappings from input functions to latent representations, while preserving the flexibility and uncertainty-awareness of nonparametric Gaussian processes. By integrating physical constraints into the training objective, the resulting model achieves reliable uncertainty quantification for both forward and inverse PDE problems with noisy data.
\subsection{Motivation}
To provide some intuitions for the key idea of our method, we first take the following simple PDE problem as an example, namely, 
\begin{equation*}
	\begin{aligned}
		&\lambda u_{xx} = f, &\quad x\in[a,b],
	\end{aligned}
\end{equation*}
with Dirichlet boundary condition. Here $f$ is unknown function and only some noisy data 
are available. The goal is to find the solution $u(x)$ and the unknown function $f(x)$. It is easy to know that the solution $u$ is a double integral of $f$:
\begin{equation*}
	u(x) = u(a) + u'(a)(x-a) + \frac{1}{\lambda}\int_a^x\int_a^s f(t)\,\mathrm{d}t\,\mathrm{d}s,
\end{equation*}
where 
\begin{equation*}
	u'(a) = \frac{1}{b-a} \left(u(b)-u(a) - \frac{1}{\lambda}\int_a^b \int_a^s f(t)\,\mathrm{d}t \,\mathrm{d}s\right).
\end{equation*}
Since the information of $f$ is incomplete, it is impossible to obtain the exact solution of $u$. Instead, given the observation data of $f$, denoted by $\{(x_f^j, f(x_f^j))\}_{j=1}^{N_f}$, one can construct a stochastic surrogate model (e.g. Gaussian process, Bayesian neural network) to approximate $f$ and characterize its associated uncertainty.

It is important to note that $u$ involves a double integration of $f$, meaning that the uncertainty of $u$ is not merely a local reflection of the uncertainty in $f$, but rather an aggregation of the uncertainty of $f$ across the entire domain. In fact, if we denote the predictive variance of $f$ by $\text{Var}(f(x))$, then the predictive variance of $u(x)$ can be heuristically understood as
\[
\text{Var}(u(x)) \sim \iint_{\Omega_{x}} \text{Var}(f(\xi)) \, \mathrm{d}\xi,
\]
where $\Omega_{x}$ denotes the relevant domain of integration. This relation illustrates that local uncertainties in $f$ are accumulated through the integration process to influence the uncertainty of $u$ at each point.

These observations highlight the intrinsic coupling between the uncertainties of 
$u$ and $f$. Specifically, the uncertainty in 
$u$ arises not only from its own observational data, but also fundamentally from the global uncertainty in 
$f$, as inferred from its limited observations. In the following, we discuss how the proposed framework enables joint quantification of the uncertainties in both 
$u$ and $f$.
\subsection{LVM-GP model for forward PDE problems}
For the PDE (\ref{pinn_pde}), we first focus on the forward problem, i.e., $\lambda$ is known. For notational simplicity, we omit the parameter $\lambda$ in the PDE (\ref{pinn_pde}). We consider the scenario where our available dataset $\mc{D}$ are scattered noisy measurements of $u$, $f$ and $b$ from sensors,
\begin{equation*}
	\mc{D} = \mc{D}_u\cup \mc{D}_f \cup \mc{D}_b,
\end{equation*}
where $\mc{D}_u = \{\bm{x}_u^{(i)}, \bar{u}^{(i)}\}_{i=1}^{N_u}$, $\mc{D}_f = \{ \bm{x}_f ^{(i)}, \bar{f}^{(i)}\}_{i=1}^{N_f}$ and $\mc{D}_b = \{\bm{x}_b^{(i)}, \bar{b}^{(i)}\}_{i=1}^{N_b}$. We assume that the measurements are contaminated with zero-mean Gaussian noise, namely,
\begin{equation}
	\begin{aligned}
		 & \bar{u}^{(i)} = u(\bm{x}_u^{(i)}) + \epsilon_u^{(i)}, \quad i=1,2,\cdots, N_u,   \\
		 & \bar{f}^{(i)} = f(\bm{x}_f^{(i)}) + \epsilon _f ^{(i)},\quad i=1, 2,\cdots, N_f, \\
		 & \bar{b}^{(i)} = b(\bm{x}_b^{(i)}) + \epsilon_b^{(i)},\quad i=1,2,\cdots, N_b,
	\end{aligned}
\end{equation}
where $\epsilon_u^{(i)}, \epsilon_f^{(i)}$ and $\epsilon_b^{(i)}$ are independent Gaussian noises with zero mean. The fidelity of each sensor may be known or unknown.

\begin{figure}[!h]
	\centering 
	\includegraphics[width=0.7\textwidth]{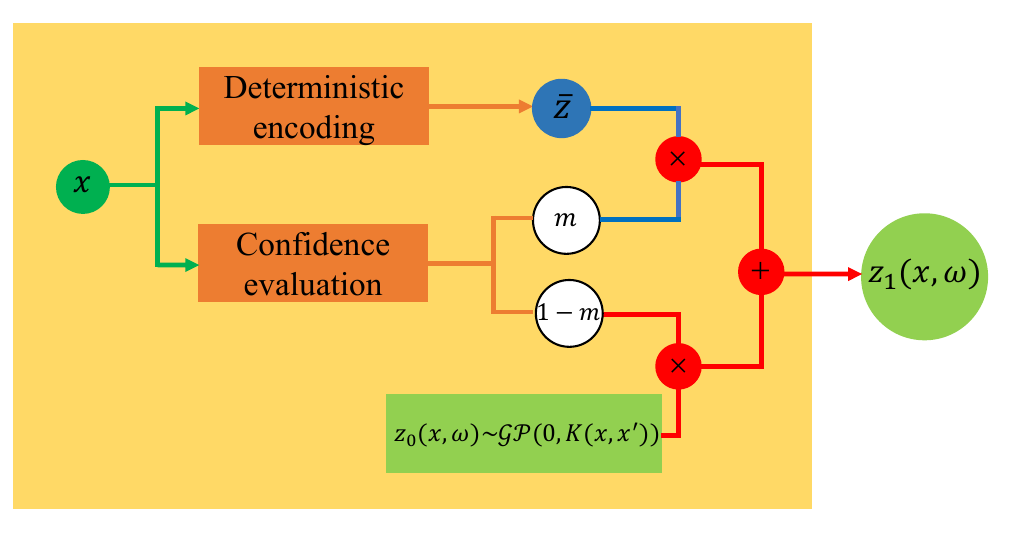}
	\caption{The encoder structure of the proposed approach.}
	\label{fig:encoder}
\end{figure}
In this work, we model the target solution $u$ as a stochastic field. While conventional latent variable models introduce low-dimensional latent random variables to capture stochasticity, they often ignore potential spatial or functional correlations.
Here, we model the latent variable as a high-dimensional stochastic process, providing a functional and infinite-dimensional latent space that naturally encodes correlations and variability. 
Specifically, as shown in Figure \ref{fig:encoder}, we model the encoder
$q_E(\bm{z}|\bm{x};\theta_E)$ as
\begin{equation}
	\bm{z}(\bm{x};\theta_E) = \mathrm{diag}(m(\bm{x};\theta_m))\bar{\bm{z}}(\bm{x};\theta_{\bar{\bm{z}}}) + \mathrm{diag}(1 - m(\bm{x};\theta_m))\bm{z}_0,
	\label{hidden_variable}
\end{equation}
where $m(\bm{x};\theta_m)$ and $\bar{\bm{z}}(\bm{x};\theta_{\bar{\bm{z}}})$ are deep neural networks, $\theta_E =\{\theta_m, \theta_{\bar{\bm{z}}}\}$. The range of each element of $m(\bm{x};\theta_m)$ is $[0,1]$ and 
$\bm{z}_0:\mathbb{R}^d\to \mathbb{R}^{d_z}$ is a vector-valued Gaussian process with zero mean and a squared exponential covariance kernel. That is,
\begin{equation*}
	\bm{z}_0 \sim \mathcal{GP}(0, K(\bm{x},\bm{x}')),
	\label{gp:z0}
\end{equation*}
where for any two inputs $\bm{x},\bm{x}'\in\mathbb{R}^{d}$,  $K(\bm{x},\bm{x}')$ is a $d_z\times d_z$ matrix-valued covariance function. Specifically, we assume that the outputs are independent across dimensions, and each component function $\bm{z}_{0,i}$ shares the same scalar kernel, namely, 
\begin{equation*}
	K(\bm{x},\bm{x}') = \sigma_K^2 \exp\left(-\frac{\|\bm{x}-\bm{x}'\|^2}{2\ell^2}\right)\bm{I}_{d_z},
\end{equation*}
where $\bm{I}_{d_z}$ denotes the $d_z$-dimensional identity matrix, $\sigma_K^2$ is the signal variance, and $\ell>0$ is the length-scale parameter. Without loss of generality, in this work we set $\sigma_K$ to be 1.
 
Here $\bar{\bm{z}}(\bm{x};\theta_{\bar{\bm{z}}})$ is a deterministic transformation and $m(\bm{x};\theta_m)$ serves as a confidence function that can be interpreted as the probability of $\bm{x}$ that is far from the training data distribution. Ideally, when the encoder is well-trained, $m(\bm{x})=1$ implies that $\bm{x}$ is close to one of training data and $\bm{z}$ becomes a deterministic latent representation. Conversely, if $m(\bm{x})=0$, $\bm{x}$ is considered distant from the observed data, which leads to increased uncertainty in $\bm{z}$. Notably, for a given $\bm{x}$, the variance of $\bm{z}(\bm{x})$ is given by  $ (1- m(\bm{x}))^2.$ From this perspective, $m(\bm{x})$ can be also interpreted as an alternative parameterization of a Gaussian process using a non-stationary kernel. Consequently, the latent variable $\bm{z}$ captures the distributional uncertainty associated with $\bm{x}$. Recalling that the uncertainty of $u$ at a given point is influenced by the global uncertainty of $f$ across the entire domain. To address the intrinsic coupling in the system, we propose an integral-based modeling framework, which establishes an infinite-dimensional mapping between a Gaussian process and the target stochastic field. Specifically, we represent the stochasticity by introducing a latent variable $\bm{z}(\bm{x}, \bm{\omega}_E)$, where $\bm{\omega}_E \sim \mathcal{N}(0, \bm{I}_M)$, and adopt a neural operator style architecture, formulated as follows:
\begin{equation}
	\begin{aligned}
		&\bm{z}_{i}(\bm{x}, \bm{\omega}_E) = \sigma \left(
		W_i \bm{z}_{i-1}(\bm{x}, \bm{\omega}_E) + b_i + \int_{\Omega} k_i(\bm{x}, \bm{x}') \bm{z}_{i-1}(\bm{x}', \bm{\omega}_E) \, \mathrm{d}\bm{x}' \right), \quad i=2,\ldots,L-1,\\
		&\bm{z}_{L}(\bm{x}, \bm{\omega}_E) = W_L \bm{z}_{L-1}(\bm{x}, \bm{\omega}_E) + b_L,
	\end{aligned}
\end{equation}
with $\bm{z}_1(\bm{x}, \bm{\omega}_E) = \bm{z}(\bm{x}, \bm{\omega}_E)$ (See \eqref{hidden_variable} for the definition of $\bm{z}$). Here, $\{W_i, b_i\}_{i=2}^{L}$ are learnable affine transformation parameters, and $k_i(\bm{x}, \bm{x}')$ denotes a kernel function. In this work, we employ a Positional Transformer kernel \cite{chen2024positional} of the form
\begin{equation}
k_i(\bm{x}, \bm{x}') = \frac{\exp\left(-\alpha_i \|\bm{x} - \bm{x}'\|^2\right)}{\int_\Omega \exp\left(-\alpha_i \|\bm{x} - \bm{y}\|^2\right) \mathrm{d}\bm{y}} V_i,
\label{decoder_kernel}
\end{equation}
where $\alpha_i > 0$ is a length-scale parameter and $V_i$ is a learnable projection matrix. And the above integral can be approximated via classic numerical algorithms (Gaussian quadrature, FFT, etc.) or Monte Carlo techniques. We refer the reader to \cite{chen2024positional} for further technical details. 
Let $\theta_D^u \coloneqq \{\alpha_i, V_i, W_i, b_i\}_{i=2}^{L-1} \cup \{W_L, b_L\}$ denote the set of learnable parameters. Then the final output $\mu_u(\bm{x}, \bm{\omega}_E; \theta_E,\theta_D^u) \coloneqq \bm{z}_L(\bm{x}, \bm{\omega}_E)$ is interpreted as the predicted mean of the target field $u$.

To further capture the inherent stochasticity, we model $u$ as conditionally Gaussian, with the mean is given by $\mu_u(\bm{x},\bm{\omega}_E;\theta_E,\theta_D^u)$, and the variance $\sigma_u(\bm{x};\theta_{\sigma}^u)$ can be either fixed or parameterized and learned from data, namely, 
\begin{equation}
q_D^u(u|\bm{x},\bm{\omega}_E;\theta_E,\theta_D^u,\theta_{\sigma}^u) \sim \mathcal{N}\left(\mu_u(\bm{x},\bm{\omega}_E;\theta_E,\theta_D^u),\sigma_u^2(\bm{x};\theta_{\sigma}^u)\right).	
\label{decoder_u}
\end{equation}
This formulation allows us to jointly characterize both the predictive mean and uncertainty of $u$ by leveraging the latent distributional representation $\bm{z}$. Again, we use $\omega_{D}^u\sim \mathcal{N}(0,1)$ to represent the randomness in the conditional Gaussian model. Thus, the output of the model $u(\bm{x},\bm{\omega}_u;\theta_u)$ can be expressed as
\begin{equation}
	\begin{aligned}
	\bm{z}_0 &\sim \mathcal{GP}(0, K(\bm{x}, \bm{x}')), \\
	\bm{z}_1(\bm{x}; \theta_E) &= \mathrm{diag}(m(\bm{x}; \theta_m)) \bar{\bm{z}}(\bm{x};\theta_{\bar{\bm{z}}}) + \mathrm{diag}(1 - m(\bm{x}; \theta_m)) \bm{z}_0, \\
	\bm{z}_{i}(\bm{x}, \bm{\omega}_E) &= \sigma \left(
		W_i \bm{z}_{i-1}(\bm{x}, \bm{\omega}_E) + b_i + \int_{\Omega} k_i(\bm{x}, \bm{x}') \bm{z}_{i-1}(\bm{x}', \bm{\omega}_E) \, \mathrm{d}\bm{x}' \right), \quad i=2,\ldots,L-1,\\
		\bm{z}_{L}(\bm{x}, \bm{\omega}_E) &= W_L \bm{z}_{L-1}(\bm{x}, \bm{\omega}_E) + b_L,\\
	u(\bm{x}, \bm{\omega}_u;\theta_u) &= \bm{z}_L(\bm{x}, \bm{\omega_E}) + \sigma_u(\bm{x};\theta_{\sigma}^u)\cdot \omega_D^u,\quad \theta_u = \{\theta_E, \theta_D^u,\theta_{\sigma}^u\}, \,\bm{\omega}_u=\{\bm{\omega}_E, \omega_D^u\}.
\end{aligned}
\label{forward_FNO_type_model}
	\end{equation}
	
We now incorporate the PDE information \eqref{pinn_pde} into the model. Note that the noise $\sigma_u(\bm{x};\theta_{\sigma}^u)$ models the aleatoric uncertainty which can not be reduced by having more observations. To avoid the need for decoding the noise into the PDE, we make the assumption that the term $\sigma_u(\bm{x};\theta_\sigma^u)$ is not differentiated with respect to $\bm{x}$. Under this assumption, the computation of derivatives in (\ref{pinn_pde}) can be efficiently
carried out using automatic differentiation:
\begin{equation*}
  \begin{aligned}
    \mathcal{N}_{\bm{x}}\left[u(\bm{x},\bm{\omega}_u;\theta_u)\right] &= \mathcal{N}_{\bm{x}}\left[\mu_u(\bm{x},\bm{\omega}_E;\theta_E,\theta_D^u)\right] = \mathcal{N}_{\bm{x}}\left[\bm{z}_L(\bm{x},\bm{\omega}_E;\theta_D^u)\right],\\
    \mathcal{N}_{\bm{x}}\left[\bm{z}_i(\bm{x},\bm{\omega}_E)\right] &= \sigma'\left(W_i \bm{z}_{i-1}(\bm{x}, \bm{\omega}_E) + b_i + \int_{\Omega} k_i(\bm{x}, \bm{x}') \bm{z}_{i-1}(\bm{x}', \bm{\omega}_E) \, \mathrm{d}\bm{x}'\right) \\
    &\quad \cdot \left(W_i\mathcal{N}_{\bm{x}}[\bm{z}_{i-1}(\bm{x},\bm{\omega}_E)] + \int_{\Omega} \mathcal{N}_{\bm{x}}[k_i(\bm{x},\bm{x}')]\bm{z}_{i-1}(\bm{x}',\bm{\omega}_E)\,\mathrm{d}\bm{x}'\right), \quad i=2,\cdots,L-1,\\
    \mathcal{N}_{\bm{x}}\left[\bm{z}_1(\bm{x},\bm{\omega}_E)\right] &= \mathcal{N}_{\bm{x}}\left[\mathrm{diag}(m(\bm{x};\theta_m)) \bar{\bm{z}}(\bm{x};\theta_{\bar{\bm{z}}})\right] + \mathcal{N}_{\bm{x}}\left[\mathrm{diag}(1 - m(\bm{x};\theta_m)) \bm{z}_0(\bm{x},\bm{\omega}_E)\right].
  \end{aligned}
\end{equation*}

The computation of the derivative of $\bm{z}_1(\bm{x}, \bm{\omega}_E)$ entails evaluating the derivative of the underlying Gaussian process $\bm{z}_0(\bm{x}, \bm{\omega}_E)$. Detailed derivations are provided in \ref{sec:gp_derivative}.
Here we employ a conditional Gaussian to model the decoder for $f$ and $b$. Specifically, the mean predictions of $f$ and $b$ can be expressed as 
\begin{equation}
	\begin{aligned}
		 & \mu_f(\bm{x},\bm{\omega}_E;\theta_E,\theta_D^u) = \mc{N}_{\bm{x}}\left[\mu_u(\bm{x},\bm{\omega}_E;\theta_E,\theta_D^u)\right],\\
		 & \mu_b(\bm{x},\bm{\omega}_E;\theta_E,\theta_D^u) = \mc{B}_{\bm{x}}\left[\mu_u(\bm{x},\bm{\omega}_E;\theta_E,\theta_D^u)\right].\\
	\end{aligned}
	\label{physics}
\end{equation}
 We use $\sigma_f(\bm{x};\theta_{\sigma}^f)$ and $\sigma_b(\bm{x};\theta_{\sigma}^b)$ to model the aleatoric uncertainty of $f$ and $b$, respectively. Then the whole model can be expressed as
\begin{equation}
	\begin{aligned}
		 f(\bm{x},\bm{\omega}_f;\theta_f) &= \mc{N}_{\bm{x}}\left[\mu_u(\bm{x},\bm{\omega}_E;\theta_E,\theta_D^u)\right] + \sigma_f(\bm{x};\theta_{\sigma}^f)\cdot \omega_D^f, \; \bm{\omega}_f = \{\bm{\omega}_E,\omega_D^f\}, \theta_f = \{\theta_E,\theta_D^u,\theta_{\sigma}^f\},\\
		 b(\bm{x},\bm{\omega}_b;\theta_b) &= \mc{B}_{\bm{x}}\left[\mu_u(\bm{x},\bm{\omega}_E;\theta_E,\theta_D^u)\right] + \sigma_b(\bm{x};\theta_{\sigma}^b)\cdot \omega_D^b, \; \bm{\omega}_b = \{\bm{\omega}_E,\omega_D^b\}, \theta_b = \{\theta_E,\theta_D^u,\theta_\sigma^b\}.\\
	\end{aligned}
	\label{decoder_f_b}
\end{equation}
In other words, the corresponding decoder models are
\begin{equation*}
  \begin{aligned}
    &q_D^f(f|\bm{x},\bm{\omega}_E;\theta_f) \sim \mathcal{N}\left(\mu_f(\bm{x},\bm{\omega}_E;\theta_E,\theta_D^u),\sigma_f^2(\bm{x};\theta_{\sigma}^f)\right),\\	
    &q_D^b(b|\bm{x},\bm{\omega}_E;\theta_b) \sim \mathcal{N}\left(\mu_b(\bm{x},\bm{\omega}_E;\theta_E,\theta_D^u),\sigma_b^2(\bm{x};\theta_{\sigma}^b)\right),	
  \end{aligned}
\end{equation*}
Then we can define the following negative log-likelihood loss function:
\begin{equation*}
  \begin{aligned}
	\mathcal{L}_{\mathrm{data}}   &= \mathcal{L}_{\mathrm{data},u} + \mathcal{L}_{\mathrm{data},f} + \mathcal{L}_{\mathrm{data},b},\\
    \mathcal{L}_{\mathrm{data},u} &=  \frac{1}{N_uN_{\bm{\omega}}}\sum_{i=1}^{N_u}\sum_{j=1}^{N_{\bm{\omega}}} \log q_D^{u}(\bar{u}^{(i)}|\bm{x}_u^{(i)}, \bm{\omega}_E^{(j)};\theta_u),\quad \bm{\omega}_E^{(j)}\sim \mc{N}(0, \bm{I}_{M}),\\
    \mathcal{L}_{\mathrm{data},f} &=  \frac{1}{N_fN_{\bm{\omega}}}\sum_{i=1}^{N_f}\sum_{j=1}^{N_{\bm{\omega}}} \log q_D^{f}(\bar{f}^{(i)}|\bm{x}_f^{(i)}, \bm{\omega}_E^{(j)};\theta_f),\quad \bm{\omega}_E^{(j)}\sim \mc{N}(0, \bm{I}_{M}),\\
    \mathcal{L}_{\mathrm{data},b} &=  \frac{1}{N_bN_{\bm{\omega}}}\sum_{i=1}^{N_b}\sum_{j=1}^{N_{\bm{\omega}}} \log q_D^{b}(\bar{b}^{(i)}|\bm{x}_b^{(i)}, \bm{\omega}_E^{(j)};\theta_b),\quad \bm{\omega}_E^{(j)}\sim \mc{N}(0, \bm{I}_{M}).\\
  \end{aligned}
  \label{forward_data_loss}
\end{equation*}

It is observed that when only data loss is used, the latent variable $\bm{z}(\bm{x};\bm{\omega}_E)$ tends to reduce uncertainty by pushing $m(\bm{x}_i)\to 1$ ($\{\bm{x}_i\}$ are observation data), causing the latent Gaussian process $\bm{z}$ to collapse into a deterministic mapping in the whole domain. Hence we employ the idea in IB-UQ \cite{guo2023ib} to regularize the model. Specifically, note that for any given $\bm{x}$, the latent representation $\bm{z}$ (or $\bm{z}_1$) follows the distribution below:
\begin{equation*}
  q_E(\bm{z}| \bm{x}) \sim \mathcal{N}\left(m(\bm{x})\bar{\bm{z}}(\bm{x}),  \diag (1-m(\bm{x}))^2\right).
\end{equation*}
Thus we can compute the KL divergence between $q_E(\bm{z}|\bm{x})$ and marginal distribution $e(\bm{z})$ on the entire physical domain as a regularization term, namely, 
\begin{equation}
    \mc{L}_{\mathrm{reg}}(Z,X) = \mathbb{E}_{\bm{x}\sim\mc{U}(\Omega)} D_{\mathrm{KL}}\left[q_E(\bm{z}|\bm{x}) \| e_1(\bm{z})\right].
    \label{mutual_information_Z_X_independent}
\end{equation}

For simplicity, in this work, we take $e_1(\bm{z})$ as a standard Gaussian distribution, i.e., $e_1(\bm{z})\sim \mathcal{N}(0,\bm{I}_{d_z})$. 

In practice, we use empirical loss to evaluate the integral involved in regularization term. For $\mathcal{L}_{\mathrm{reg}}(Z,X)$, we have the following approximation:
\begin{equation}
  \begin{aligned}
    \mathcal{L}_{\mathrm{reg}} = \frac{1}{N}\sum_{i=1}^{N} \left[\log q_E(\bm{z_i}|\bm{x}_i) - \log e_1(\bm{z_i})\right], \quad \bm{z}_i\sim q_E(\cdot|\bm{x}_i), \quad \bm{x}_i\sim \mc{U}(\Omega),
    \end{aligned}
	\label{regularization_independent}
\end{equation}
where $\mc{U}(\Omega)$ denotes the uniform distribution over the domain $\Omega$.   
Finally, we can maximize the overall loss function as follows:
\begin{equation}
	\mathcal{L} = \mathcal{L}_{\mathrm{data}} - \beta \cdot  \mathcal{L}_{\mathrm{reg}},
    \label{forward_loss}
\end{equation}
here $\beta$ is a hyper-parameter that balances the trade-off between the data loss and the regularization term. This objective allows us to jointly optimize the model parameters and latent variables in a non-Bayesian manner, without requiring posterior inference.

\begin{figure}[!h]
	\centering
	\includegraphics[width=0.9\linewidth]{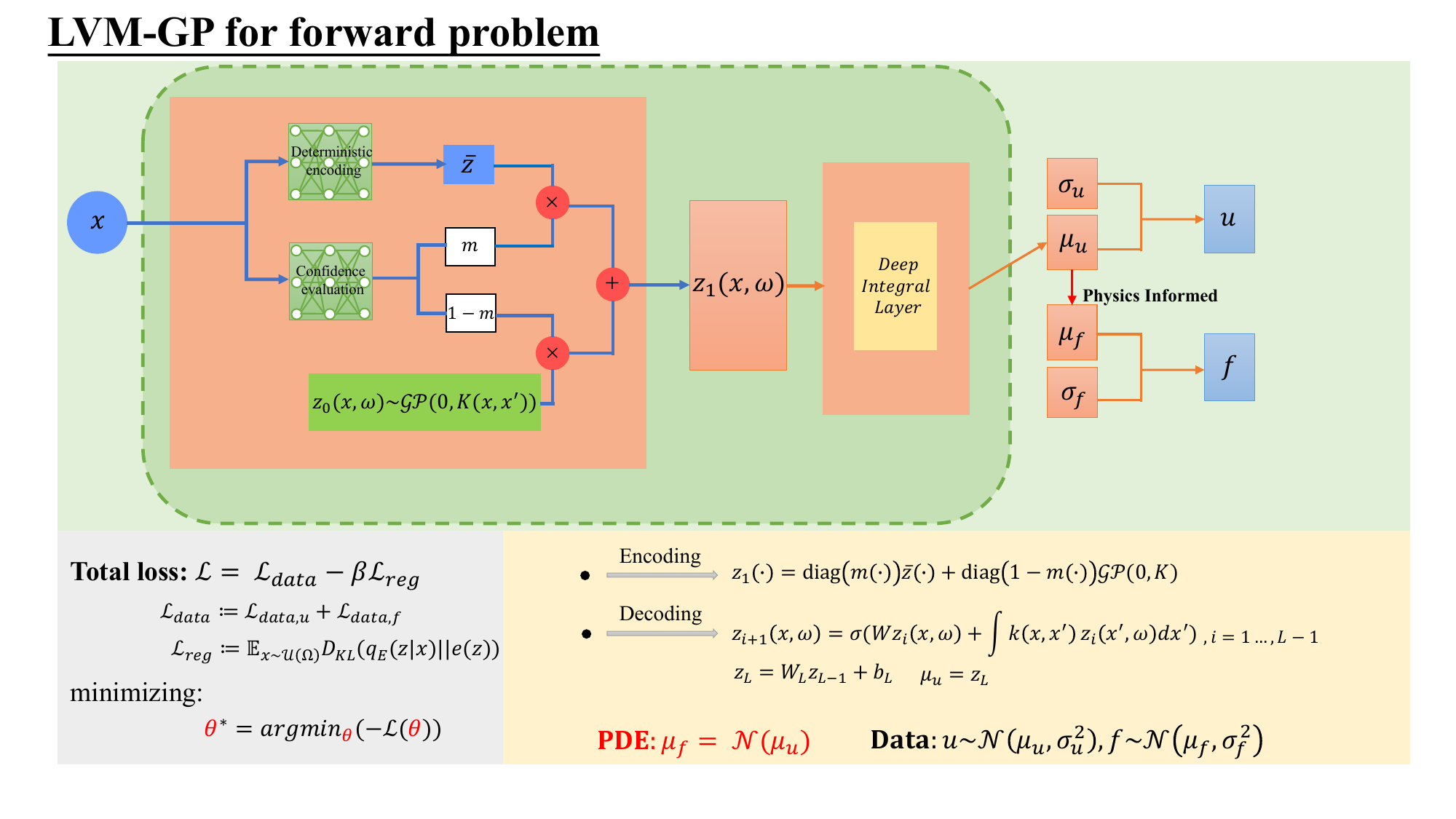}
	\caption{LVM-GP for forward PDE problems. The input $\bm{x}$ is first processed by a confidence-aware encoder, which combines deterministic features $\bar{\bm{z}}$ and spatially varying uncertainty $\bm{z}_0(\bm{x},\bm{\omega})$ (defined by a Gaussian process) through a confidence function $m$. Then the latent representation $\bm{z}_1(\bm{x};\bm{\omega})$ is propagated through a deep integral operator, which models nonlinear mappings in function space and enables the representation of complex PDE solutions. This operator outputs a latent variable $\bm{z}_L(\bm{x};\bm{\omega})$ from which both the predicted solution $u$ and its PDE residual $f$ are jointly inferred via learned mean and variance functions.}
	\label{fig:forward_structure}
\end{figure}

The detail schematic of our LVM-GP model is shown in Figure \ref{fig:forward_structure} (without loss of generality, we omit the module corresponding to $b$). And the corresponding algorithm is summarized in Algorithm \ref{algorithm}.

\begin{algorithm}[H]
	\begin{itemize}
		\item\textbf{1.} Specify the training set $\{\bm{x}_u^{(i)}\}_{i=1}^{N_u}, \{\bm{x}_f^{(i)}\}_{i=1}^{N_f}$ and $\{\bm{x}_b^{(i)}\}_{i=1}^{N_b}$, parameter $\beta$, learning rate $\eta$
		\item\textbf{2.} Sample random variables $\bm{\omega}_E^{(j)} \sim \mc{N}(0,\bm{I}_M), j=1,\cdots,N_{\bm{\omega}}$, and calculate the loss $\mc{L}$ via \eqref{forward_loss}
		\item\textbf{3.} Update all the involved parameters $W$ of $\theta_u,\theta_f,\theta_b$ in (\ref{forward_loss}) using the Adam optimizer
		      \begin{equation*}
			      W = W + lr \cdot \frac{\partial {\mc{L}}}{\partial W}
		      \end{equation*}
		\item\textbf{4.} Repeat \textbf{Step 2-3} until convergence
	\end{itemize}
	\caption{LVM-GP for forward PDE problems}
	\label{algorithm}
\end{algorithm}

\begin{remark}
	As shown in \eqref{forward_FNO_type_model}, we employ an FNO-type architecture for the decoder, which can be regarded as a specific type of neural operator. The integral computation involved in this formulation has a computational complexity of $\mathcal{O}(N^2)$, where $N$ denotes the number of integration mesh points. This cost can be significantly reduced using the Fast Fourier Transform (FFT), resulting in a more efficient complexity of $\mathcal{O}(N \log N)$ (see FNO \cite{li2020fourier}). Alternatively, we can use a DeepONet-type structure to model the decoder. Specifically, the mean prediction of $u$ can be expressed as 
	\begin{equation*}
		\begin{aligned}
			\bm{z}_0(\bm{x},\bm{\omega}_E) & \sim \mathcal{GP}(0, k(\bm{x},\bm{x}')), \\
			 \bm{z}(\bm{x},\bm{\omega}_E) & = m(\bm{x})\bar{\bm{z}}(\bm{x}) + (1 - m(\bm{x})) \bm{z}_0(\bm{x};\bm{\omega}_E), \\ 
			 \mu_u(\bm{x},\bm{\omega}_E) & = \mathrm{Branch}(\bm{z}(\bm{x}_{\mathrm{grid},1},\bm{\omega}_E),\cdots, \bm{z}(\bm{x}_{\mathrm{grid},p},\bm{\omega}_E))\cdot \mathrm{Trunk}(\bm{z}(\bm{x},\bm{\omega}_E)),\\
			u(\bm{x},\bm{\omega}_u) &= \mu_u(\bm{x}, \bm{\omega_E}) + \sigma_u(\bm{x}, \theta^{u}_{\sigma})\cdot \omega_D^u,\quad \bm{\omega}_u=\{\bm{\omega}_E, \omega_D^u\}.
		\end{aligned}
		\end{equation*}
		where $\mathrm{Branch}:\mathbb{R}^{p\times d_z}\to \mathbb{R}^h$, $\mathrm{Trunk}:\mathbb{R}^{d_z}\to \mathbb{R}^{h}$ are neural networks, $\{\bm{x}_{\mathrm{grid}, 1},\cdots,\bm{x}_{\mathrm{grid}, p}\}$ are predefined mesh points in the domain. It is worth noting that the input dimension of the Branch network is $p\times d_z$, which can be significantly large when $d_z$ and $p$ are large.	
\end{remark}

\begin{remark}
	In this section, we present a model for the noise parameters $\sigma_u$, $\sigma_f$, and $\sigma_b$ that
	accounts for their dependence on the input, resulting in a heteroscedastic noise model.
	In practice, the heteroscedastic noise \cite{kendall2017uncertainties,le2005heteroscedastic} can be implemented by applying a
	softplus or exponential transformation to the output. Alternatively, we can model the
	noise parameters using scalar, namely, $\sigma_u$,
	$\sigma_f$, and $\sigma_b$ are either pre-determined or can be learned during
	the training process.
\end{remark}

\begin{remark}
Note that the regularization term \eqref{mutual_information_Z_X_independent} does not capture the interactions between different inputs $\bm{x}$. In fact, we may consider the following revised regularization term:
\begin{equation}
		\widetilde{\mc{L}}_{\mathrm{reg}}(Z,X) = \frac{1}{B}\mathbb{E}_{\bm{x}_{1:B}\sim \mc{U}(\Omega^B)}D_{\mathrm{KL}}\big[q_E(\bm{z}_{1:B}| \bm{x}_{1:B})\| e_2(\bm{z}_{1:B}|\bm{x}_{1:B})\big],
        \label{mutual_information_Z_X_dependent}
\end{equation}
where $\bm{x}_{1:B} = \{\bm{x}_1,\cdots,\bm{x}_B\}$,  $\bm{z}_{1:B}=(\bm{z}(\bm{x}_1), \cdots, \bm{z}(\bm{x}_B))$ and $e_2(\bm{z}_{1:B}|\bm{x}_{1:B})\sim \mc{N}(0,\bm{K}),$ $\bm{K}=(K(\bm{x}_{i},\bm{x}_j))_{i,j=1,\cdots,B} \in \mathbb{R}^{B\times B}.$  Similarly, for $\widetilde{\mc{L}}_{\mathrm{reg}}(Z,X),$ we can derive exact expression
\begin{equation}
    \widetilde{\mathcal{L}}_{\mathrm{reg}}=\mathbb{E}_{\bm{x}_{1:B}\sim \mathcal{U}(\Omega^B)}\frac{1}{2B}\left[-2 \log \det (\bm{D_m}) - B + \mathrm{tr}(\bm{K}^{-1}\bm{D_m}\bm{K}\bm{D_m}) + (\bm{m\bar{z}})^T\bm{K}^{-1}(\bm{m\bar{z}}) \right],
\label{revised_regularization}
\end{equation}
where $\bm{D}_m = \diag(1-m(\bm{x}_1),\cdots,1-m(\bm{x}_B)), \bm{m\bar{z}}=(m(\bm{x}_1)\bar{z}(\bm{x}_1),\cdots, m(\bm{x}_B)\bar{z}(\bm{x}_B)).$
Hence, minimizing $\widetilde{\mathcal{L}}_{\mathrm{reg}}$ allows for the simultaneous optimization of  $m$ and the hyperparameters of the Gaussian process prior.
This revised regularization term is adopted in the numerical experiments (see Figure \ref{fig:1D_poisson_equation_with_learnable_lc}), where we demonstrate its effectiveness in capturing spatial correlations among inputs $\bm{x}$. It is worth noting that the statistical properties of $\widetilde{\mc{L}}_{\mathrm{reg}}$ in the limit $B \to \infty$ remain to be theoretically analyzed.
\end{remark}

\subsection{LVM-GP model for inverse PDE problems}
In the context of inverse problem, assume that we have some observations for $\lambda$ 
$$\mathcal{D}_{\mb{\lambda}}=\{\bm{x}_{{\lambda}}^{(i)},\bar{\lambda}^{(i)}\}_{i=1}^{N_{\lambda}},\quad \bar{\lambda}^{(i)} = \lambda(\bm{x}_{\lambda}^{(i)}) + \epsilon_\lambda^{(i)}, \quad i=1,2,\cdots, N_{\lambda},   \\
$$ where $\epsilon_\lambda^{(i)}$ is independent Gaussian noise with zero mean. Since the value of $\lambda$ is unknown in this case, we need to remodel both $f$ and $b$. To this end, we introduce a new decoder
\begin{equation}
	q_D^{\lambda}(\lambda|\bm{x},\bm{\omega}_{E};\theta_{\lambda})\sim \mathcal{N}\left(\mu_{\lambda}(\bm{x},\bm{\omega}_E;\theta_D^{\lambda}), \sigma_{\lambda}^2(\bm{x},\theta^{\lambda}_{\sigma})\right),
\end{equation}
where $ \mu_{\lambda}(\bm{x}, \bm{\omega}_E; \theta_D^{\lambda}) $ and $ \sigma_{\lambda}(\bm{x}, \theta^{\lambda}_{\sigma}) $ represent the predicted mean and standard deviation of $ \lambda $, respectively. The modeling structure follows the same design as that used for the solution $ u(\bm{x},\bm{\omega}_E;\theta_u) $.
Notably, both $ \lambda $ and $ u $ share the same encoder $ \bm{\omega}_E $, ensuring that their representations are learned in a consistent latent space.

\begin{figure}[H]
	\centering
	\includegraphics[width=0.85\linewidth]{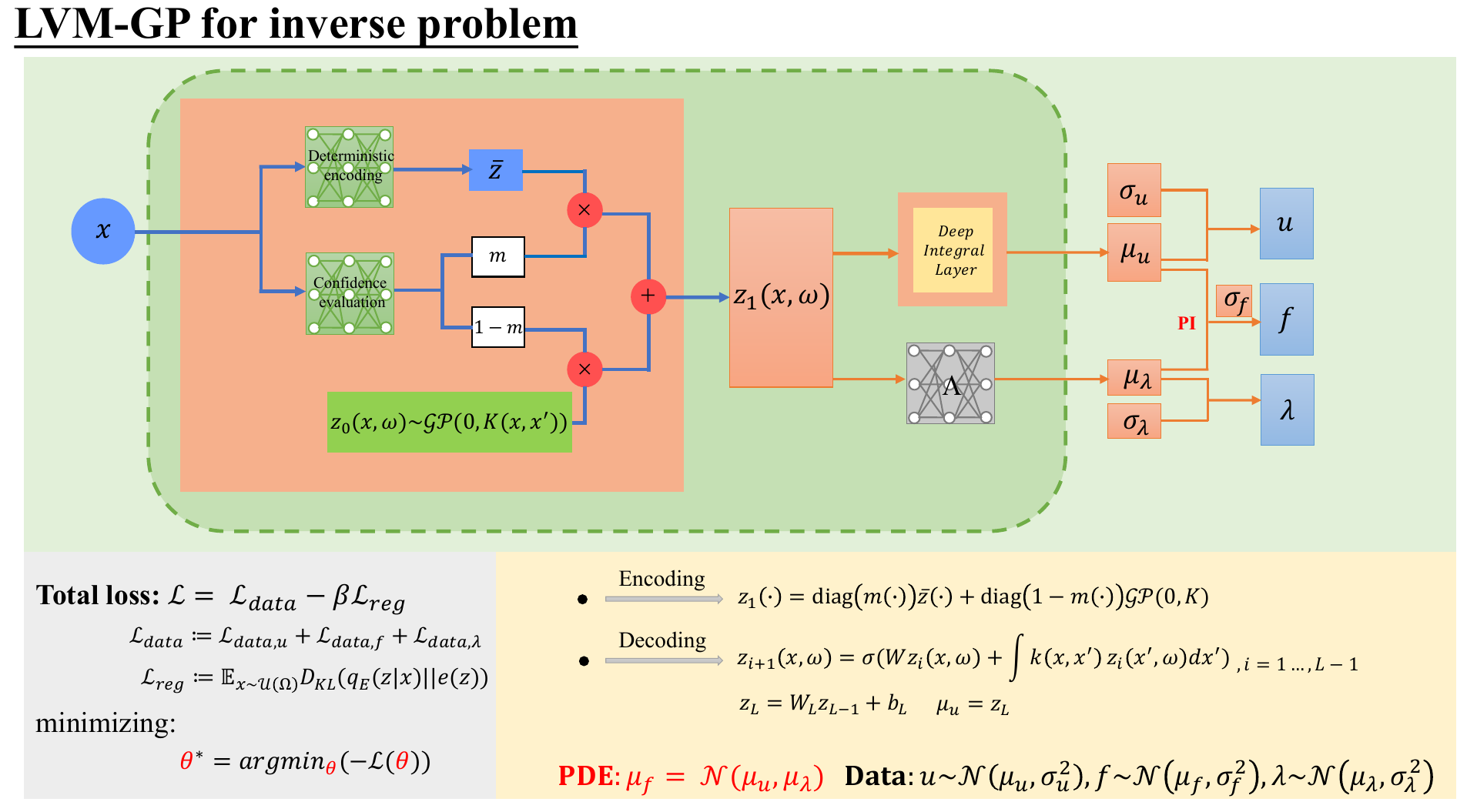}
	\caption{LVM-GP for inverse PDE problem. The input $\bm{x}$ is first processed by a confidence-aware encoder, which combines deterministic features $\bar{\bm{z}}$ and spatially varying uncertainty $\bm{z}_0(\bm{x},\bm{\omega})$ (defined by a Gaussian process) through a confidence function $m$. Then the latent representation $\bm{z}_1(\bm{x};\bm{\omega})$ is propagated through a deep integral operator, which models nonlinear mappings in function space and enables the representation of complex PDE solutions. This operator outputs a latent variable $\bm{z}_L(\bm{x};\bm{\omega})$ from which the predicted solution $u$, unknown vector $\lambda$ and its PDE residual $f$ are jointly inferred via learned mean and variance functions.}
	\label{fig:arch2}
\end{figure}

When $\lambda$ is considered as a vector (as opposed to a function), the following formulation is adopted:
\begin{equation*}
    \mu_{\lambda}(\bm{\omega}_E;\theta_D^{\lambda}) = \int _{\Omega} \mu_{\lambda}(\bm{x},\bm{\omega}_E;\theta_D^{\lambda}) \mathrm{d} \bm{x}.
\end{equation*} Then we can represent $f$ and $b$ as follows:
\begin{equation}
	\begin{aligned}
		  f(\bm{x},\bm{\omega}_f;\theta_f) & = \mc{N}_{\bm{x}}\left[\mu_u(\bm{x},\bm{\omega}_E;\theta_u); \mu_\lambda(\bm{x};\bm{\omega}_{E};\theta_{\lambda})\right] + \sigma_f(\bm{x},\theta_\sigma^f)\cdot \omega_D^f,\\
		  b(\bm{x},\bm{\omega}_b;\theta_b) & = \mc{B}_{\bm{x}}\left[\mu_u(\bm{x},\bm{\omega}_E;\theta_u); \mu_\lambda(\bm{x};\bm{\omega}_{E};\theta_{\lambda})\right] + \sigma_b(\bm{x},\theta_\sigma^b)\cdot \omega_D^b,\\
		   \bm{\omega}_b &= \{\bm{\omega}_E,\omega_D^b\}, \theta_b = \{\theta_u,\theta_\sigma^b\},\quad \bm{\omega}_f = \{\bm{\omega}_E,\omega_D^f\}, \theta_f = \{\theta_u,\theta_\sigma^f\}.
        \end{aligned}
	\label{inverse_physics}
\end{equation}
Hence, we can derive similar loss function
\begin{equation}
	\begin{aligned}
		\mathcal{L}   &= \mathcal{L}_{\mathrm{data},u} + \mathcal{L}_{\mathrm{data},f} + \mathcal{L}_{\mathrm{data},b} + \mathcal{L}_{\mathrm{data},\lambda} - \beta \cdot \mathcal{L}_{\mathrm{reg}},\\
		\mathcal{L}_{\mathrm{data},\lambda} &= \frac{1}{N_{\lambda}N_{\bm{\omega}}}\sum_{i=1}^{N_{\lambda}}\sum_{j=1}^{N_{\bm{\omega}}} \log q_D^{\lambda}(\bar{\lambda}^{(i)}|\bm{x}_{\lambda}^{(i)}, \bm{\omega}_{E}^{(j)};\theta_{\lambda}),\quad \bm{\omega}_{E}^{(j)}\sim \mc{N}(0, \bm{I}_{M}),\\
	  \end{aligned}	
\end{equation}
where $\mathcal{L}_{\mathrm{data},u}, \mathcal{L}_{\mathrm{data},f}, \mathcal{L}_{\mathrm{data},b}, \mathcal{L}_{\mathrm{reg}}$ are defined in \eqref{forward_data_loss} and \eqref{regularization_independent}.
The remaining is same as before. Moreover, similar schematic is also provided in Figure \ref{fig:arch2}.
\begin{remark}
	In the absence of observations for $\lambda$, we may use the predicted mean
	$\mu_{\lambda}(\bm{x}; \bm{\omega}_E; \theta_{\lambda})$ as a surrogate for $\lambda$. In this case, modeling the aleatoric uncertainty of $\lambda$ is unnecessary.
\end{remark}
\section{Numerical examples}
In this section, we report the performance of our LVM-GP method on PDE forward and inverse problems.
For both forward and inverse PDE problems, the LVM-GP model comprises encoder networks $m(\bm{x})$ and $\bar{z}(\bm{x})$, as well as a decoder network. Unless otherwise specified, these networks are implemented as 3-layer fully-connected neural networks with 64 neurons per hidden layer, using Mish \cite{misra2019mish} activation function. Since the hyperparameter $\alpha_i$ in \eqref{decoder_kernel} is positive, here we use 
\begin{equation*}
	\alpha_i = \tan(\tilde{\alpha}_i),\quad 0<\tilde{\alpha}_i<\pi / 2
\end{equation*}
to parameterize it. In the case where input dimension $d_{\bm{x}}=1$, we use the Karhunen-Lo\`eve expansion to represent the Gaussian process prior $\bm{z}_0$ in \eqref{gp:z0} \cite{nobile2008sparse}
\begin{equation*}
	\bm{z}_{0,i}(\bm{x},\bm{\omega}_i) = \omega_{i,1}\left(\frac{\sqrt {\pi }L}{2}\right)^{1/2}+\sum_{j=2}^{N}\zeta_jq_j(\bm{x})\omega_{i,j},\quad i=1,\cdots,d_z,
\end{equation*}
where 
\begin{equation*}
	\begin{aligned}
		\zeta_j & =\left(\sqrt{\pi}L\right)^{1/2}\exp\left(\frac{-(\lfloor j/2\rfloor \pi L)^2}{8}\right), \quad \mbox{if } j>1,\\
		q_j(x) & = \left\{\begin{array}{l}
			\sin(\frac{\lfloor j/2 \rfloor \pi \bm{x}}{L_p}), \quad \mbox{if } i \mbox{ even},\\
			\cos(\frac{\lfloor j/2 \rfloor \pi \bm{x}}{L_p}), \quad \mbox{if } i \mbox{ odd}.
		\end{array}\right.
	\end{aligned}
\end{equation*}
Here $\{\{\omega_{i,j}\}_{j=1}^N\}_{i=1}^{d_z}$ are independent standard Gaussian random variables, and $L_c$ is a desired physical correlation length, $L_p=\max\{1,2L_c\},L=L_c/L_p$. In the case when $d_{\bm{x}}=2$, we use a separate kernel to represent the Gaussian process prior, namely,
\begin{equation*}
    k((x_1,x_2),(x_1',x_2')) = \tilde{k}(x_1,x_1')\tilde{k}(x_2,x_2'),
\end{equation*}
where $\tilde{k}(\cdot,\cdot)$ is the one-dimensional kernel used in the $d_{\bm{x}}=1$ case. 

The latent dimension is set to 20, and the hyper-parameter $\beta$ is set to 0.01(for extrapolation tasks, $\beta$ is set to 0.1). During training, we utilize the Adam optimizer with an initial learning rate of 0.001 for 10000 steps, which decays at a rate of 0.7 every 1000 steps.  It is noteworthy that during the first 5,000 iterations, we exclusively optimize the predictive mean; in the subsequent 5,000 iterations, we begin optimizing the predictive standard deviation while simultaneously fine-tuning the predictive mean.

Additionally, we compare the performance of the LVM-GP method with the B-PINN-HMC \cite{yang2021b} and deep ensembles methods \cite{lakshminarayanan2017simple,fort2019deep}.
In the B-PINN-HMC method, the neural network architecture consists of a two hidden layer with 50 neurons per layer. The prior distribution for the weights and biases in the neural networks is set as an independent standard Gaussian distribution for each component. In HMC, the mass matrix is set to the identity matrix, $\bm{M} = \bm{I}$, the leapfrog step is set to $L = 50$, the time step is $\delta_t = 0.1$, the burn-in steps are set to 2000, and the total number of samples is 5000. For deep ensembles method, we use networks of the same size as HMC and set the decay rate to 4e-6 and run the DNN approximation code for 20 times to get the ensembles result, then use these ensembles to provide the corresponding estimations.

\subsection{PDE forward problem}
	\subsubsection{1D Poisson equation}\label{sec:1D_poisson}
We consider the following linear Poisson equation:
\begin{equation}
	\lambda \partial ^2_x u=f,\quad x\in[-0.7, 0.7],
	\label{1D_poisson}
\end{equation}
where $\lambda=0.01.$ The exact solution is given by $u(x)=\sin^3(6x)$, and the force term $f$ can be calculated via the equation. Here we assume the exact expression of $f$ is unknown, but we have 32 sensors of $f$, which is equidistantly distributed in $x\in[-0.7,0.7]$. In addition, we put two sensors at $x=-0.7$ and $x=0.7$ to provide the Dirichlet boundary condition for $u$. We assume that all measurements from the sensors are noisy, and we consider the following two different cases: (1) $\epsilon_f\sim \mc{N}(0, 0.01^2), \epsilon_b\sim \mc{N}(0, 0.01^2)$, and (2) $\epsilon_f \sim \mc{N}(0, 0.1^2), \epsilon_b\sim \mc{N}(0, 0.1^2)$.

\begin{figure}
	\begin{center}
		\begin{overpic}[width=0.25\textwidth, trim=0 0 0 0, clip=True]{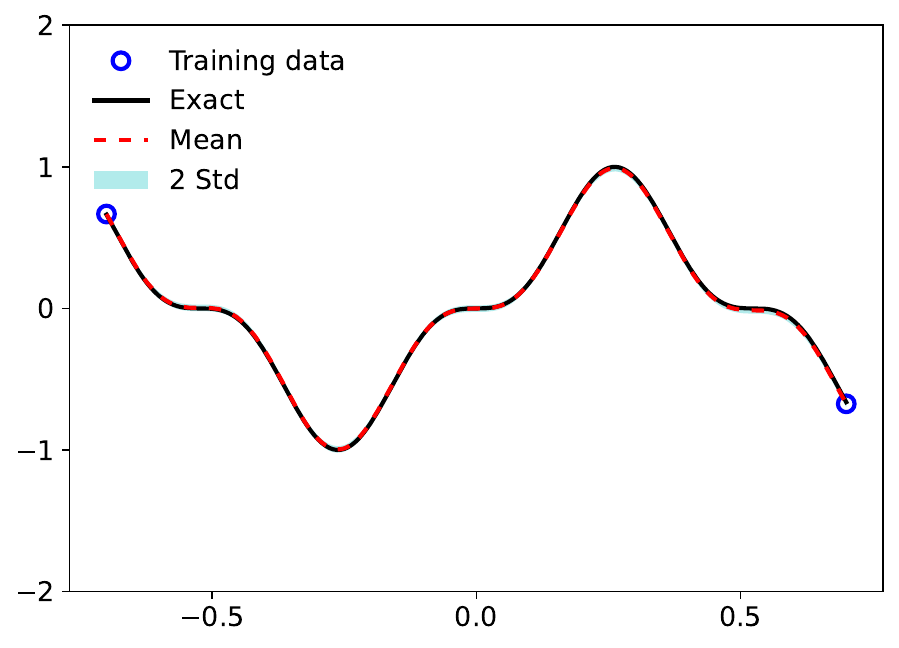}
			\put(35,72.5){{LVM-GP}}
			\put(-5, 35) {$u$}
		\end{overpic}
		\begin{overpic}[width=0.25\textwidth, trim=0 0 0 0, clip=True]{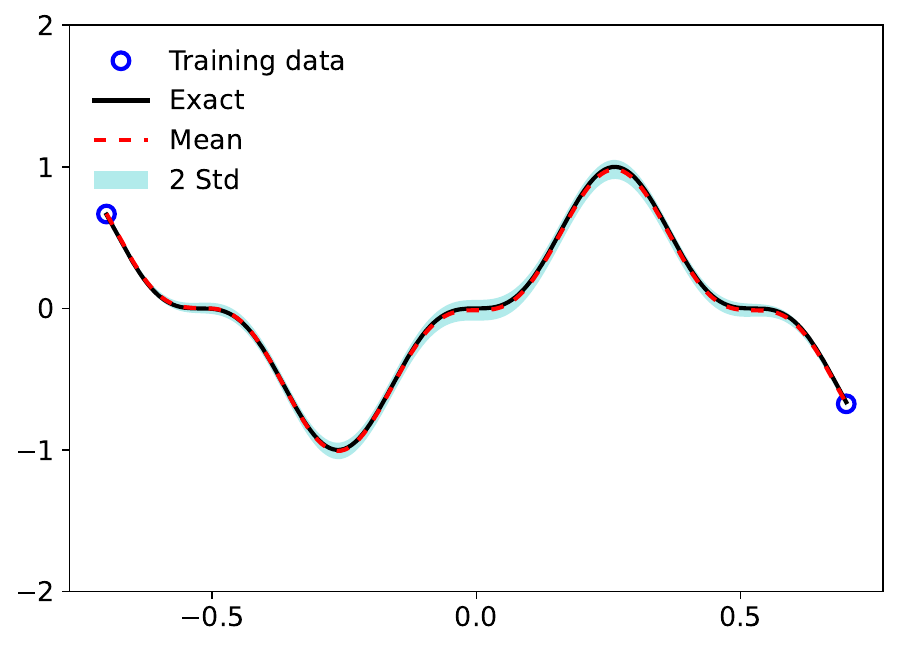}
			\put(25,72.5){{B-PINN-HMC}}
		\end{overpic}
		\begin{overpic}[width=0.25\textwidth, trim=0 0 0 0, clip=True]{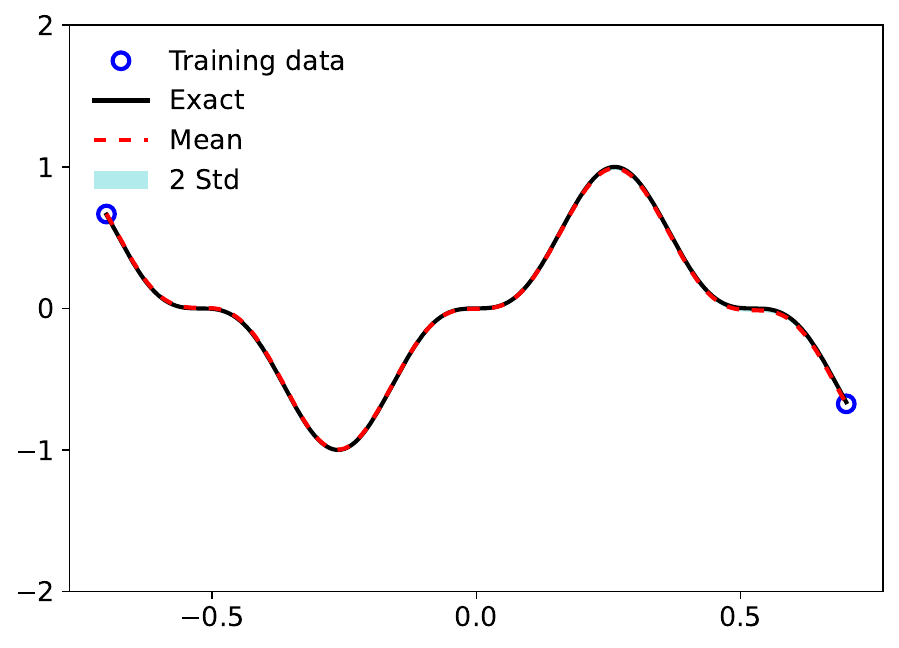}
			\put(20,72.5){{Deep Ensemble}}
		\end{overpic}

		\begin{overpic}[width=0.25\textwidth, trim=0 0 0 0, clip=True]{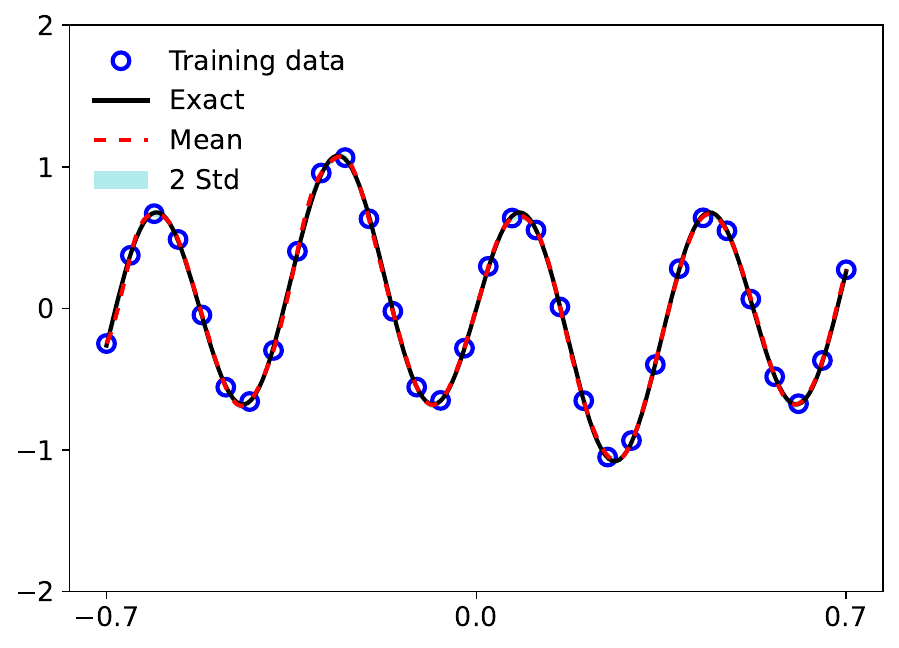}
			\put(-5, 35) {$f$}
		\end{overpic}
		\begin{overpic}[width=0.25\textwidth, trim=0 0 0 0, clip=True]{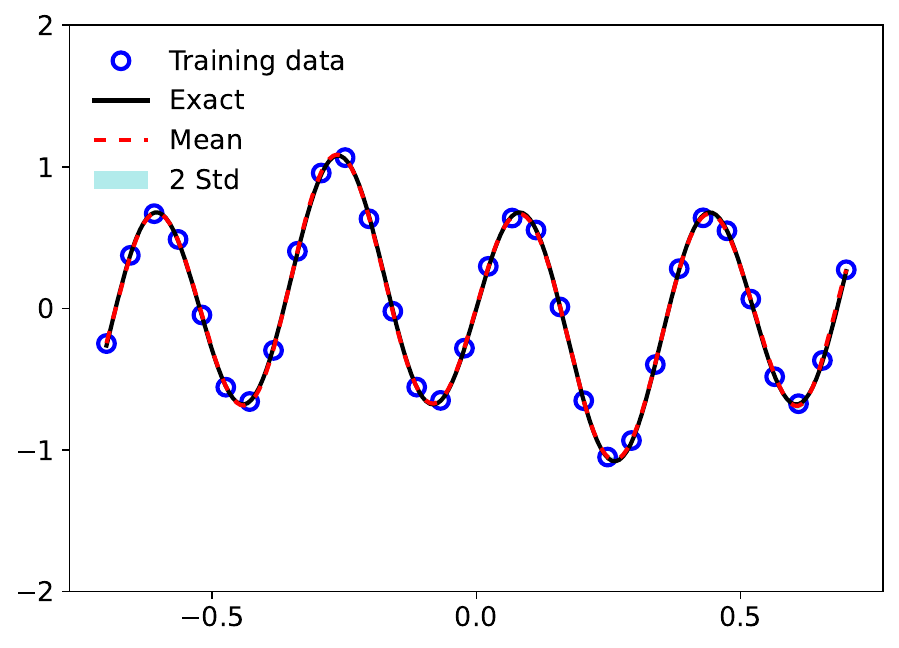}
		\end{overpic}
		\begin{overpic}[width=0.25\textwidth, trim=0 0 0 0, clip=True]{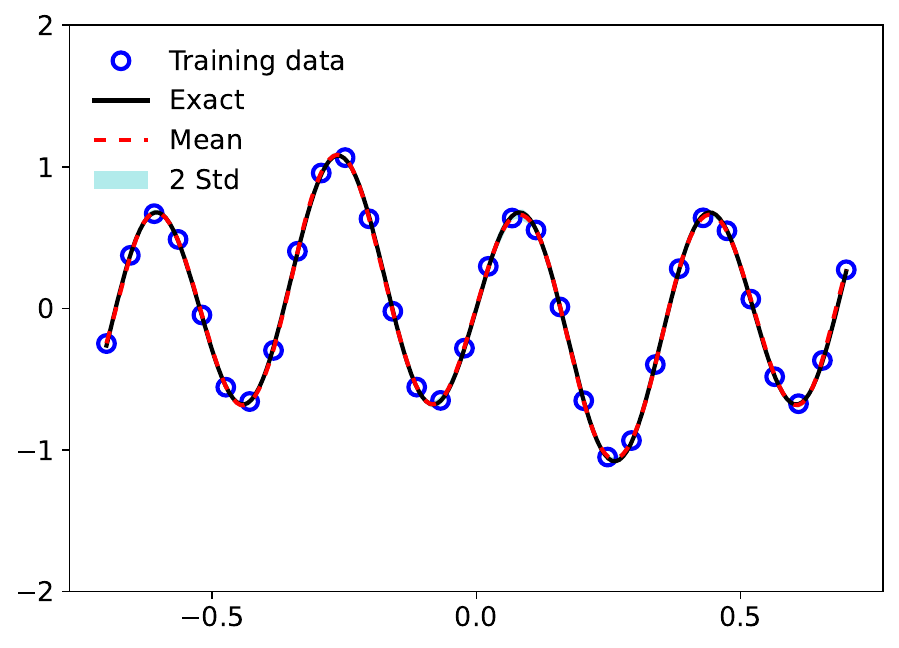}
		\end{overpic}
		\subcaption{}

		\begin{overpic}[width=0.25\textwidth, trim=0 0 0 0, clip=True]{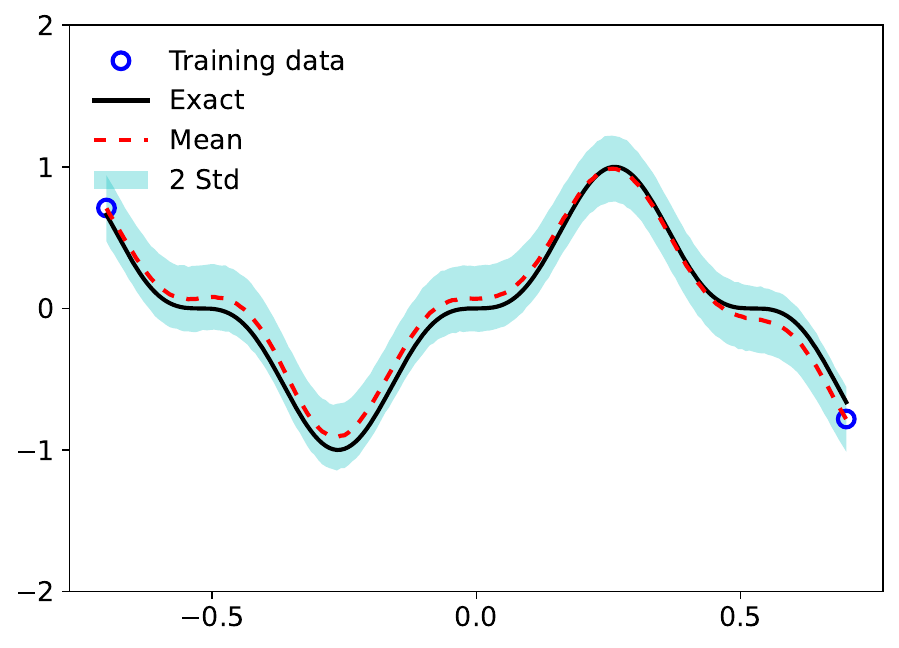}
			\put(-5, 35) {$u$}
		\end{overpic}
		\begin{overpic}[width=0.25\textwidth, trim=0 0 0 0, clip=True]{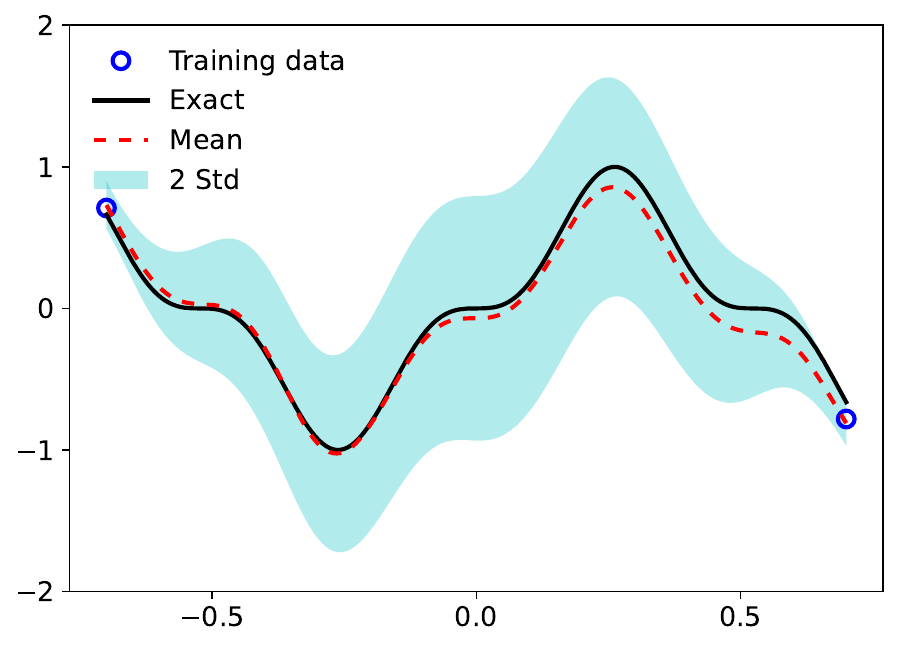}
		\end{overpic}
		\begin{overpic}[width=0.25\textwidth, trim=0 0 0 0, clip=True]{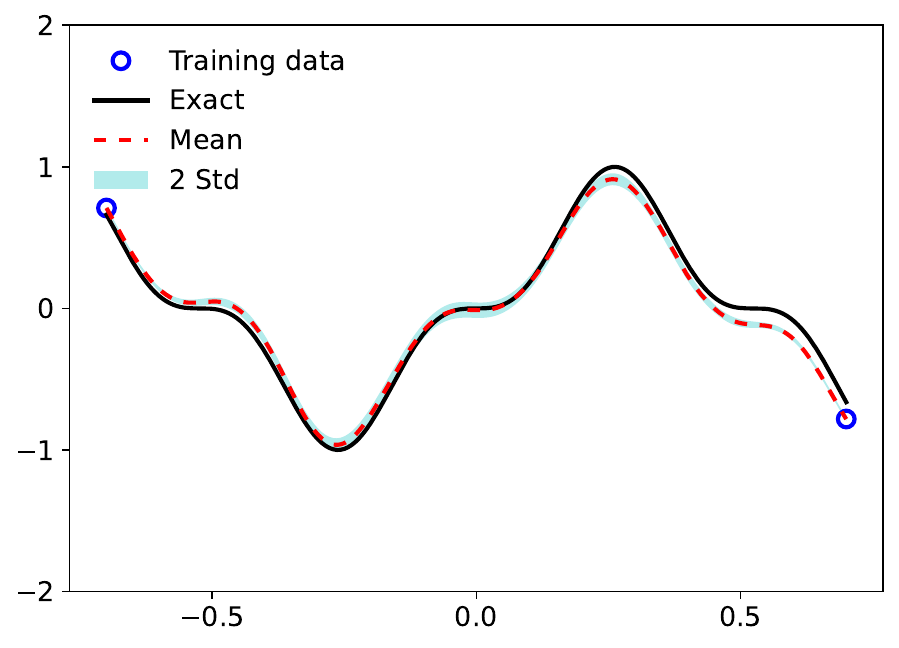}
		\end{overpic}

		\begin{overpic}[width=0.25\textwidth, trim=0 0 0 0, clip=True]{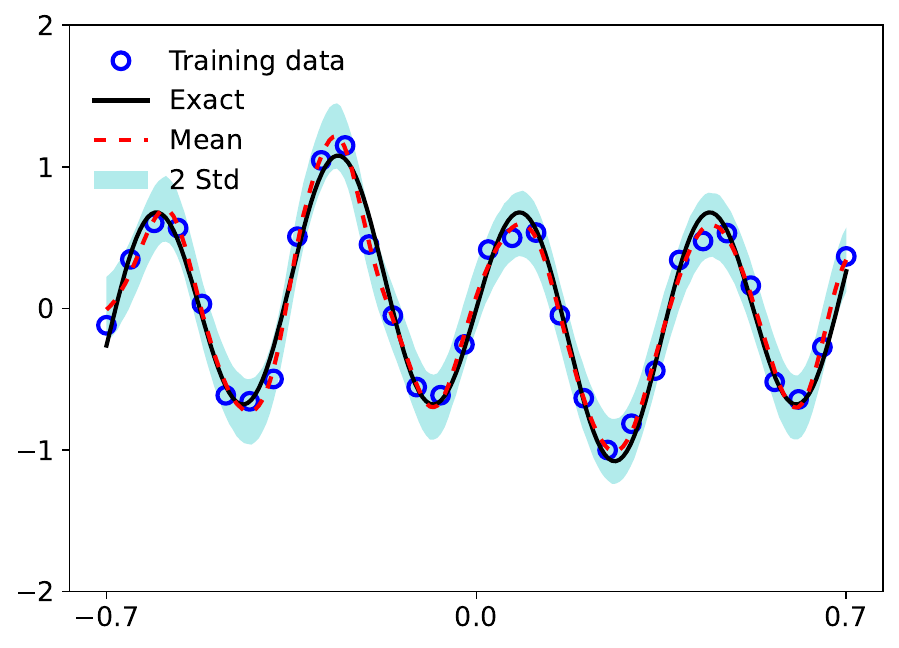}
			\put(-5, 35) {$f$}
		\end{overpic}
		\begin{overpic}[width=0.25\textwidth, trim=0 0 0 0, clip=True]{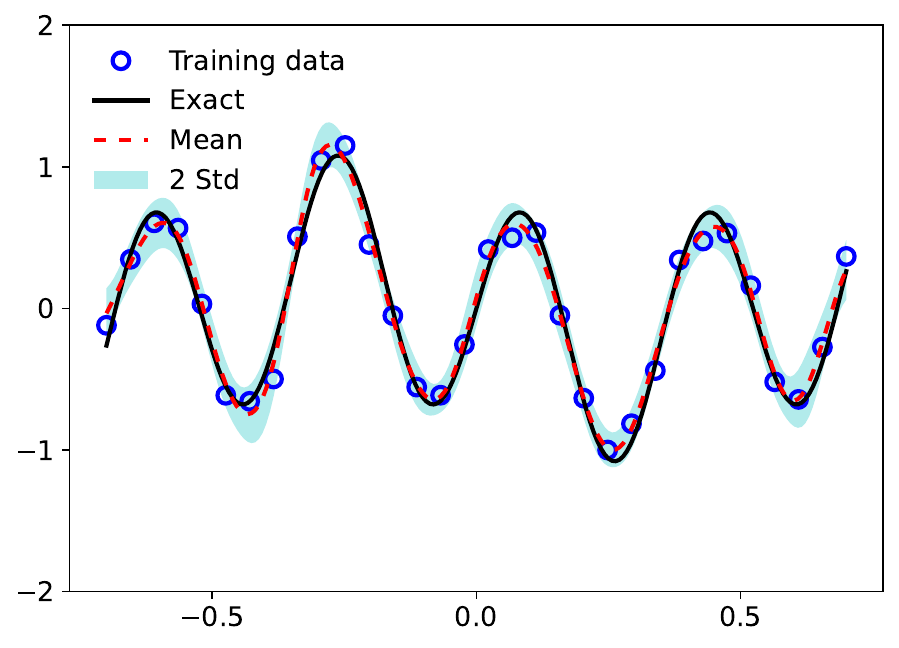}
		\end{overpic}
		\begin{overpic}[width=0.25\textwidth, trim=0 0 0 0, clip=True]{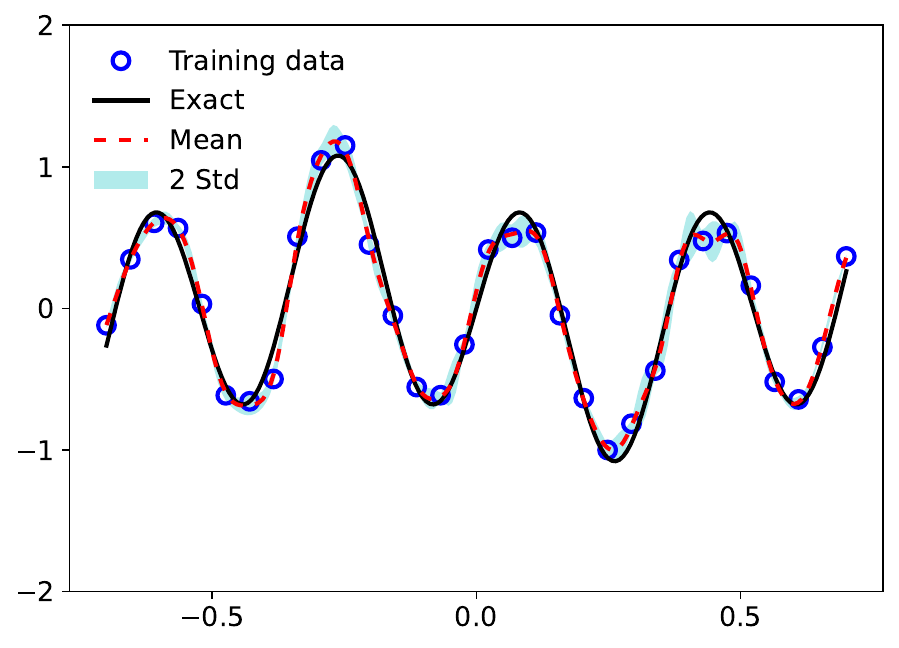}
		\end{overpic}
		\subcaption{}
	\end{center}
	\caption{1D Poisson equation. Predicted $u$ and $f$ from different methods with two data noise scales. (a) $\epsilon_f\sim \mc{N}(0, 0.01^2), \epsilon_b\sim \mc{N}(0, 0.01^2)$. (b) $\epsilon_f \sim \mc{N}(0, 0.1^2), \epsilon_b\sim \mc{N}(0, 0.1^2)$.  }
	\label{1D_poisson_unknown_noise}
\end{figure}

We evaluate the model performance under unknown data observation noise, where the term $\sigma_f, \sigma_u$ in (\ref{decoder_u},\ref{decoder_f_b}) are implemented as learnable scalars initialized at twice the empirical noise level(e.g., $\sigma_u = 0.02$ when $\epsilon \sim \mathcal{N}(0, 0.01^{2})$). The results are shown in Figure \ref{1D_poisson_unknown_noise}. One can clearly observe that LVM-GP, B-PINN-HMC and deep ensembles perform well when noise level is relatively low. However, as the noise level increases, deep ensembles fail to provide a reliable uncertainty quantification for the prediction of $f$. In comparison, the LVM-GP produces a better-predicted mean and tighter uncertainty estimation than B-PINN-HMC.

\begin{figure}[!h]
    \centering
    \includegraphics[width=0.5\linewidth]{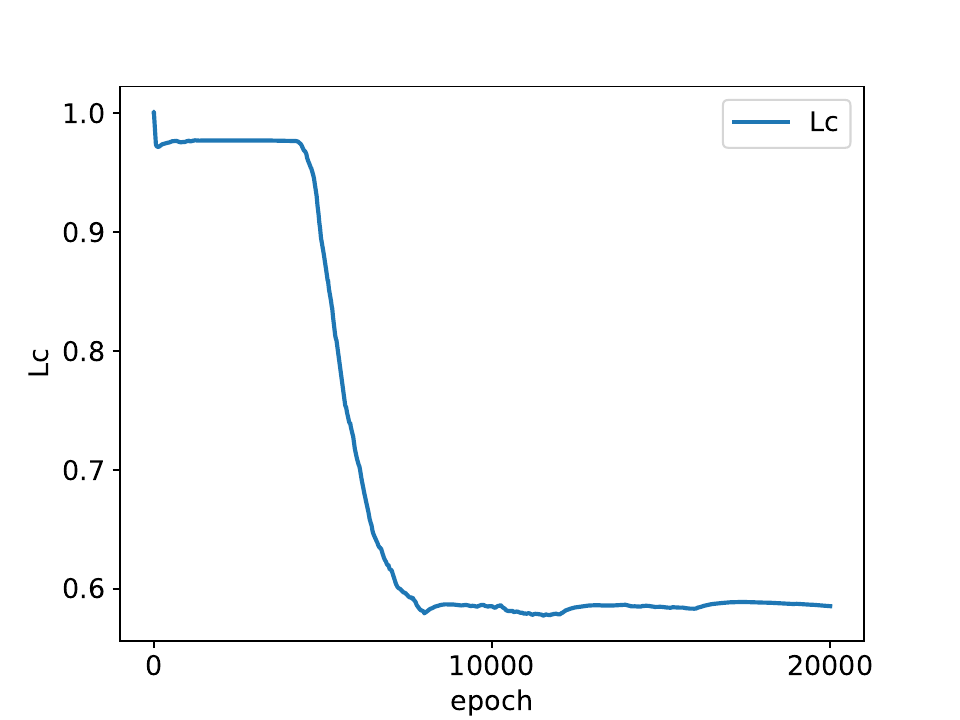}
    \caption{1D Poisson equation. Training dynamics of the hyperparameter in the Gaussian process prior. }
    \label{fig:learnable_curve_lc}
\end{figure}

\begin{figure}[!h]
    \centering
    \includegraphics[width=0.4\linewidth]{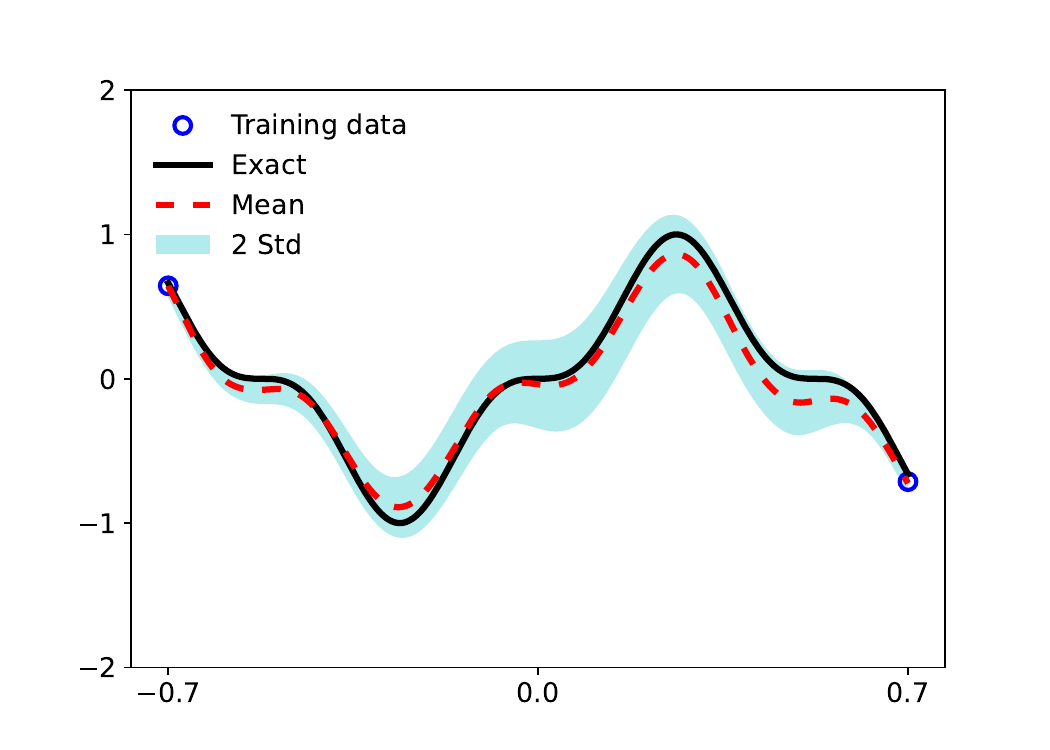}
    \includegraphics[width=0.4\linewidth]{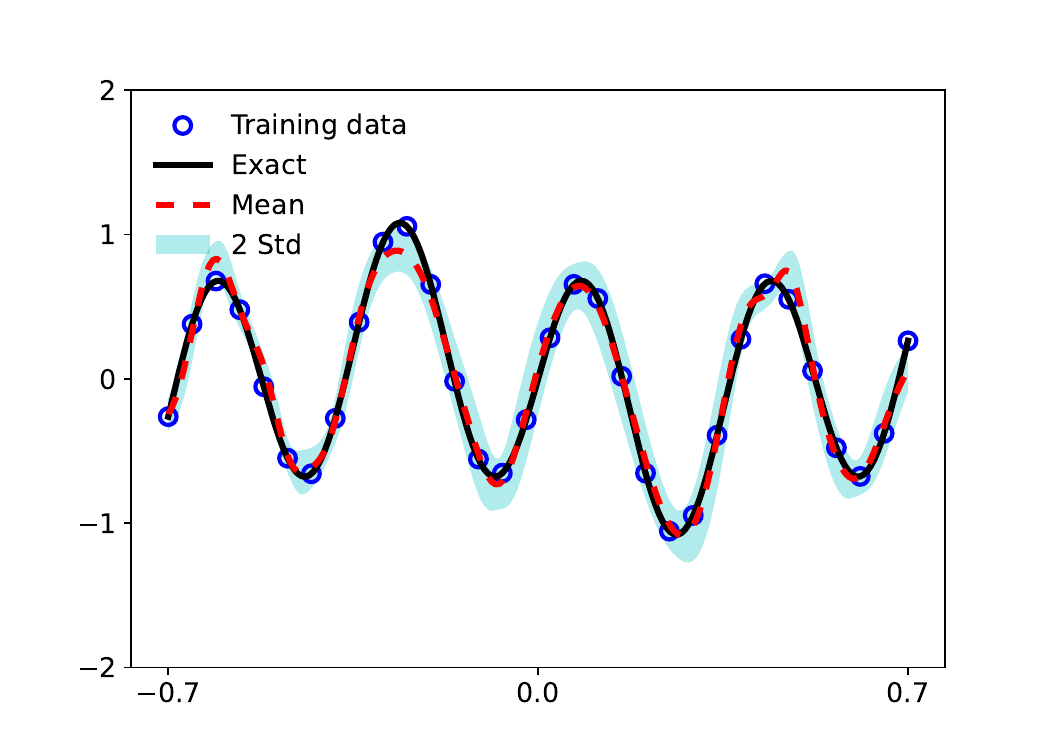}
    
    \caption{1D Poisson equation. Predicted $u$ and $f$ with learnable $L_c$, $\epsilon_u\sim\mc{N}(0,0.1^2),\epsilon_f\sim \mc{N}(0,0.1^2).$}
    \label{fig:1D_poisson_equation_with_learnable_lc}
\end{figure}

We also consider the LVM-GP framework with the revised regularization term \eqref{revised_regularization}, where the hyperparameter $L_c$ in the Gaussian process prior $\bm{z}_0$ is learned jointly with the parameters of the neural network. The numerical results are shown in Figure \ref{fig:learnable_curve_lc}  and Figure \ref{fig:1D_poisson_equation_with_learnable_lc}. Figure \ref{fig:learnable_curve_lc} illustrates the training dynamics of the kernel length scale $L_c$. Starting from an initial value close to 1.0, the length scale gradually decreases and stabilizes around 0.58 after about 8000 iterations, indicating that the model automatically adapts the correlation scale in the latent space as learning progresses. This adaptive behavior suggests that the model is able to refine the prior structure based on the observed data. Figure \ref{fig:1D_poisson_equation_with_learnable_lc} shows the predicted solution 
$u$ and source term $f$ for the 1D Poisson equation using a learnable kernel length scale $L_c$. The model accurately captures both the mean and uncertainty of the target functions, with predictions closely matching the exact solutions. Moreover, numerical results for the decoder with DeepONet-Type architecture are provided in the Appendix \ref{sec: DeepONet-Type forward}. 

\subsubsection{1D flow through porous media with boundary layer}
We now consider flows through a horizontal channel filled with a uniform porous medium governed by the following equation: 
\begin{equation*}
    -\frac{\nu_e}{\phi}\partial_x^2 u  + \frac{\nu u}{K}=g, \quad x\in[0,1],
\end{equation*}
where $g$ denotes the external force, $\nu_e$ is the effective viscosity, $\phi$ is the porosity, $\nu$  is the fluid viscosity, $K$ is the permeability. The coordinate $ x $ is orthogonal to the channel, and $ u(x) $ denotes the velocity along the channel. No-slip boundary conditions are enforced at both ends, i.e., $ u(0) = u(1) = 0 $. The analytical solution to this problem is
\begin{equation*}
    u=\frac{gK}{\nu} \left[1 - \frac{\cosh(r(x-H/2))}{\cosh(rH/2)}\right],
\end{equation*}
where $ r = \sqrt{\frac{\nu \phi}{\nu_e K}} $. For this case, we set $ \nu_e = \nu = 10^{-3} $, $ \phi = 0.4 $, and $ K = 10^{-3} $. The unknown external force is fixed at $ g = 1 $, and corresponding data are collected from sensors.

\begin{figure}[H]
	\begin{center}
		\begin{overpic}[width=0.25\textwidth, trim=0 0 0 0, clip=True]{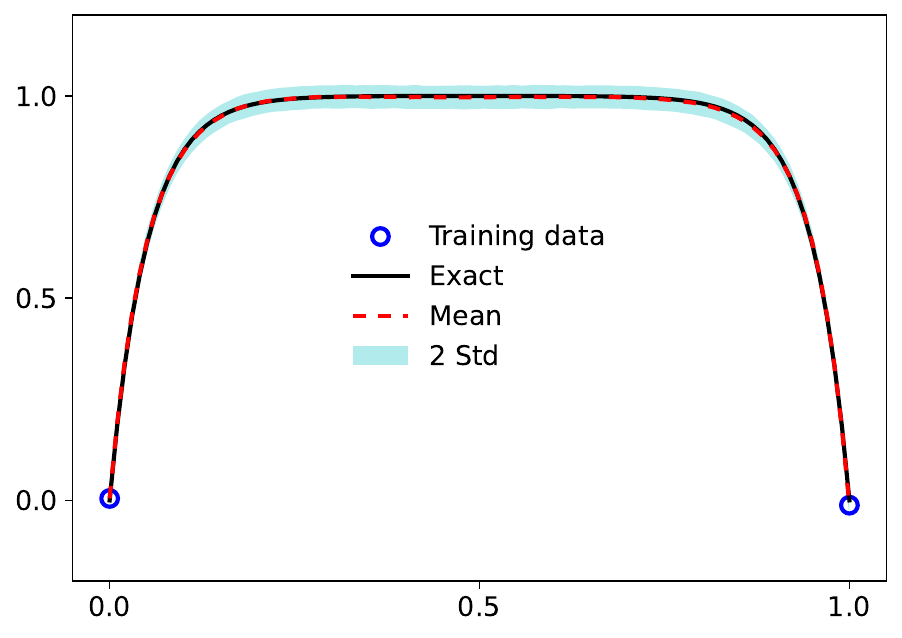}
			\put(35,72.5){{LVM-GP}}
			\put(-8, 35) {$u$}
		\end{overpic}
		\begin{overpic}[width=0.25\textwidth, trim=0 0 0 0, clip=True]{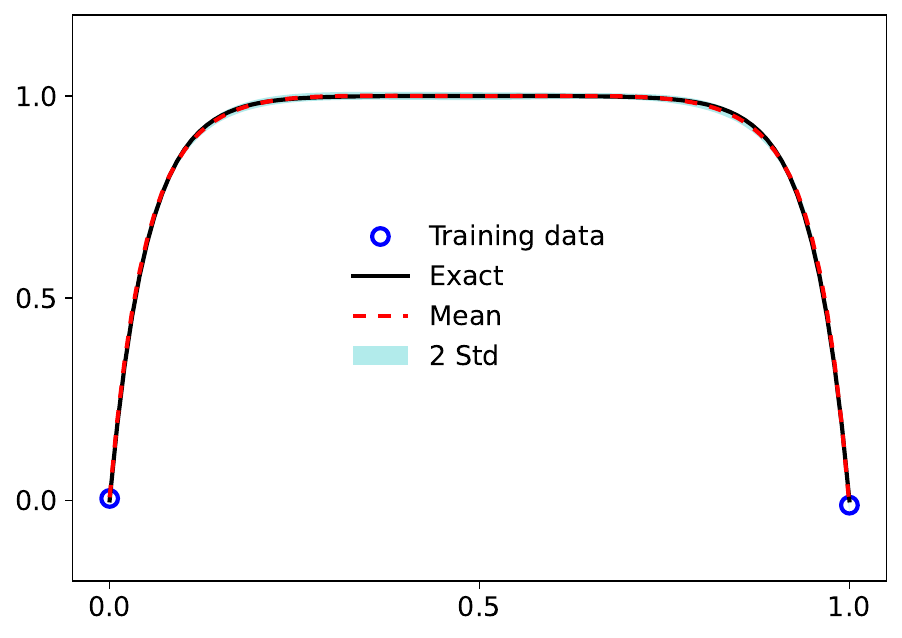}
			\put(25,72.5){{B-PINN-HMC}}
		\end{overpic}
		\begin{overpic}[width=0.25\textwidth, trim=0 0 0 0, clip=True]{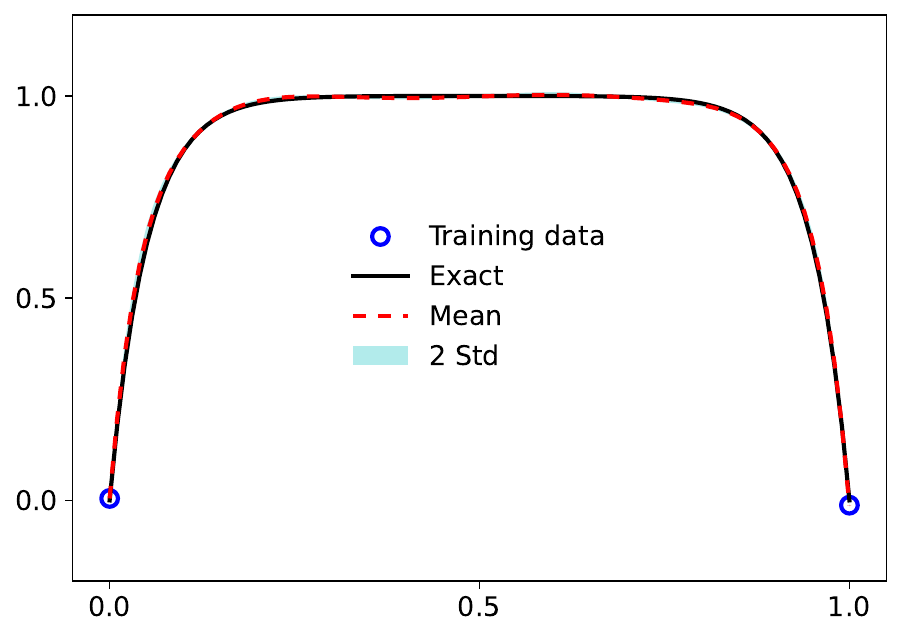}
			\put(20,72.5){{Deep Ensemble}}
		\end{overpic}
		\subcaption{}

		\begin{overpic}[width=0.25\textwidth, trim=0 0 0 0, clip=True]{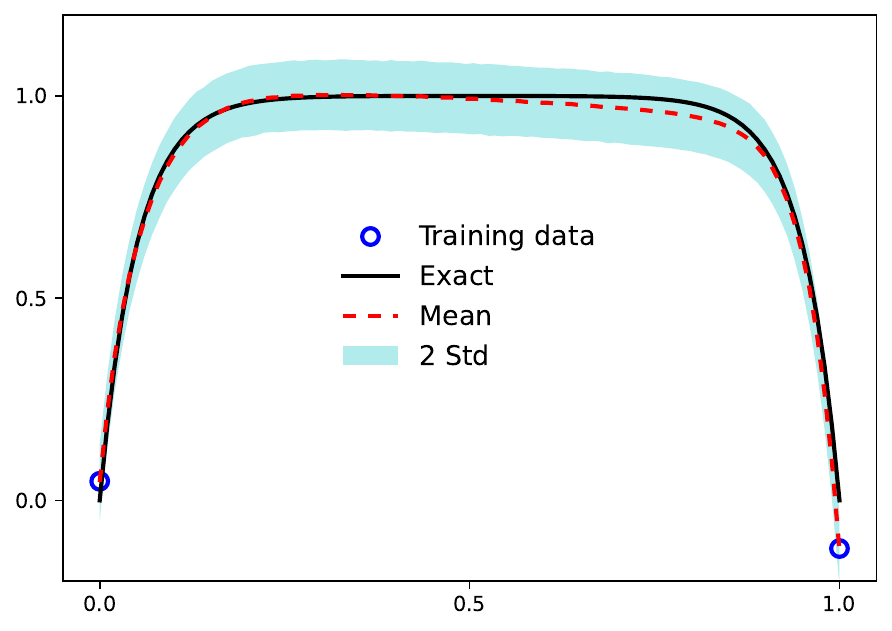}
			\put(-5, 35) {$u$}
		\end{overpic}
		\begin{overpic}[width=0.25\textwidth, trim=0 0 0 0, clip=True]{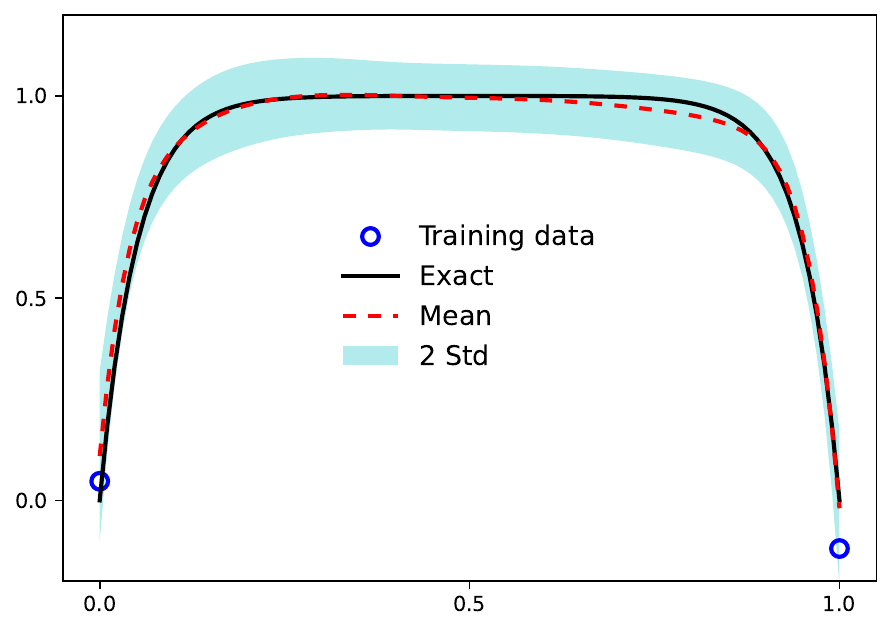}
		\end{overpic}
		\begin{overpic}[width=0.25\textwidth, trim=0 0 0 0, clip=True]{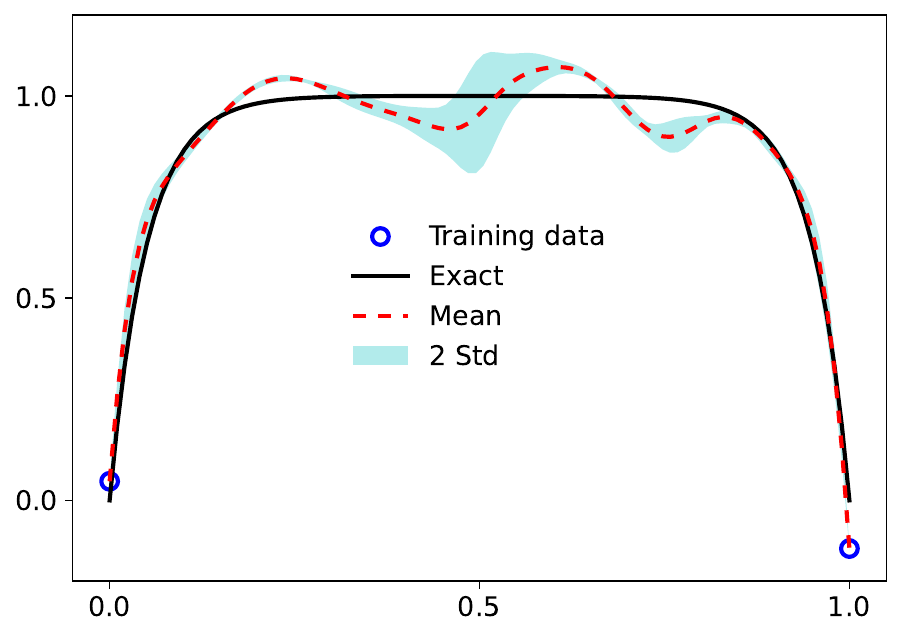}
		\end{overpic}
		\subcaption{}
	\end{center}
	\caption{1D flow through porous media with boundary layer. Predicted $u$ from different methods with two data noise scales. (a) $\epsilon_g\sim \mc{N}(0, 0.01^2), \epsilon_b\sim \mc{N}(0, 0.01^2)$. (b) $\epsilon_g \sim \mc{N}(0, 0.1^2), \epsilon_b\sim \mc{N}(0, 0.1^2)$.  }
	\label{1D_flow_noise}
\end{figure}

Here we assume that we have 16 sensors for $g$  equidistantly placed in $x \in [0, 1]$. In addition, two sensors for $u$ are placed at $x = 0$ and $1$ for Dirichlet boundary conditions. Two different scales of Gaussian noise in the measurements are considered, i.e., (1) $g \sim \mathcal{N}(0, 0.01^2), b \sim \mathcal{N}(0, 0.01^2)$, and (2) $g \sim \mathcal{N}(0, 0.1^2), b \sim \mathcal{N}(0, 0.1^2)$. The results of different methods are presented in Figure \ref{1D_flow_noise}. The LVM-GP method achieves results comparable to HMC, while deep ensembles yield inaccurate estimates and unreliable uncertainty quantification under high-noise conditions. In this case study, both the encoder and decoder are implemented as 3-layer fully-connected neural networks with 20 neurons per hidden layer. Notably, both the  HMC and deep ensembles adopt network architectures identical to LVM-GP. The training process is divided into two stages: during the first 2000 steps, only the predictive mean is optimized; in the subsequent 8000 steps, the predictive standard deviation is optimized while the predictive mean continues to be fine-tuned.

\subsection{PDE inverse problem}
\subsubsection{1D nonlinear Poisson equation}
\begin{figure}[H]
	\begin{center}
		\begin{overpic}[width=0.25\textwidth, trim=0 0 0 0, clip=True]{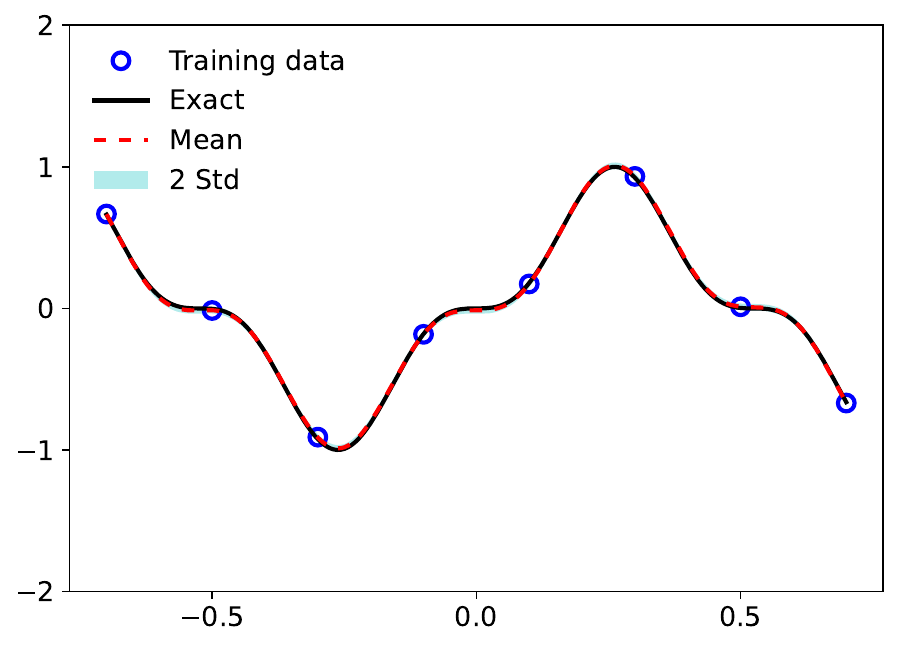}
			\put(35,72.5){{LVM-GP}}
			\put(-5, 35) {$u$}
		\end{overpic}
		\begin{overpic}[width=0.25\textwidth, trim=0 0 0 0, clip=True]{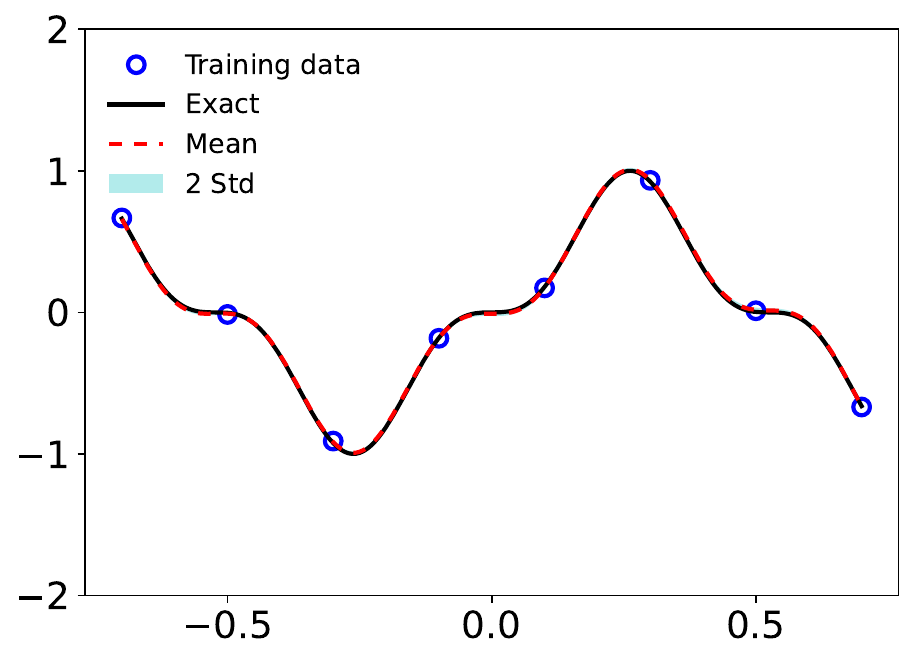}
			\put(25,72.5){{B-PINN-HMC}}
		\end{overpic}
		\begin{overpic}[width=0.25\textwidth, trim=0 0 0 0, clip=True]{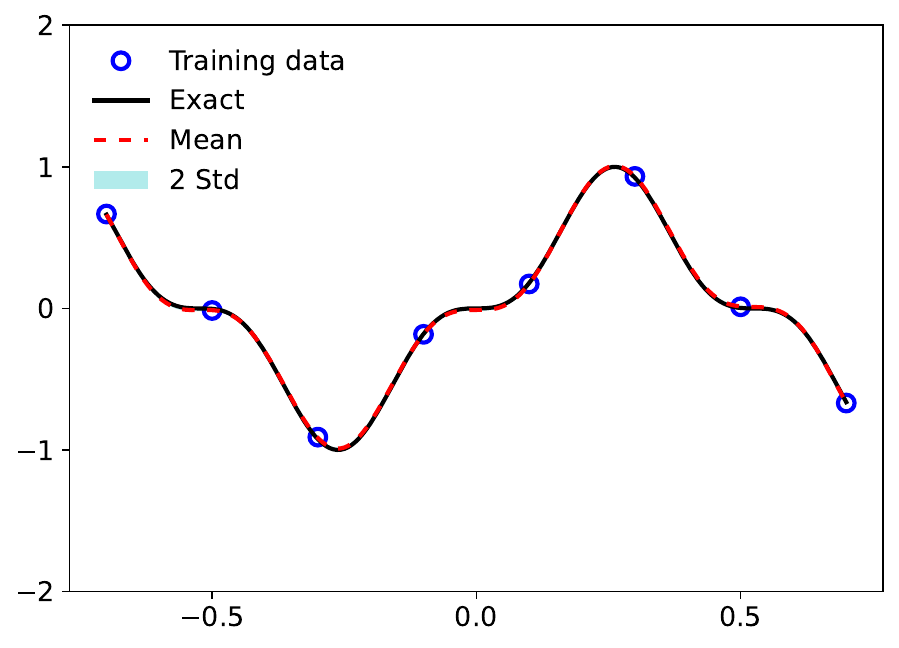}
			\put(20,72.5){{Deep Ensemble}}
		\end{overpic}

		\begin{overpic}[width=0.25\textwidth, trim=0 0 0 0, clip=True]{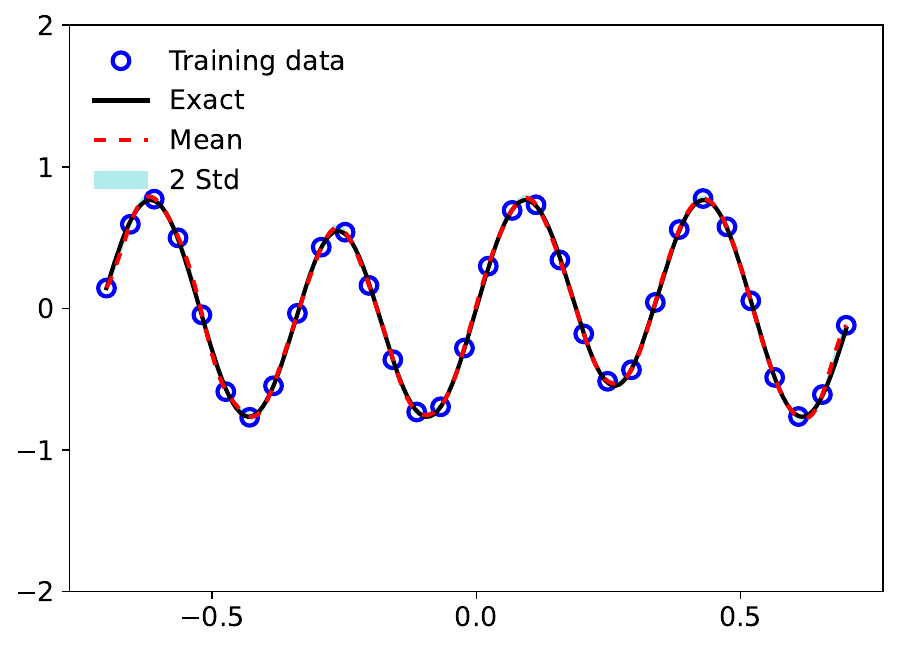}
			\put(-5, 35) {$f$}
		\end{overpic}
		\begin{overpic}[width=0.25\textwidth, trim=0 0 0 0, clip=True]{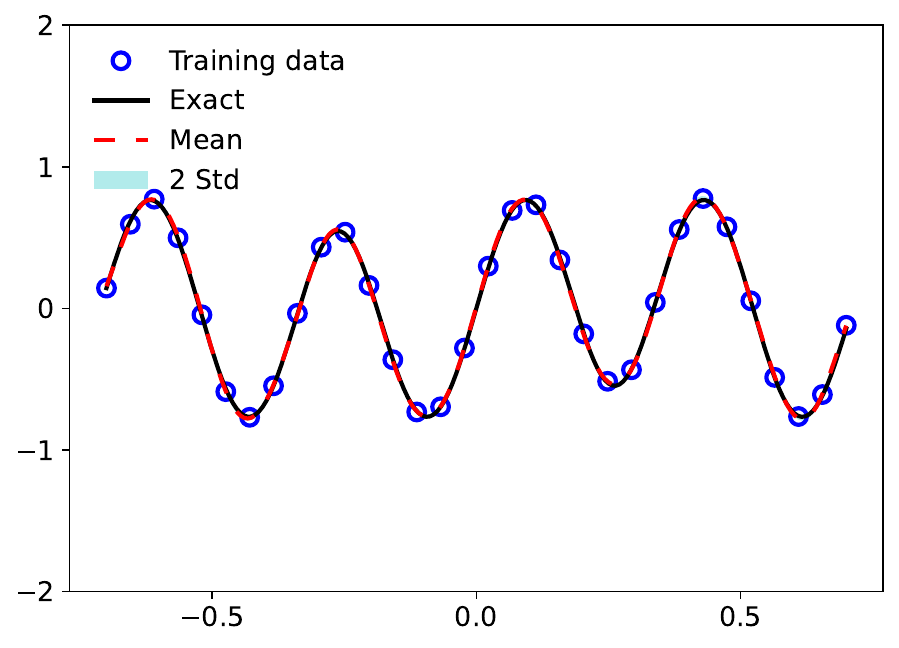}
		\end{overpic}
		\begin{overpic}[width=0.25\textwidth, trim=0 0 0 0, clip=True]{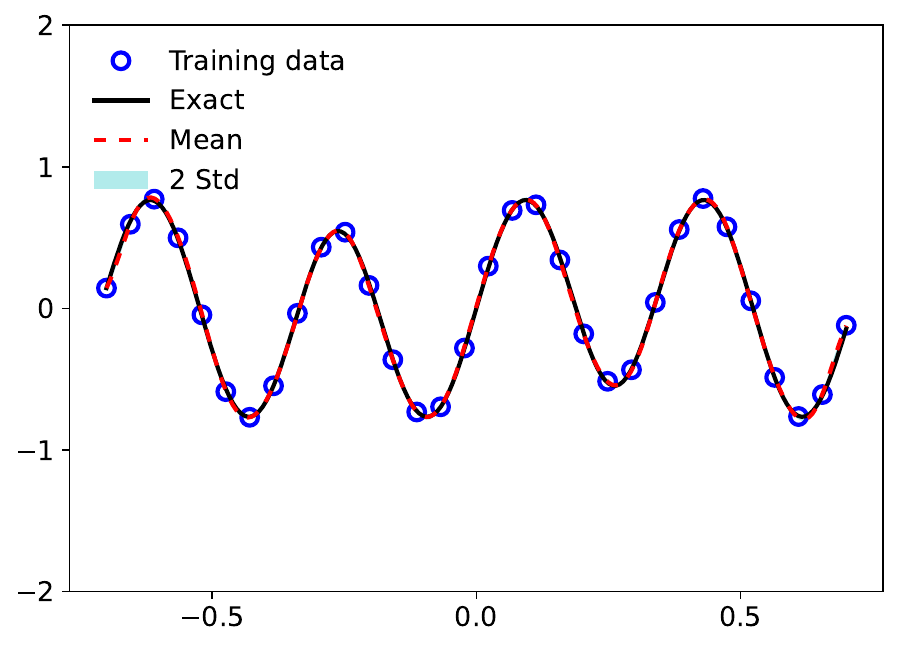}
		\end{overpic}
		\subcaption{}

		\begin{overpic}[width=0.25\textwidth, trim=0 0 0 0, clip=True]{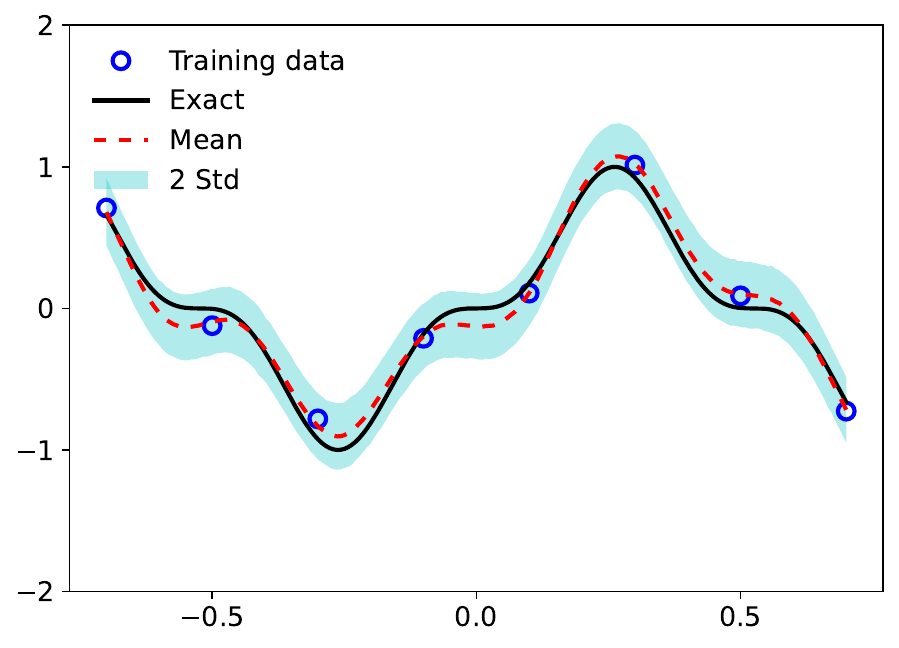}
			\put(-5, 35) {$u$}
		\end{overpic}
		\begin{overpic}[width=0.25\textwidth, trim=0 0 0 0, clip=True]{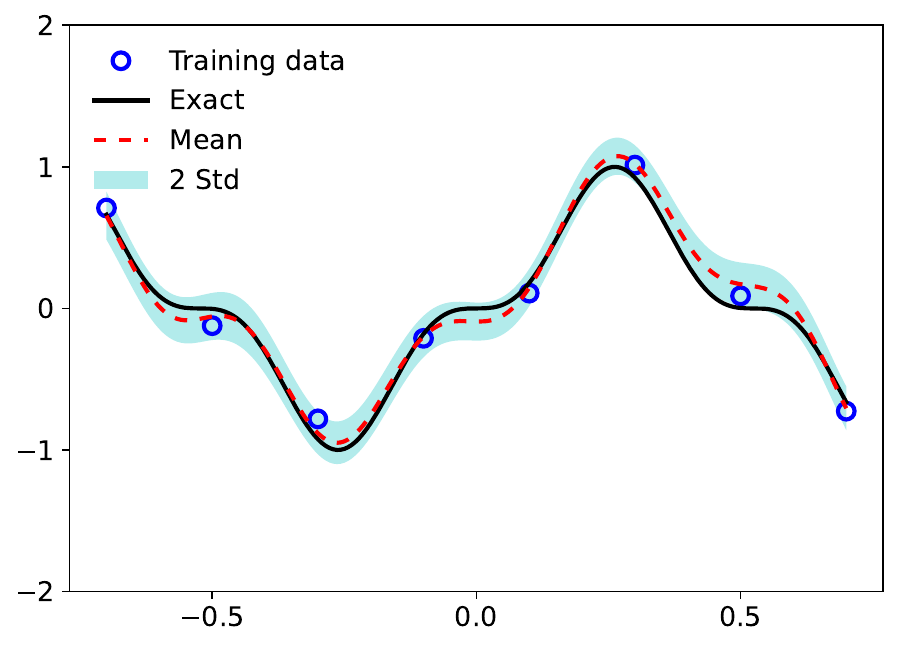}
		\end{overpic}
		\begin{overpic}[width=0.25\textwidth, trim=0 0 0 0, clip=True]{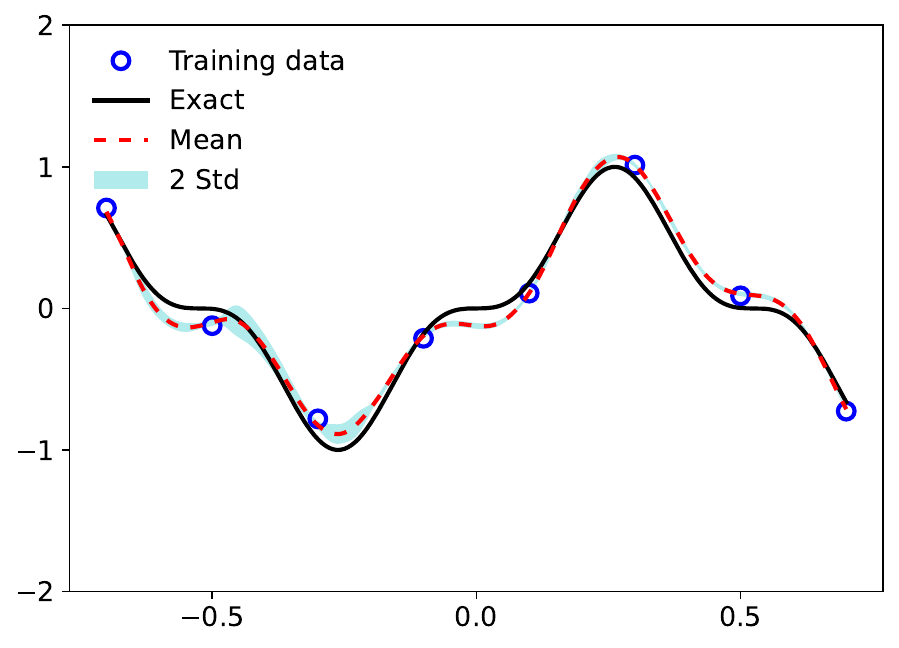}
		\end{overpic}

		\begin{overpic}[width=0.25\textwidth, trim=0 0 0 0, clip=True]{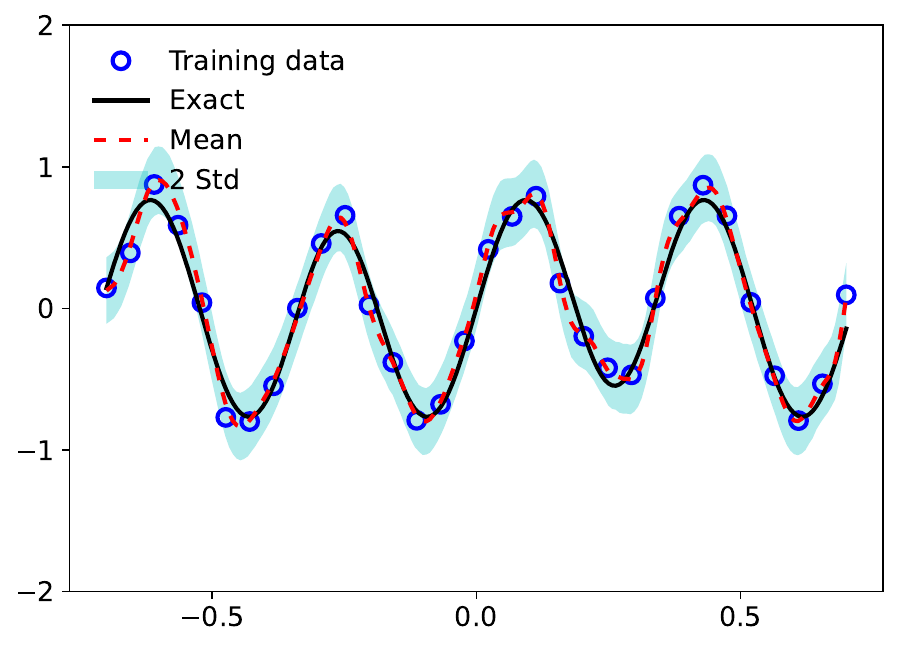}
			\put(-5, 35) {$f$}
		\end{overpic}
		\begin{overpic}[width=0.25\textwidth, trim=0 0 0 0, clip=True]{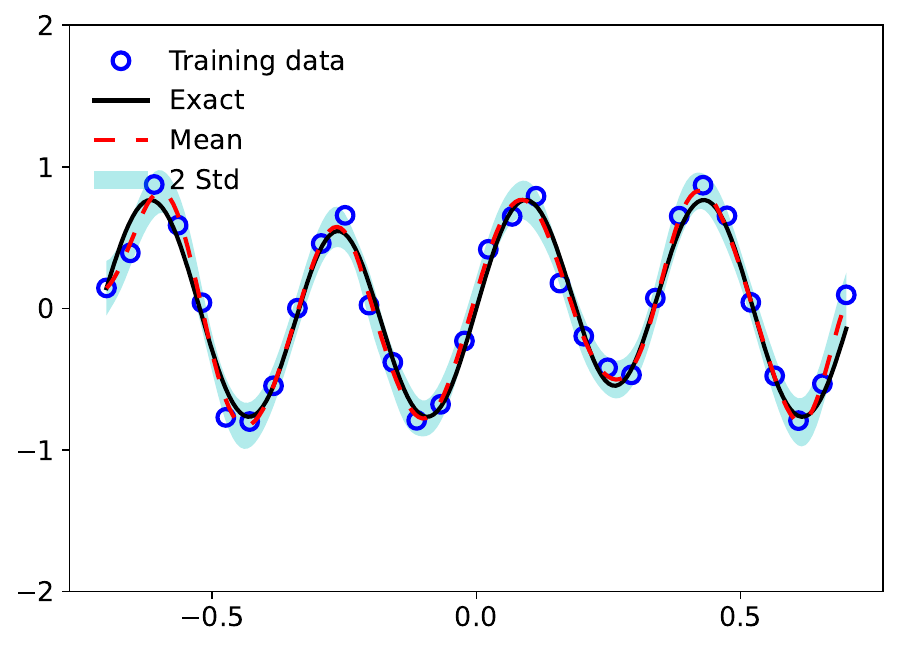}
		\end{overpic}
		\begin{overpic}[width=0.25\textwidth, trim=0 0 0 0, clip=True]{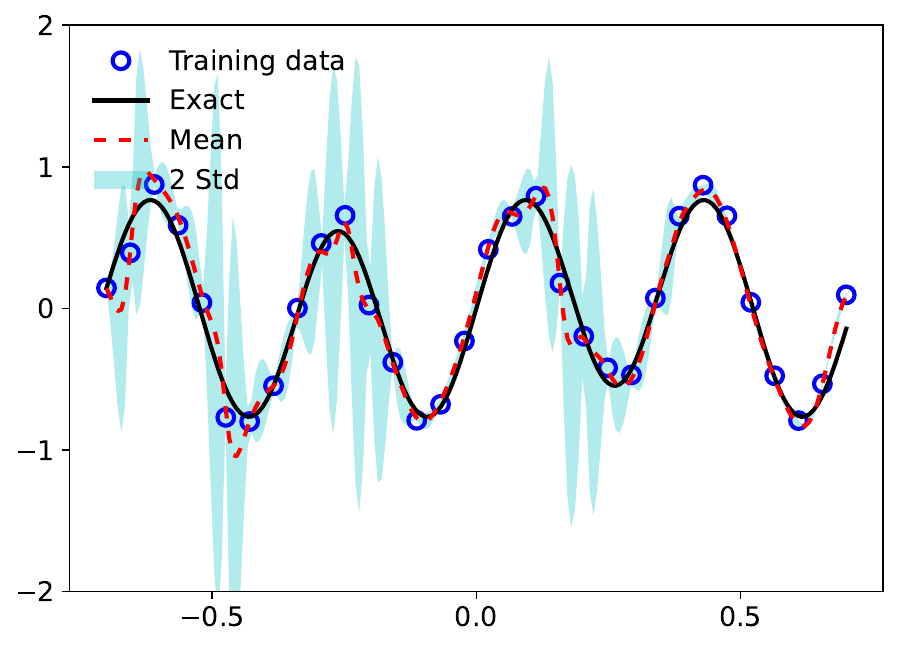}
		\end{overpic}
		\subcaption{}
	\end{center}
	\caption{1D nonlinear Poisson equation. Predicted $u$ and $f$ from different methods with two data noise scales. (a) $\epsilon_f\sim \mc{N}(0, 0.01^2), \epsilon_b\sim \mc{N}(0, 0.01^2)$. (b) $\epsilon_f \sim \mc{N}(0, 0.1^2), \epsilon_b\sim \mc{N}(0, 0.1^2)$.}
	\label{1D_nonlinear_unknown}
\end{figure}

We consider the following 1D nonlinear equation
\begin{equation}
	k \partial_x^2 u + \lambda \tanh(u) = f, \quad x\in[-0.7, 0.7],
\end{equation}
and the exact solution $u(x)=\sin^3(6x)$. In addition, $k=0.01$ and $\lambda$ is the unknown reaction rate. The objective here is to identify $\lambda$ based on partial measurements of $f$ and $u$. The exact value of $\lambda$ is 0.7 and we assume that we have 32 sensors for $f$, which are equidistantly placed in $x\in [-0.7, 0.7]$. In addition, two sensors for $u$ are placed at $x=-0.7$ and $0.7$ to provide  Dirichlet boundary conditions. Apart from the boundary conditions, another 6 sensors for $u$ are placed in the interior of the domain to help identify $\lambda$.
Similar to \ref{1D_poisson}, we evaluate the performance of the LVM-GP method using predetermined decoder noises that correspond to the unknow noise level. In contrast to the forward problem formulation, the predictive network $\Lambda$ for parameter $\lambda$ estimation in this implementation adopts a simplified architecture comprising a single-neuron hidden layer with hyperbolic tangent  activation function. The results are depicted in Figure \ref{1D_nonlinear_unknown}. The predicted results of LVM-GP and B-PINN-HMC exhibit a remarkable agreement with the ground truth. Nevertheless, it is evident that the results of deep ensembles method display significant oscillations when the observation noise is relatively high. The predicted values for $\lambda$ from different methods are displayed in Table \ref{table:1D_nonlinear_poisson_kpred}.

\begin{table}[H]
	\centering
	{\footnotesize
		\begin{tabular}{cc|ccc}
			\hline \hline
			\multicolumn{2}{c|}{ Noise scale} & { LVM-GP} & { B-PINN-HMC} & { Deep Ensemble}                                                              \\
			\hline
			\multirow{2}{*}{0.01}             & Mean    & 0.6976        & 0.6967                               & 0.6966                                \\
			                                  & Std     & {\footnotesize $9.816 \times 10^{-3}$}             & {\footnotesize $4.225 \times 10^{-3}$} & {\footnotesize $2.493 \times 10^{-4}$} \\
			\multirow{2}{*}{0.1}              & Mean    & 0.6965        & 0.6787                                & 0.6959                                \\
			                                  & Std     & {\footnotesize $6.954 \times 10^{-2}$}            & {\footnotesize $4.166 \times 10^{-2}$} & {\footnotesize $3.691 \times 10^{-2}$} \\
			\hline \hline
		\end{tabular}
	}
	\caption{1D nonlinear Poisson equation: Predicted mean and standard deviation for the reaction rate $\lambda$ using different uncertainty-induced methods. $\lambda = 0.7$ is the exact solution.}
	\label{table:1D_nonlinear_poisson_kpred}
\end{table}

\begin{figure}[H]
	\begin{center}
		\begin{overpic}[width=0.45\textwidth, trim= 0 0 0 0,
				clip=True]{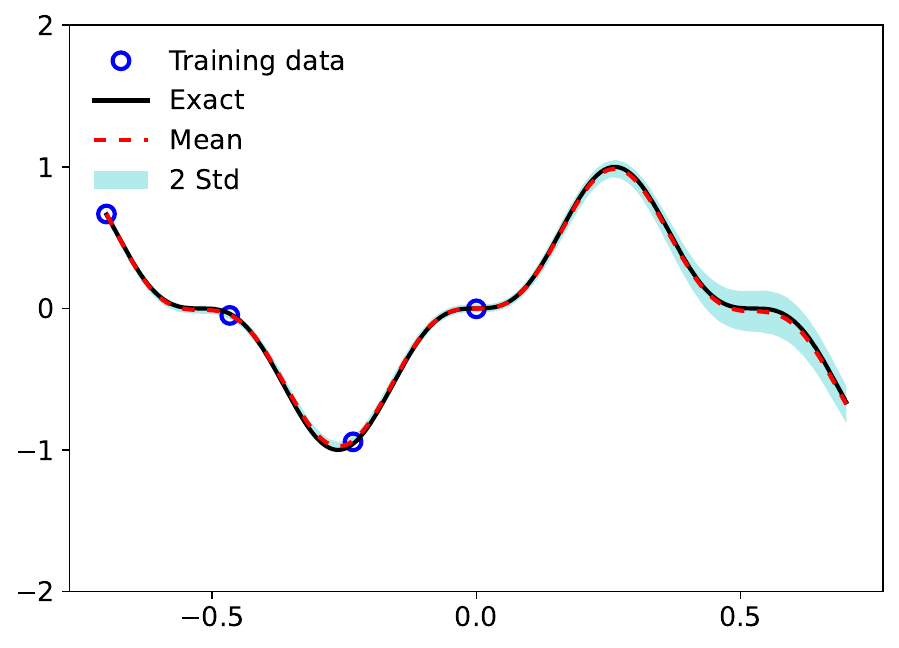}
			\put(50, 72.5){$u$}
		\end{overpic}
		\begin{overpic}[width=0.45\textwidth, trim= 0 0 0 0,
				clip=True]{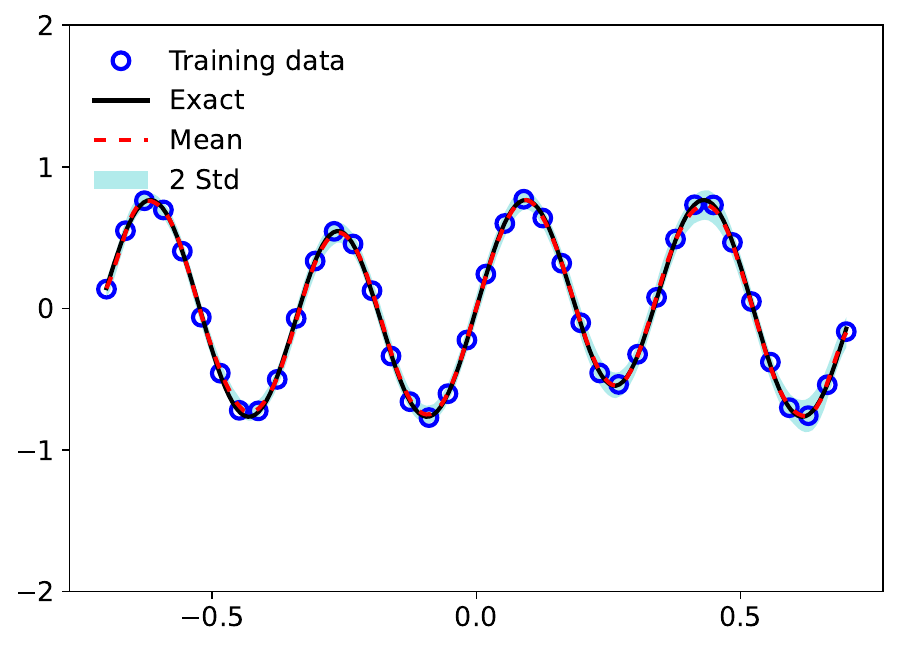}
			\put(50, 72.5){$f$}
		\end{overpic}
	\end{center}
	\caption{Extrapolation of LVM-GP: predicted $u$ and $f$ with a limited number of observations on $u$. $\epsilon_u\sim \mc{N}(0, 0.01^2), \epsilon_f \sim \mc{N}(0, 0.01^2)$.}
	\label{extrapolation}
\end{figure}

Furthermore, we evaluate the LVM-GP method on the inverse problem with a missing one-sided boundary condition of $u$. Specifically, we utilize only four measurements of $u$ at $x\leq 0$ to estimate the parameter $k$ and extrapolate $u$ at $x>0$ using the PDE constraint and 40 uniformly distributed measurements of $f$. The measurements are corrupted by Gaussian noise with a standard deviation of $0.01$. The results are depicted in Figure \ref{extrapolation}, which demonstrates the effectiveness of our method for the extrapolation of $u$ with the PDE constraint. In addition, the prediction mean for $k$ is $0.7040$ with a standard deviation of $1.361 \times 10^{-2}$, which is in good agreement with the exact value.
	
\subsubsection{2D nonlinear diffusion-reaction system}
We consider the following PDE here
\begin{equation}
	k (\partial _{x_1}^2 u + \partial _{x_2}^2 u) + \lambda u^2 = f, \quad x_1,x_2\in[-1, 1],
\end{equation}
where $k=0.01$ is the diffusion coefficient, $\lambda$ represents the reaction rate, which is a constant, and $f$ denotes the source term. Here we assume that the exact value for $\lambda$ is unknown, and we only have sensors for $u$ and $f$. We have a total of 100 and 484 uniformly distributed sensors for $u$ and $f$ respectively in the physical domain. Additionally, we have 25 equally distributed sensors for $u$ at each boundary to satisfy the Dirichlet boundary condition. All measurements are noisy, and two noise scales are considered: (1) $\epsilon_f \sim \mathcal{N}(0, 0.01^2)$, $\epsilon_u \sim \mathcal{N}(0, 0.01^2)$; (2) $\epsilon_f \sim \mathcal{N}(0, 0.1^2)$, $\epsilon_u \sim \mathcal{N}(0, 0.1^2)$. The exact solution for $u$ is given by $u(x_1, x_2) = \sin(\pi x_1) \sin(\pi x_2)$.

\begin{figure}[H]
	\begin{center}
		\begin{overpic}[height=30mm, keepaspectratio, trim=80 20 80 10, clip=False]{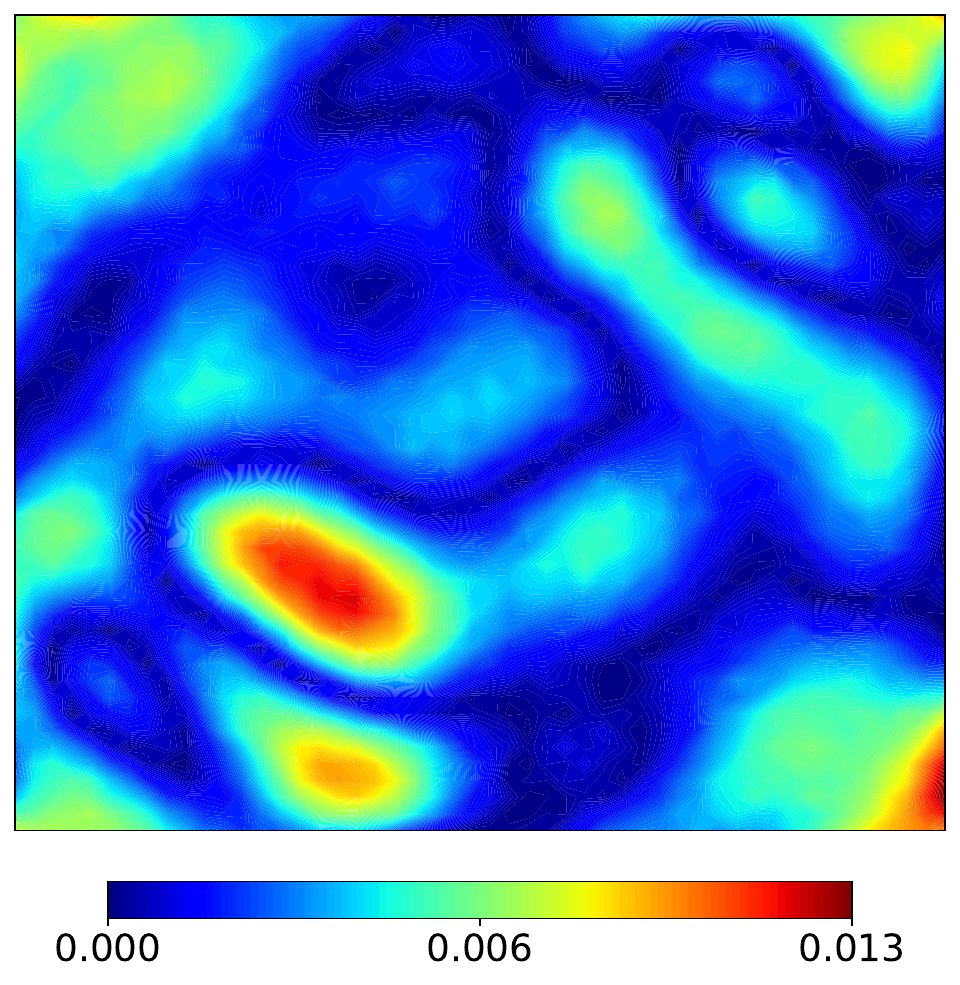}
			\put(32.5, 105) {$u$}
			\put(-55, 50) {errors}
		\end{overpic}
        \hspace{10mm} 
		\begin{overpic}[height=30mm, keepaspectratio, trim=80 20 80 10, clip=False]{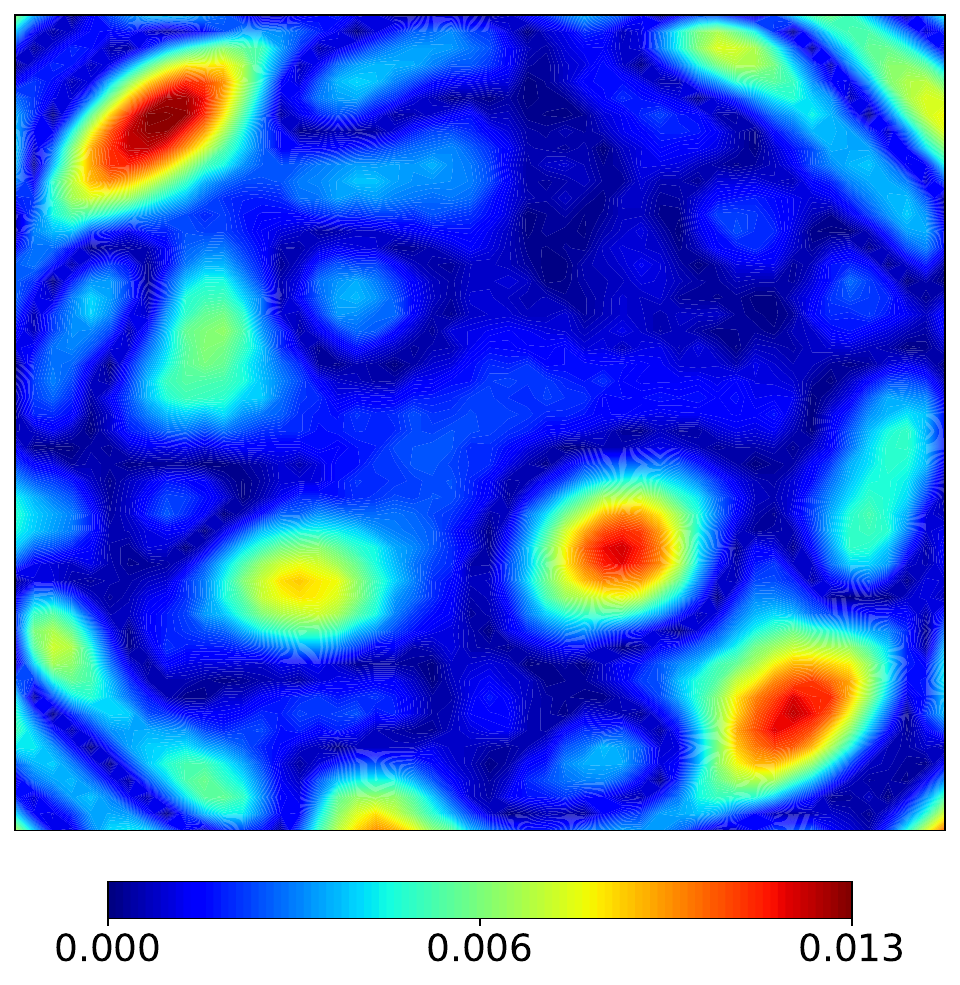}
			\put(32.5, 105) {$f$}
		\end{overpic}
        \hspace{10mm} 
		\begin{overpic}[height=30mm, keepaspectratio, trim=80 20 80 10, clip=False]{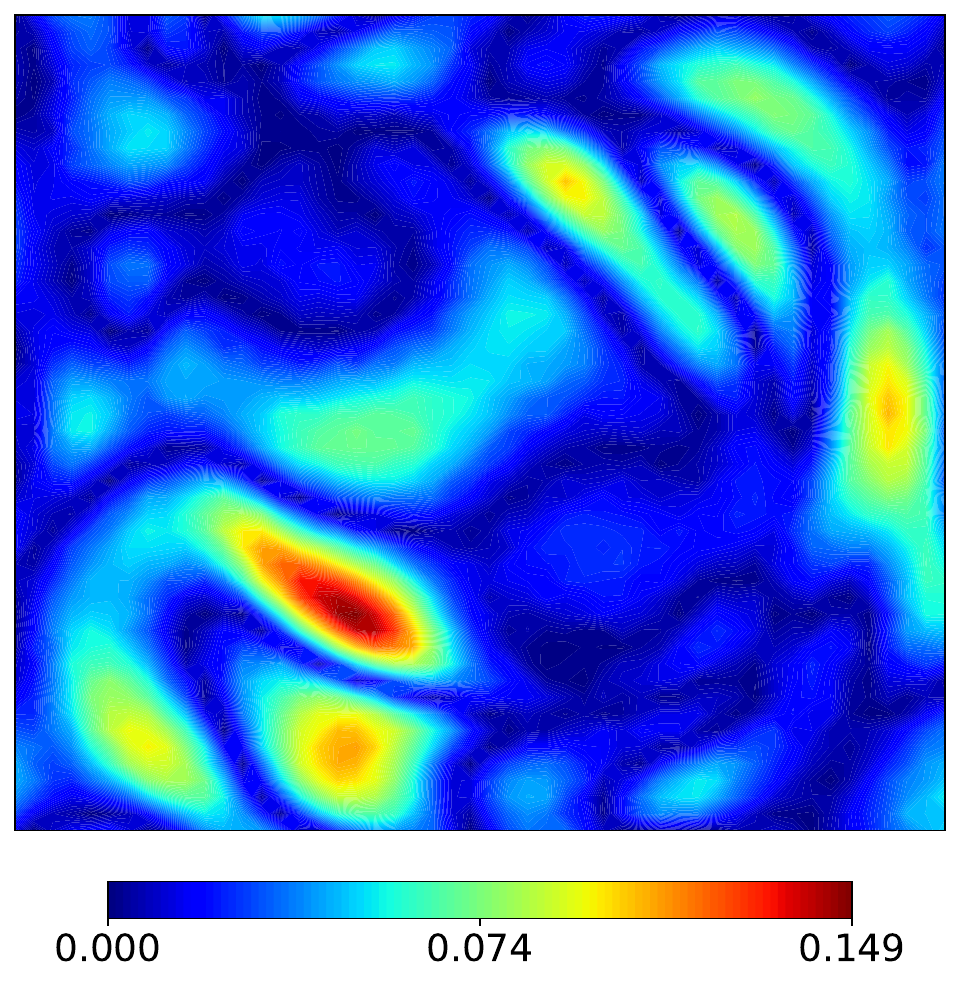}
			\put(32.5, 105) {$u$}
		\end{overpic}
        \hspace{10mm} 
		\begin{overpic}[height=30mm, keepaspectratio, trim=80 20 80 10, clip=False]{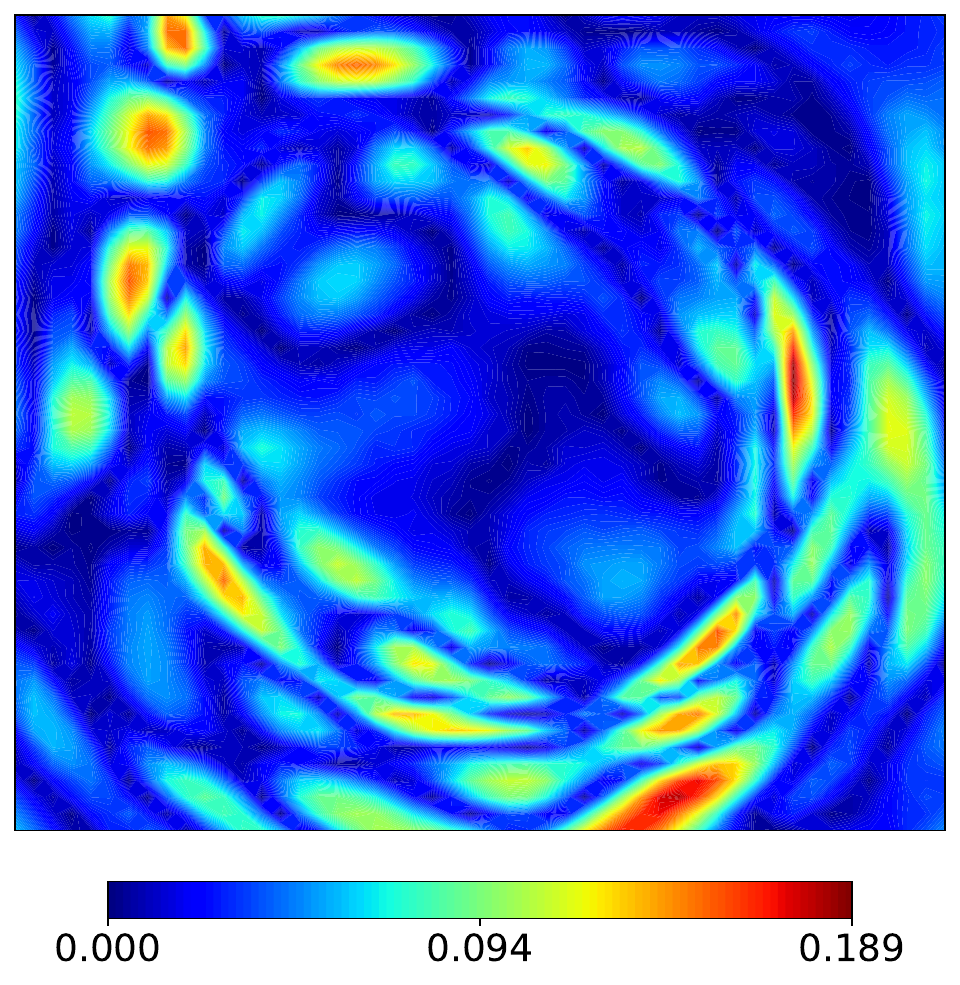}
			\put(32.5, 105) {$f$}
		\end{overpic}
        \hspace{-1.2mm}
		\vspace{5mm} 

		\begin{overpic}[height=30mm, keepaspectratio, trim=80 20 80 10, clip=False]{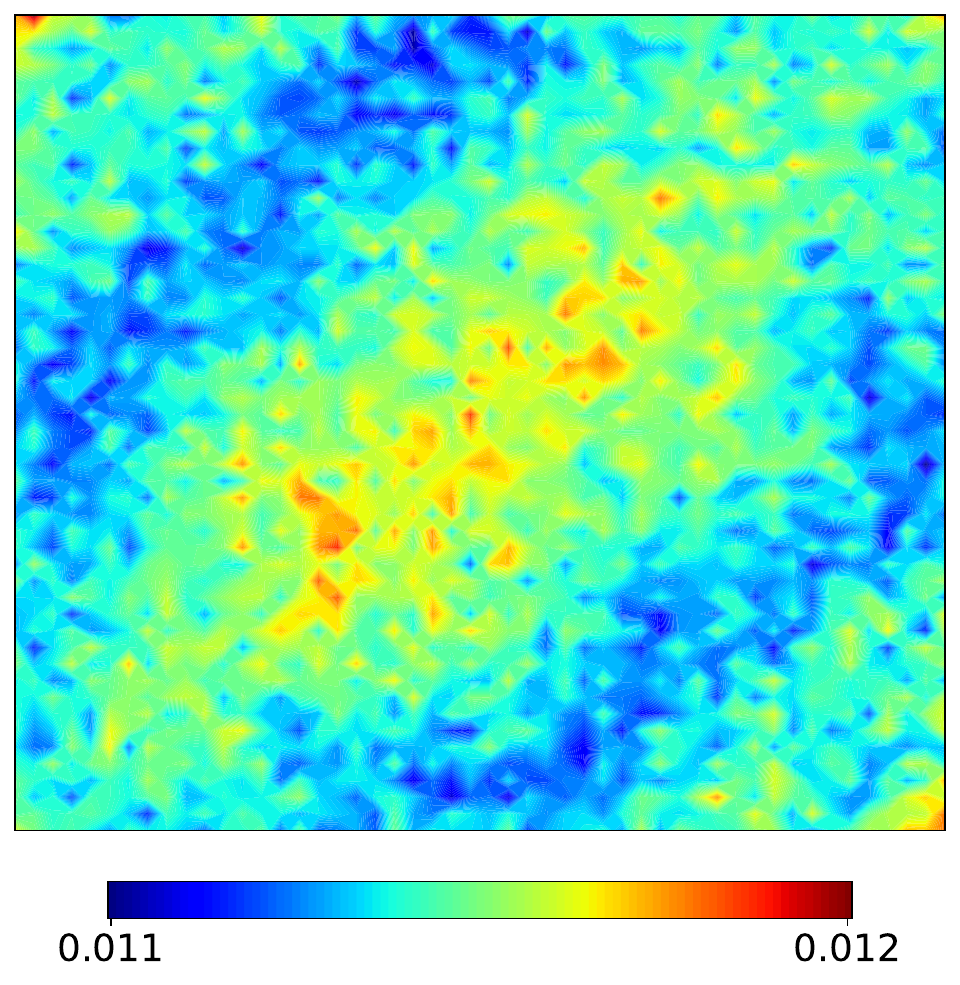}
			\put(-55, 50) {stds}
		\end{overpic}
        \hspace{10mm} 
		\begin{overpic}[height=30mm, keepaspectratio, trim=80 20 80 10, clip=False]{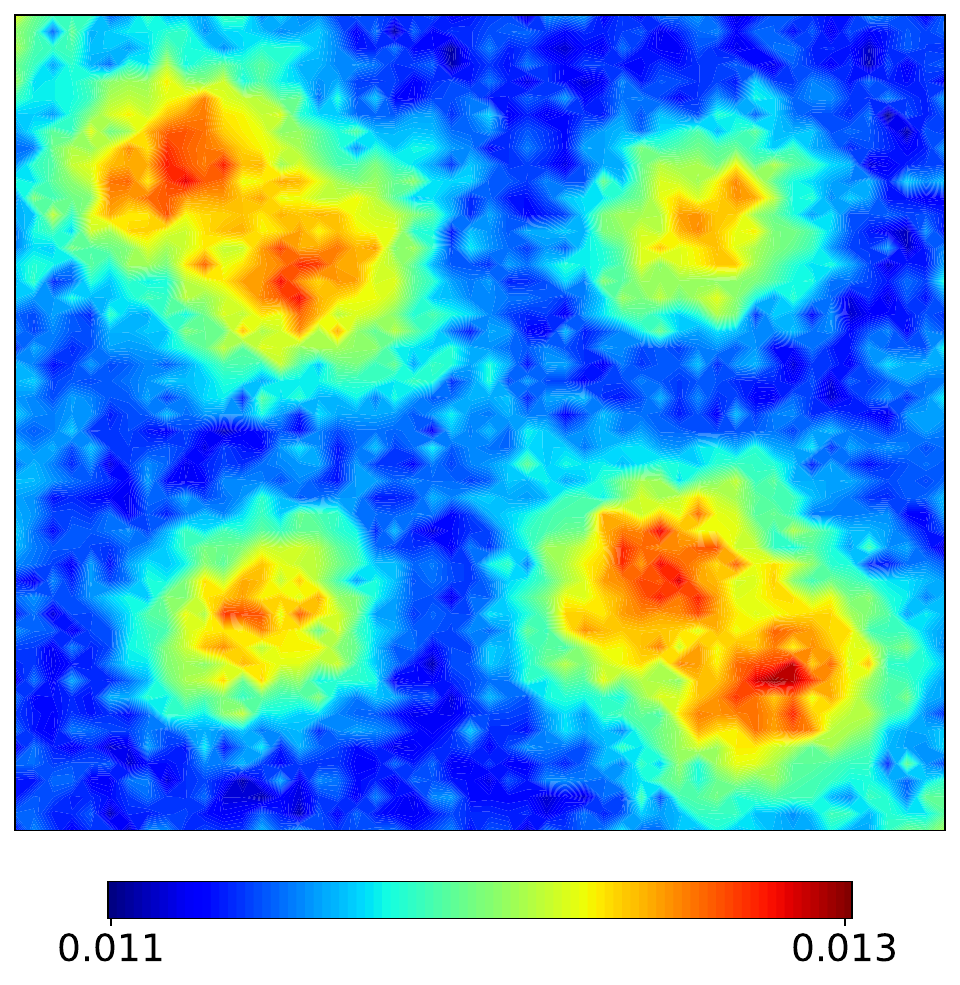}
			\put(-30, -17){(a)}
		\end{overpic}
        \hspace{10mm} 
		\begin{overpic}[height=30mm, keepaspectratio, trim=80 20 80 10, clip=False]{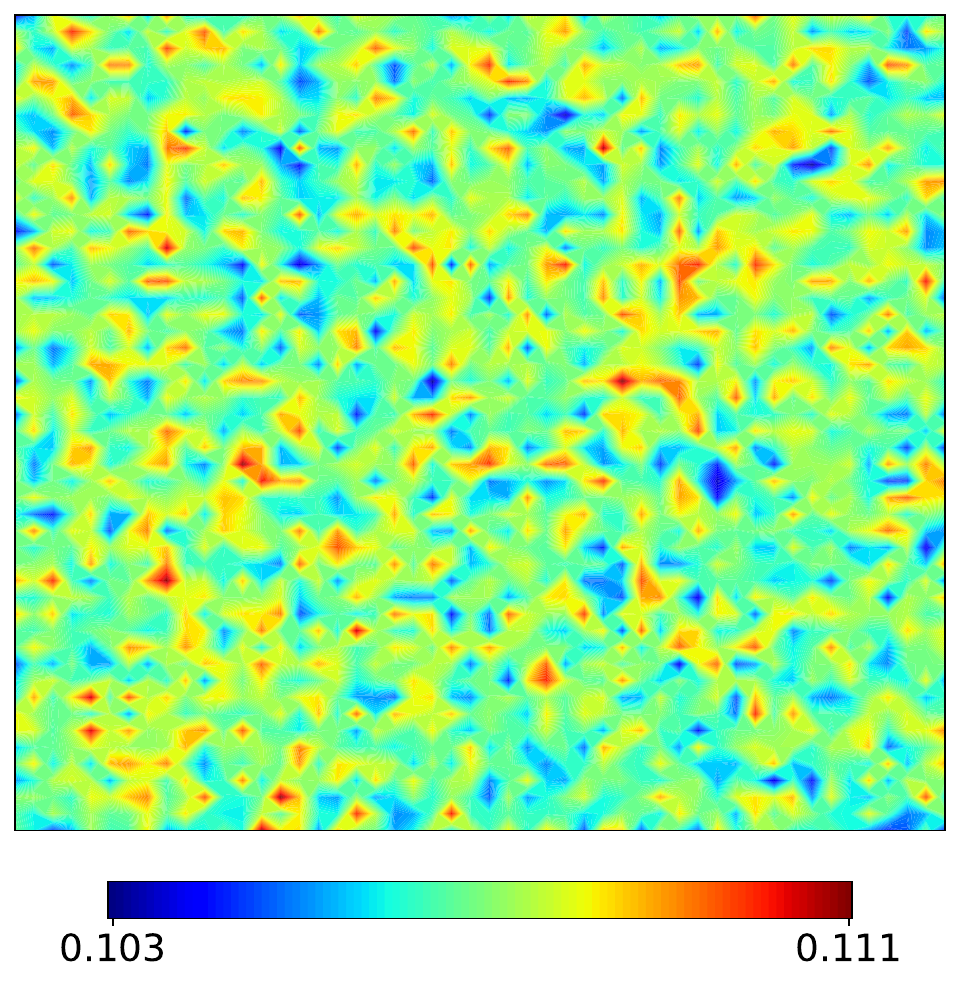}
		\end{overpic}
        \hspace{10mm} 
		\begin{overpic}[height=30mm, keepaspectratio, trim=80 20 80 10, clip=False]{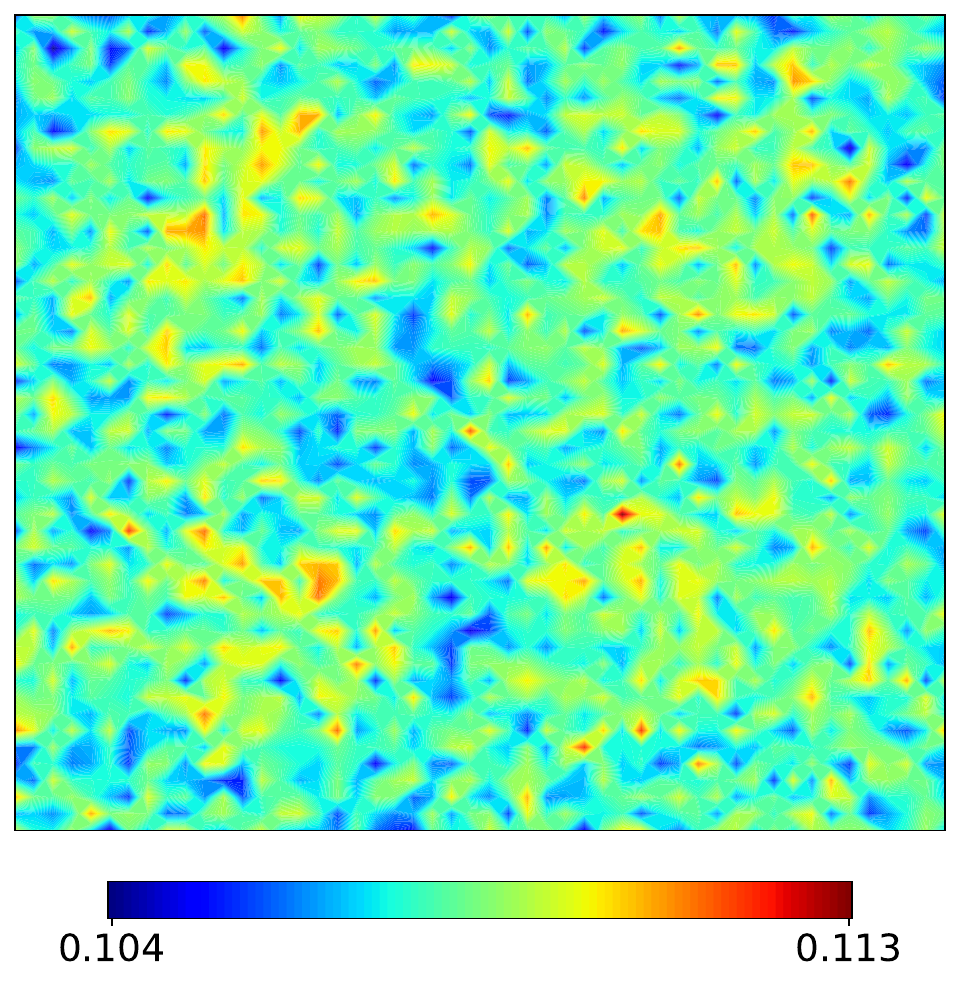}
			\put(-30, -17){(b)}
		\end{overpic}
        \vspace{5mm} 
	\end{center}
	\caption{2D nonlinear diffusion-reaction equation. Predicted errors and standard deviations for $u$ from different methods with two data noise scales. (a) $\epsilon_f\sim \mc{N}(0, 0.01^2),\epsilon_u\sim\mc{N}(0, 0.01^2)$. (b) $\epsilon_f\sim \mc{N}(0, 0.1^2),\epsilon_u\sim\mc{N}(0, 0.1^2)$.}
	\label{2D_nonlinear_inverse}
\end{figure}

The predicted values for $\lambda$ with different methods are displayed in Table \ref{table:kpred}. The predicted mean and standard deviation of the LVM-GP method is also provided in Figure \ref{2D_nonlinear_inverse}. All numerical results demonstrate the effectiveness of the proposed method.  In this numerical example, the networks employed in both the encoder and decoder adopt a 3-layer fully-connected neural network architecture, with each hidden layer containing 128 neurons. All other parameters are the same as in the first mention.

\begin{table}[H]
	\centering
	{\footnotesize
		\begin{tabular}{cc|ccc}
			\hline \hline
			\multicolumn{2}{c|}{Noise scale} & {LVM-GP} & {B-PINN-HMC} & {Deep Ensemble}                                                              \\
			\hline
			\multirow{2}{*}{0.01}             & Mean    & 1.0003        & 1.0005                                & 1.0047                                \\
			                                  & Std     &  {\footnotesize $4.58 \times 10^{-3}$}           & {\footnotesize $5.75 \times 10^{-3}$} & {\footnotesize $4.12 \times 10^{-3}$} \\
			\multirow{2}{*}{0.1}              & Mean    & 0.9916        & 0.9781                                & 0.9302                                \\
			                                  & Std     & {\footnotesize $5.70 \times 10^{-3}$}        & {\footnotesize $4.98 \times 10^{-2}$} & {\footnotesize $2.60 \times 10^{-2}$} \\
			\hline \hline
		\end{tabular}
	}
	\caption{2D nonlinear diffusion-reaction system: Predicted mean and standard deviation for
		the reaction rate $\lambda$ using different uncertainty-induced methods. $\lambda = 1$ is the exact
		solution. }
	\label{table:kpred}
\end{table}

\subsubsection{Six-dimensional source inverse problem}
We now apply the LVM-GP to a high-dimensional contaminant source inversion problem, in which the locations
of three contaminant sources are inferred based on a number of noisy observations. The governing PDE is given as follows:
\begin{equation}
	-\lambda (\partial_{x_1}^2 + \partial_{x_2}^2) u -f_2 = f_1, \quad (x_1,x_2)\in [0,1]^2,
\end{equation}
with zero boundary conditions. Here $u$ is the concentration of the contaminant, $\lambda=0.02$ is the diffusion coefficient, and $f_1$ is the known source term. The unknown source term $f_2$ is parameterized as
\begin{equation}
	f_2 = \sum_{i=1}^{3} k_i \exp\left[-0.5\frac{\Vert \bm{x}-\bm{x}_{c,i}\Vert^2}{0.15^2}\right],
\end{equation}
where $\bm{k}=(2, -3, 0.5)$ are known constants, and ${\bm{x}_{c,i}}$ denote the centers of the contaminant, which will
be inferred by noise observations of $u$ and $f_1$.

Here,  we set the exact $f_1$ as $f_1=0.1\sin(\pi x_1)\sin(\pi x_2)$ and the exact locations for three sources are:
$\bm{x}_{c,1}=(0.3, 0.3), \bm{x}_{c,2}=(0.75, 0.75)$ and $\bm{x}_{c,3}=(0.2, 0.7)$. We consider the case where
the noise for the measurements is: $\epsilon_u\sim \mathcal{N}(0, 0.1^2)$ and $\epsilon_{f_1}\sim \mathcal{N}(0, 0.01^2)$. The exact solution of $u$ is obtained via finite method method in Fenics.
We then utilize 1000 and 200 random samples for $u$ and $f_1$ as training data, respectively. In this numerical experiment, both the encoder and decoder networks employ a 3-layer fully-connected architecture with 128 neurons per hidden layer with Mish activation functions, while the $\lambda$-prediction network $\Lambda$ adopts a 2-layer fully-connected structure with 128 neurons per layer with hyperbolic tangent activation. Here, the term $\sigma_{u}$ in (\ref{decoder_u}) is initialized to 0.1. The optimization procedure consists of 20,000 iterations, the first 10,000 iterations exclusively optimize the mean-related network parameters, followed by another 10,000 iterations that simultaneously optimize the standard deviation parameters while fine-tuning the mean-related parameters. Throughout the training procedure, the Adam optimizer is employed with a persistently maintained learning rate of $0.001$. The numerical results in Table \ref{table:inversion_6D} demonstrate the effectiveness of the proposed method for high-dimensional source inversion problems. 

\begin{table}[H]
	\centering
	{\footnotesize
		\begin{tabular}{cc|ccc}
			\hline \hline
			\multicolumn{2}{c|}{ }      & $\bm{x}_{c,1}$      & $\bm{x}_{c,2}$   & $\bm{x}_{c,3}$                      \\
			\hline
			\multirow{2}{*}{LVM-GP}       & Mean                & (0.2927, 0.3022) & (0.7433, 0.7542) & (0.2065, 0.7569) \\
			                            & Std ($\times 10^2$)                & (2.79, 6.25)               & (2.87, 2.36)               & (5.49, 6.34)               \\
			\multirow{2}{*}{B-PINN-HMC} & Mean                & (0.3014, 0.2883) & (0.7473,0.7496)  & (0.2268, 0.6519) \\
			                            & Std ($\times 10^3$) & (3.08, 3.45)     & (3.51, 2.52)     & (18.97, 11.47)   \\
			\hline \hline
		\end{tabular}
	}
	\caption{Source inversion problem. Predicted mean and standard deviation for the locations using different
		uncertainty-induced methods.}
	\label{table:inversion_6D}
\end{table}
\section{Conclusion}
 In this work, we introduced LVM-GP, a novel physics-informed probabilistic framework for solving forward and inverse PDE problems with noisy data. Our method constructs a high-dimensional latent representation by combining a Gaussian process prior with a data-dependent confidence function, and employs a neural operator to map the latent representation to solution predictions. Physical laws are incorporated as soft constraints to ensure consistency with the governing PDEs. The experimental results demonstrate that our proposed approach can deliver accurate predictions while providing reliable uncertainty estimates.

However, there are many important issues need to be addressed. Theoretically, a rigorous analysis of stability and accuracy is still lacking, the mathematical well-posedness of the revised regularization term has not yet been fully established. In addition, the capabilities of LVM-GP methods for time-dependent PDE problems should be further explored and developed. Finally, incorporating the proposed framework into the setting of physics-informed operator learning represents a promising direction for future research.

\section*{Acknowledge}
The first author is partially supported by Guangdong and Hong Kong Universities “1+1+1” Joint Research Collaboration Scheme.
\appendix

\section{Derivative of Gaussian Process}
\label{sec:gp_derivative}

In this section, we provide some details for computing derivative of a 
Gaussian process (GP). Here we focus on two complementary approaches for computing the derivative of a Gaussian process, assuming a scalar-valued output: 
(i) an exact formulation based on the Karhunen–Lo\`eve (KL) expansion, and (ii) a finite-dimensional joint sampling approach derived from multivariate Gaussian identities.

Let $\bm{z}_0(\bm{x}) \sim \mathcal{GP}(0, k(\bm{x}, \bm{x}'))$ be a zero-mean Gaussian process defined on a compact domain $\mathcal{X} \subset \mathbb{R}^d$. Assuming that the kernel $k$ is continuous, symmetric, and positive semi-definite, Mercer’s theorem ensures that $k$ admits the spectral decomposition:
\[
k(\bm{x}, \bm{x}') = \sum_{n=1}^{\infty} \lambda_n \phi_n(\bm{x}) \phi_n(\bm{x}'),
\]
where $\{\lambda_n\}$ are the non-negative eigenvalues and $\{\phi_n\}$ are the corresponding orthonormal eigen-functions in $L^2(\mathcal{X})$.

The GP can then be represented as:
\[
\bm{z}_0(\bm{x}) = \sum_{n=1}^{\infty} \sqrt{\lambda_n} \xi_n \phi_n(\bm{x}), \quad \xi_n \sim \mathcal{N}(0, 1) \text{ i.i.d.}
\]

Assuming sufficient smoothness of the eigen-functions $\phi_n$, we can differentiate term-by-term to obtain the derivative process:
\[
\mc{N}_{\bm{x}} [\bm{z}_0(\bm{x})] = \sum_{n=1}^{\infty} \sqrt{\lambda_n} \xi_n \mc{N}_{\bm{x}} [\phi_n(\bm{x})],
\]
which is itself a mean-zero Gaussian process. This representation is exact and infinite-dimensional. In practice, one may truncate the series at a finite number of terms $N$ for numerical approximation.

\subsection{Joint Sampling of Function and Derivative Values}
Take the input dimension $ d = 1 $  and output dimension $d_z=1$ for simplicity. Assume we adopt the squared exponential kernel
\[
k(x,x') = \sigma_K^2 \exp\left(-\frac{(x - x')^2}{2\ell^2}\right),
\]
where $ \ell $ is the length scale and $ \sigma_K^2 $ is the variance of the kernel. Let the observation data consist of function values $ \{x_i, u(x_i)\}_{i=1}^{N_u} $ and differential observations $ \{x_i, f(x_i)\}_{i=1}^{N_f} $. We consider the joint distribution of the Gaussian process $ z_0 $ and its first and second derivatives evaluated at input locations $ X_u $ and $ X_f $:

\begin{equation*}
	\begin{bmatrix}
		z_0(X_u) \\
		z_0(X_f) \\
		\frac{\partial z_0}{\partial x}(X_f) \\
		\frac{\partial^2 z_0}{\partial x^2}(X_f)
	\end{bmatrix} \sim \mathcal{N}\left(0, K\right),
\end{equation*}
where the covariance matrix $ K $ is given by:
	\begin{equation*}
		K = \begin{bmatrix}
			k(X_u, X_u) & k(X_u, X_f) & k_y(X_u, X_f) & k_{yy}(X_u,X_f) \\
			k(X_f, X_u) & k(X_f, X_f) & k_y(X_f, X_f) & k_{yy}(X_f, X_f) \\
			k_x(X_f, X_u) & k_x(X_f, X_f) & k_{xy}(X_f, X_f) & k_{xyy}(X_f, X_f) \\
			k_{xx}(X_f,X_u) & k_{xx}(X_f, X_f) & k_{xxy}(X_f, X_f) & k_{xxyy}(X_f, X_f)
		\end{bmatrix}.
	\end{equation*}
Here, the entries such as $ k_x(x, x') = \frac{\partial}{\partial x'} k(x, x') $, $ k_{xx}(x, x') = \frac{\partial^2}{\partial x \partial x'} k(x, x') $, and so on, denote the mixed partial derivatives of the kernel function. Each sub-block represents the covariance between function values and derivatives, computed analytically from the chosen kernel.

This indicates that we can directly sample from the joint distribution of $ z_0(X_u) $, $ z_0(X_f) $, $ \frac{\partial z_0}{\partial x}(X_f) $, and $ \frac{\partial^2 z_0}{\partial x^2}(X_f) $ using standard multivariate Gaussian sampling techniques. Suppose the nonlinear differential operator $ \mathcal{N}_x(\cdot) $ takes the form
\begin{equation*}
	\mathcal{N}_x[z_0](x) = F\left(z_0(x), \frac{\partial z_0}{\partial x}(x), \frac{\partial^2 z_0}{\partial x^2}(x)\right),
\end{equation*}
where $ F $ is a given nonlinear function. Then, for each sample realization indexed by a random variable $ \bm{\omega} $ drawn from the joint Gaussian distribution, the operator response is computed as
\begin{equation*}
	\mathcal{N}_x[z_0](x; \bm{\omega}) = F\left(z_0(x; \bm{\omega}), \frac{\partial z_0}{\partial x}(x; \bm{\omega}), \frac{\partial^2 z_0}{\partial x^2}(x; \bm{\omega})\right).
\end{equation*}
Here, the evaluation of $ \mathcal{N}_x $ is performed pointwise for each realization of the GP and its derivatives, yielding a non-Gaussian output distribution that can be analyzed empirically across samples.

\section{Supplementary Numerical Experiments for DeepONet-Type Decoder}
\label{sec: DeepONet-Type forward}

\begin{figure} [!h]
	\begin{center}
		\begin{overpic}[width=0.25\textwidth, trim=0 0 0 0, clip=True]{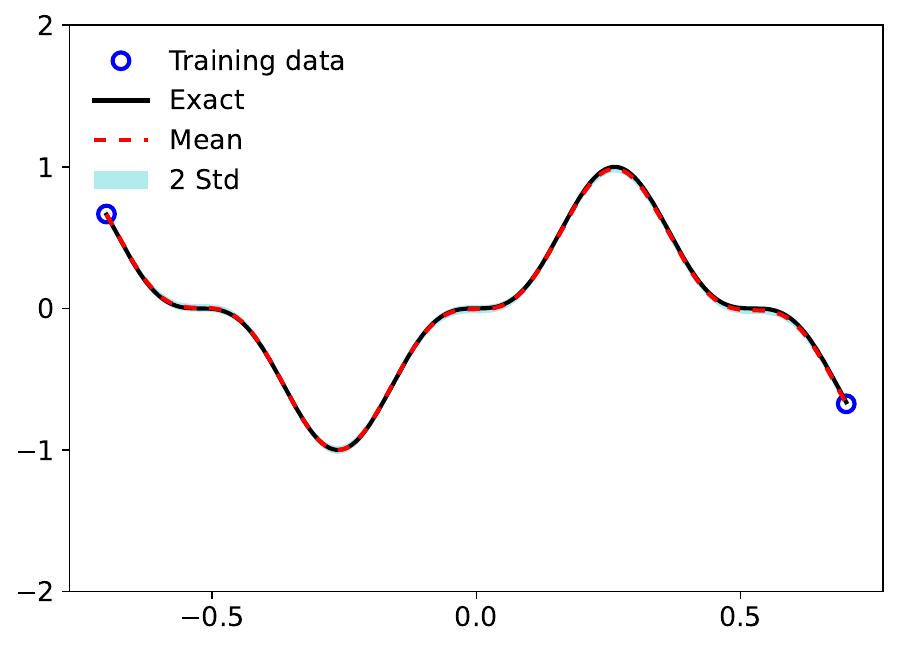}
		\end{overpic}
		\begin{overpic}[width=0.25\textwidth, trim=0 0 0 0, clip=True]{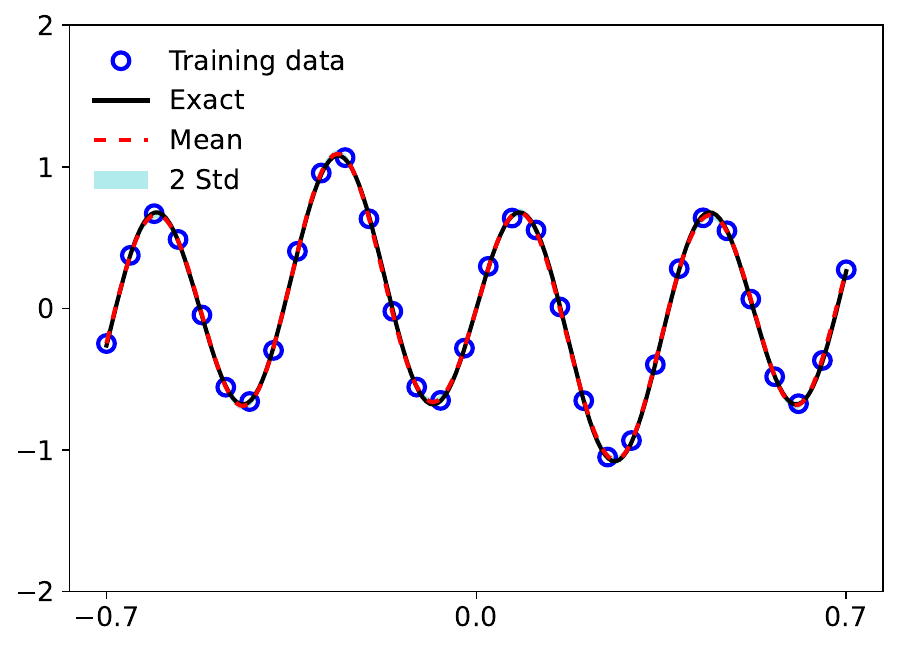}
		\end{overpic}
        \subcaption{}

		\begin{overpic}[width=0.25\textwidth, trim=0 0 0 0, clip=True]{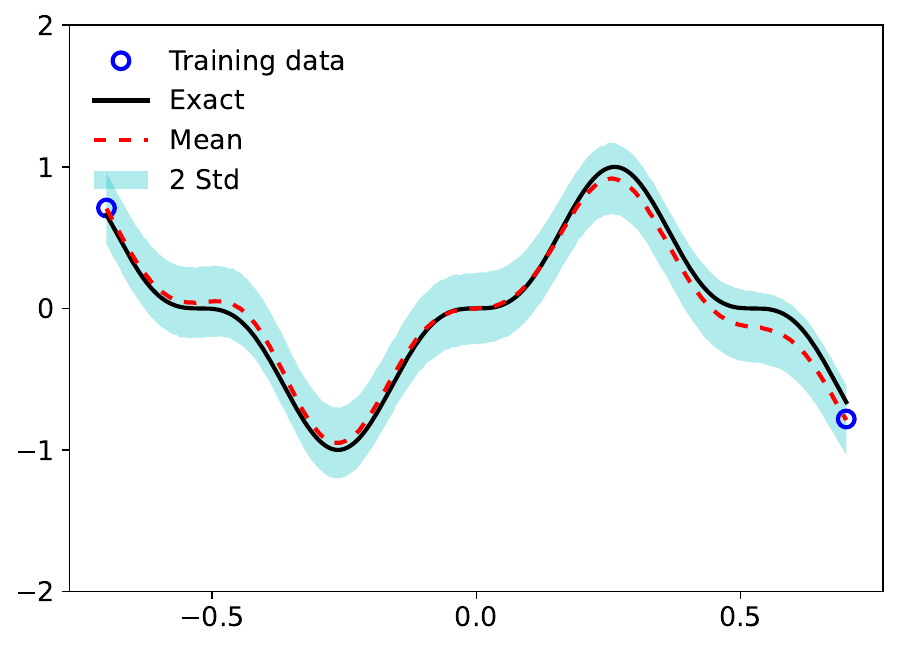}
		\end{overpic}
		\begin{overpic}[width=0.25\textwidth, trim=0 0 0 0, clip=True]{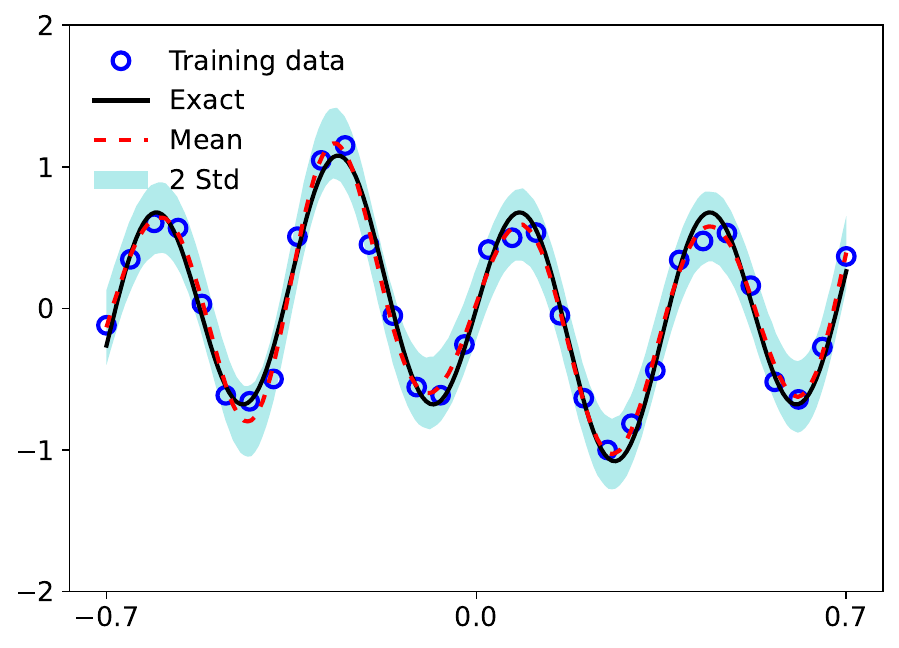}
		\end{overpic}
		\subcaption{}
	\end{center}
	\caption{1D Poisson equation. Predicted $u$ and $f$ from DeepONet-type structure with two data noise scales. (a) $\epsilon_f\sim \mc{N}(0, 0.01^2), \epsilon_b\sim \mc{N}(0, 0.01^2)$. (b) $\epsilon_f \sim \mc{N}(0, 0.1^2), \epsilon_b\sim \mc{N}(0, 0.1^2)$.  }
	\label{1D_poisson_unknown_noise_deeponet}
\end{figure}

In the section \ref{1D_poisson}, we presented the numerical results for the decoder model with an FNO-type architecture. Here, we further provide experimental results for the decoder employing a DeepONet-type structure. All parameter configurations in these experiments remain consistent with those of the neural operator type structure to ensure a fair comparison. As shown in Figure \ref{1D_poisson_unknown_noise_deeponet}, the DeepONet-type framework is equally effective, delivering well-performing mean predictions and reliable uncertainty quantification. 

\bibliography{ref.bib}

\begin{thebibliography}{51}
\expandafter\ifx\csname natexlab\endcsname\relax\def\natexlab#1{#1}\fi
\providecommand{\bibinfo}[2]{#2}
\ifx\xfnm\relax \def\xfnm[#1]{\unskip,\space#1}\fi
\bibitem[{Karniadakis et~al.(2021)Karniadakis, Kevrekidis, Lu, Perdikaris, Wang, and Yang}]{karniadakis2021physics}
\bibinfo{author}{G.~E. Karniadakis}, \bibinfo{author}{I.~G. Kevrekidis}, \bibinfo{author}{L.~Lu}, \bibinfo{author}{P.~Perdikaris}, \bibinfo{author}{S.~Wang}, \bibinfo{author}{L.~Yang},
\newblock \bibinfo{title}{Physics-informed machine learning},
\newblock \bibinfo{journal}{Nature Reviews Physics} \bibinfo{volume}{3} (\bibinfo{year}{2021}) \bibinfo{pages}{422--440}.
\bibitem[{Willard et~al.(2022)Willard, Jia, Xu, Steinbach, and Kumar}]{willard2022integrating}
\bibinfo{author}{J.~Willard}, \bibinfo{author}{X.~Jia}, \bibinfo{author}{S.~Xu}, \bibinfo{author}{M.~Steinbach}, \bibinfo{author}{V.~Kumar},
\newblock \bibinfo{title}{Integrating scientific knowledge with machine learning for engineering and environmental systems},
\newblock \bibinfo{journal}{ACM Computing Surveys} \bibinfo{volume}{55} (\bibinfo{year}{2022}) \bibinfo{pages}{1--37}.
\bibitem[{E(2020)}]{weinan2020machine}
\bibinfo{author}{W.~E},
\newblock \bibinfo{title}{Machine learning and computational mathematics},
\newblock \bibinfo{journal}{arXiv preprint arXiv:2009.14596}  (\bibinfo{year}{2020}).
\bibitem[{Raissi et~al.(2019)Raissi, Perdikaris, and Karniadakis}]{raissi2019physics}
\bibinfo{author}{M.~Raissi}, \bibinfo{author}{P.~Perdikaris}, \bibinfo{author}{G.~E. Karniadakis},
\newblock \bibinfo{title}{Physics-informed neural networks: A deep learning framework for solving forward and inverse problems involving nonlinear partial differential equations},
\newblock \bibinfo{journal}{Journal of Computational Physics} \bibinfo{volume}{378} (\bibinfo{year}{2019}) \bibinfo{pages}{686--707}.
\bibitem[{E and Yu(2018)}]{weinan2018deep}
\bibinfo{author}{W.~E}, \bibinfo{author}{B.~Yu},
\newblock \bibinfo{title}{The deep {R}itz method: A deep learning-based numerical algorithm for solving variational problems},
\newblock \bibinfo{journal}{Communications in Mathematics and Statistics} \bibinfo{volume}{6} (\bibinfo{year}{2018}).
\bibitem[{Sirignano and Spiliopoulos(2018)}]{sirignano2018dgm}
\bibinfo{author}{J.~Sirignano}, \bibinfo{author}{K.~Spiliopoulos},
\newblock \bibinfo{title}{{DGM}: A deep learning algorithm for solving partial differential equations},
\newblock \bibinfo{journal}{Journal of computational physics} \bibinfo{volume}{375} (\bibinfo{year}{2018}) \bibinfo{pages}{1339--1364}.
\bibitem[{Wang et~al.(2017)Wang, Wu, and Xiao}]{wang2017physics}
\bibinfo{author}{J.-X. Wang}, \bibinfo{author}{J.-L. Wu}, \bibinfo{author}{H.~Xiao},
\newblock \bibinfo{title}{Physics-informed machine learning approach for reconstructing reynolds stress modeling discrepancies based on {DNS} data},
\newblock \bibinfo{journal}{Physical Review Fluids} \bibinfo{volume}{2} (\bibinfo{year}{2017}) \bibinfo{pages}{034603}.
\bibitem[{Raissi et~al.(2020)Raissi, Yazdani, and Karniadakis}]{raissi2020hidden}
\bibinfo{author}{M.~Raissi}, \bibinfo{author}{A.~Yazdani}, \bibinfo{author}{G.~E. Karniadakis},
\newblock \bibinfo{title}{Hidden fluid mechanics: Learning velocity and pressure fields from flow visualizations},
\newblock \bibinfo{journal}{Science} \bibinfo{volume}{367} (\bibinfo{year}{2020}) \bibinfo{pages}{1026--1030}.
\bibitem[{Han et~al.(2018)Han, Jentzen, and E}]{han2018solving}
\bibinfo{author}{J.~Han}, \bibinfo{author}{A.~Jentzen}, \bibinfo{author}{W.~E},
\newblock \bibinfo{title}{Solving high-dimensional partial differential equations using deep learning},
\newblock \bibinfo{journal}{Proceedings of the National Academy of Sciences} \bibinfo{volume}{115} (\bibinfo{year}{2018}) \bibinfo{pages}{8505--8510}.
\bibitem[{Huang et~al.(2022)Huang, Wang, and Zhou}]{huang2022augmented}
\bibinfo{author}{J.~Huang}, \bibinfo{author}{H.~Wang}, \bibinfo{author}{T.~Zhou},
\newblock \bibinfo{title}{An augmented {L}agrangian deep learning method for variational problems with essential boundary conditions},
\newblock \bibinfo{journal}{Communications in Computational Physics} \bibinfo{volume}{31} (\bibinfo{year}{2022}) \bibinfo{pages}{966--986}.
\bibitem[{Zang et~al.(2020)Zang, Bao, Ye, and Zhou}]{zang2020weak}
\bibinfo{author}{Y.~Zang}, \bibinfo{author}{G.~Bao}, \bibinfo{author}{X.~Ye}, \bibinfo{author}{H.~Zhou},
\newblock \bibinfo{title}{Weak adversarial networks for high-dimensional partial differential equations},
\newblock \bibinfo{journal}{Journal of Computational Physics} \bibinfo{volume}{411} (\bibinfo{year}{2020}) \bibinfo{pages}{109409}.
\bibitem[{Lu et~al.(2021)Lu, Jin, Pang, Zhang, and Karniadakis}]{lu2021learning}
\bibinfo{author}{L.~Lu}, \bibinfo{author}{P.~Jin}, \bibinfo{author}{G.~Pang}, \bibinfo{author}{Z.~Zhang}, \bibinfo{author}{G.~E. Karniadakis},
\newblock \bibinfo{title}{Learning nonlinear operators via {D}eeponet based on the universal approximation theorem of operators},
\newblock \bibinfo{journal}{Nature machine intelligence} \bibinfo{volume}{3} (\bibinfo{year}{2021}) \bibinfo{pages}{218--229}.
\bibitem[{Chen and Chen(1995)}]{chen1995universal}
\bibinfo{author}{T.~Chen}, \bibinfo{author}{H.~Chen},
\newblock \bibinfo{title}{Universal approximation to nonlinear operators by neural networks with arbitrary activation functions and its application to dynamical systems},
\newblock \bibinfo{journal}{IEEE Transactions on Neural Networks} \bibinfo{volume}{6} (\bibinfo{year}{1995}) \bibinfo{pages}{911--917}.
\bibitem[{Li et~al.(2020)Li, Kovachki, Azizzadenesheli, Liu, Bhattacharya, Stuart, and Anandkumar}]{li2020fourier}
\bibinfo{author}{Z.~Li}, \bibinfo{author}{N.~Kovachki}, \bibinfo{author}{K.~Azizzadenesheli}, \bibinfo{author}{B.~Liu}, \bibinfo{author}{K.~Bhattacharya}, \bibinfo{author}{A.~Stuart}, \bibinfo{author}{A.~Anandkumar},
\newblock \bibinfo{title}{Fourier neural operator for parametric partial differential equations},
\newblock \bibinfo{journal}{arXiv preprint arXiv:2010.08895}  (\bibinfo{year}{2020}).
\bibitem[{Cao(2021)}]{cao2021choose}
\bibinfo{author}{S.~Cao},
\newblock \bibinfo{title}{Choose a transformer: Fourier or galerkin},
\newblock \bibinfo{journal}{Advances in neural information processing systems} \bibinfo{volume}{34} (\bibinfo{year}{2021}) \bibinfo{pages}{24924--24940}.
\bibitem[{Kissas et~al.(2022)Kissas, Seidman, Guilhoto, Preciado, Pappas, and Perdikaris}]{kissas2022learning}
\bibinfo{author}{G.~Kissas}, \bibinfo{author}{J.~H. Seidman}, \bibinfo{author}{L.~F. Guilhoto}, \bibinfo{author}{V.~M. Preciado}, \bibinfo{author}{G.~J. Pappas}, \bibinfo{author}{P.~Perdikaris},
\newblock \bibinfo{title}{Learning operators with coupled attention},
\newblock \bibinfo{journal}{Journal of Machine Learning Research} \bibinfo{volume}{23} (\bibinfo{year}{2022}) \bibinfo{pages}{1--63}.
\bibitem[{Wang et~al.(2021)Wang, Wang, and Perdikaris}]{wang2021learning}
\bibinfo{author}{S.~Wang}, \bibinfo{author}{H.~Wang}, \bibinfo{author}{P.~Perdikaris},
\newblock \bibinfo{title}{Learning the solution operator of parametric partial differential equations with physics-informed deeponets},
\newblock \bibinfo{journal}{Science advances} \bibinfo{volume}{7} (\bibinfo{year}{2021}) \bibinfo{pages}{eabi8605}.
\bibitem[{Wang et~al.(2022)Wang, Wang, and Perdikaris}]{wang2022improved}
\bibinfo{author}{S.~Wang}, \bibinfo{author}{H.~Wang}, \bibinfo{author}{P.~Perdikaris},
\newblock \bibinfo{title}{Improved architectures and training algorithms for deep operator networks},
\newblock \bibinfo{journal}{Journal of Scientific Computing} \bibinfo{volume}{92} (\bibinfo{year}{2022}) \bibinfo{pages}{35}.
\bibitem[{Geneva and Zabaras(2020)}]{geneva2020modeling}
\bibinfo{author}{N.~Geneva}, \bibinfo{author}{N.~Zabaras},
\newblock \bibinfo{title}{Modeling the dynamics of {PDE} systems with physics-constrained deep auto-regressive networks},
\newblock \bibinfo{journal}{Journal of Computational Physics} \bibinfo{volume}{403} (\bibinfo{year}{2020}) \bibinfo{pages}{109056}.
\bibitem[{Psaros et~al.(2023)Psaros, Meng, Zou, Guo, and Karniadakis}]{psaros2023uncertainty}
\bibinfo{author}{A.~F. Psaros}, \bibinfo{author}{X.~Meng}, \bibinfo{author}{Z.~Zou}, \bibinfo{author}{L.~Guo}, \bibinfo{author}{G.~E. Karniadakis},
\newblock \bibinfo{title}{Uncertainty quantification in scientific machine learning: {M}ethods, metrics, and comparisons},
\newblock \bibinfo{journal}{Journal of Computational Physics}  (\bibinfo{year}{2023}) \bibinfo{pages}{111902}.
\bibitem[{Raissi et~al.(2017)Raissi, Perdikaris, and Karniadakis}]{raissi2017machine}
\bibinfo{author}{M.~Raissi}, \bibinfo{author}{P.~Perdikaris}, \bibinfo{author}{G.~E. Karniadakis},
\newblock \bibinfo{title}{Machine learning of linear differential equations using {Gaussian} processes},
\newblock \bibinfo{journal}{Journal of Computational Physics} \bibinfo{volume}{348} (\bibinfo{year}{2017}) \bibinfo{pages}{683--693}.
\bibitem[{Kendall and Gal(2017)}]{kendall2017uncertainties}
\bibinfo{author}{A.~Kendall}, \bibinfo{author}{Y.~Gal},
\newblock \bibinfo{title}{What uncertainties do we need in bayesian deep learning for computer vision?},
\newblock \bibinfo{journal}{Advances in neural information processing systems} \bibinfo{volume}{30} (\bibinfo{year}{2017}).
\bibitem[{Chen et~al.(2021)Chen, Hosseini, Owhadi, and Stuart}]{chen2021solving}
\bibinfo{author}{Y.~Chen}, \bibinfo{author}{B.~Hosseini}, \bibinfo{author}{H.~Owhadi}, \bibinfo{author}{A.~M. Stuart},
\newblock \bibinfo{title}{Solving and learning nonlinear {PDEs} with {Gaussian} processes},
\newblock \bibinfo{journal}{Journal of Computational Physics} \bibinfo{volume}{447} (\bibinfo{year}{2021}) \bibinfo{pages}{110668}.
\bibitem[{Chen et~al.(2023)Chen, Owhadi, and Sch{\"a}fer}]{chen2023sparse}
\bibinfo{author}{Y.~Chen}, \bibinfo{author}{H.~Owhadi}, \bibinfo{author}{F.~Sch{\"a}fer},
\newblock \bibinfo{title}{Sparse cholesky factorization for solving nonlinear {PDE}s via {Gaussian} processes},
\newblock \bibinfo{journal}{arXiv preprint arXiv:2304.01294}  (\bibinfo{year}{2023}).
\bibitem[{Yang et~al.(2021)Yang, Meng, and Karniadakis}]{yang2021b}
\bibinfo{author}{L.~Yang}, \bibinfo{author}{X.~Meng}, \bibinfo{author}{G.~E. Karniadakis},
\newblock \bibinfo{title}{B-{PINNs}: Bayesian physics-informed neural networks for forward and inverse {PDE} problems with noisy data},
\newblock \bibinfo{journal}{Journal of Computational Physics} \bibinfo{volume}{425} (\bibinfo{year}{2021}) \bibinfo{pages}{109913}.
\bibitem[{Linka et~al.(2022)Linka, Sch{\"a}fer, Meng, Zou, Karniadakis, and Kuhl}]{linka2022bayesian}
\bibinfo{author}{K.~Linka}, \bibinfo{author}{A.~Sch{\"a}fer}, \bibinfo{author}{X.~Meng}, \bibinfo{author}{Z.~Zou}, \bibinfo{author}{G.~E. Karniadakis}, \bibinfo{author}{E.~Kuhl},
\newblock \bibinfo{title}{Bayesian physics informed neural networks for real-world nonlinear dynamical systems},
\newblock \bibinfo{journal}{Computer Methods in Applied Mechanics and Engineering} \bibinfo{volume}{402} (\bibinfo{year}{2022}) \bibinfo{pages}{115346}.
\bibitem[{Lin et~al.(2023)Lin, Moya, and Zhang}]{lin2023b}
\bibinfo{author}{G.~Lin}, \bibinfo{author}{C.~Moya}, \bibinfo{author}{Z.~Zhang},
\newblock \bibinfo{title}{{B-DeepONet}: An enhanced {B}ayesian {DeepONet} for solving noisy parametric pdes using accelerated replica exchange {SGLD}},
\newblock \bibinfo{journal}{Journal of Computational Physics} \bibinfo{volume}{473} (\bibinfo{year}{2023}) \bibinfo{pages}{111713}.
\bibitem[{Lotfi et~al.(2022)Lotfi, Izmailov, Benton, Goldblum, and Wilson}]{lotfi2022bayesian}
\bibinfo{author}{S.~Lotfi}, \bibinfo{author}{P.~Izmailov}, \bibinfo{author}{G.~Benton}, \bibinfo{author}{M.~Goldblum}, \bibinfo{author}{A.~G. Wilson},
\newblock \bibinfo{title}{Bayesian model selection, the marginal likelihood, and generalization},
\newblock in: \bibinfo{booktitle}{International Conference on Machine Learning}, \bibinfo{organization}{PMLR}, pp. \bibinfo{pages}{14223--14247}.
\bibitem[{Lakshminarayanan et~al.(2017)Lakshminarayanan, Pritzel, and Blundell}]{lakshminarayanan2017simple}
\bibinfo{author}{B.~Lakshminarayanan}, \bibinfo{author}{A.~Pritzel}, \bibinfo{author}{C.~Blundell},
\newblock \bibinfo{title}{Simple and scalable predictive uncertainty estimation using deep ensembles},
\newblock \bibinfo{journal}{Advances in neural information processing systems} \bibinfo{volume}{30} (\bibinfo{year}{2017}).
\bibitem[{Fort et~al.(2019)Fort, Hu, and Lakshminarayanan}]{fort2019deep}
\bibinfo{author}{S.~Fort}, \bibinfo{author}{H.~Hu}, \bibinfo{author}{B.~Lakshminarayanan},
\newblock \bibinfo{title}{Deep ensembles: A loss landscape perspective},
\newblock \bibinfo{journal}{arXiv preprint arXiv:1912.02757}  (\bibinfo{year}{2019}).
\bibitem[{Malinin and Gales(2018)}]{malinin2018predictive}
\bibinfo{author}{A.~Malinin}, \bibinfo{author}{M.~Gales},
\newblock \bibinfo{title}{Predictive uncertainty estimation via prior networks},
\newblock \bibinfo{journal}{Advances in neural information processing systems} \bibinfo{volume}{31} (\bibinfo{year}{2018}).
\bibitem[{Yang et~al.(2022)Yang, Kissas, and Perdikaris}]{yang2022scalable}
\bibinfo{author}{Y.~Yang}, \bibinfo{author}{G.~Kissas}, \bibinfo{author}{P.~Perdikaris},
\newblock \bibinfo{title}{Scalable uncertainty quantification for deep operator networks using randomized priors},
\newblock \bibinfo{journal}{Computer Methods in Applied Mechanics and Engineering} \bibinfo{volume}{399} (\bibinfo{year}{2022}) \bibinfo{pages}{115399}.
\bibitem[{Amini et~al.(2020)Amini, Schwarting, Soleimany, and Rus}]{amini2020deep}
\bibinfo{author}{A.~Amini}, \bibinfo{author}{W.~Schwarting}, \bibinfo{author}{A.~Soleimany}, \bibinfo{author}{D.~Rus},
\newblock \bibinfo{title}{Deep evidential regression},
\newblock \bibinfo{journal}{Advances in Neural Information Processing Systems} \bibinfo{volume}{33} (\bibinfo{year}{2020}) \bibinfo{pages}{14927--14937}.
\bibitem[{Akhare et~al.(2023)Akhare, Luo, and Wang}]{akhare2023diffhybrid}
\bibinfo{author}{D.~Akhare}, \bibinfo{author}{T.~Luo}, \bibinfo{author}{J.-X. Wang},
\newblock \bibinfo{title}{Diffhybrid-uq: uncertainty quantification for differentiable hybrid neural modeling},
\newblock \bibinfo{journal}{arXiv preprint arXiv:2401.00161}  (\bibinfo{year}{2023}).
\bibitem[{Du et~al.(2024)Du, Parikh, Fan, Liu, and Wang}]{du2024conditional}
\bibinfo{author}{P.~Du}, \bibinfo{author}{M.~H. Parikh}, \bibinfo{author}{X.~Fan}, \bibinfo{author}{X.-Y. Liu}, \bibinfo{author}{J.-X. Wang},
\newblock \bibinfo{title}{Conditional neural field latent diffusion model for generating spatiotemporal turbulence},
\newblock \bibinfo{journal}{Nature Communications} \bibinfo{volume}{15} (\bibinfo{year}{2024}) \bibinfo{pages}{10416}.
\bibitem[{Guo et~al.(2023)Guo, Wu, Zhou, and Zhou}]{guo2023ib}
\bibinfo{author}{L.~Guo}, \bibinfo{author}{H.~Wu}, \bibinfo{author}{W.~Zhou}, \bibinfo{author}{T.~Zhou},
\newblock \bibinfo{title}{{IB-UQ}: Information bottleneck based uncertainty quantification for neural function regression and neural operator learning},
\newblock \bibinfo{journal}{arXiv preprint arXiv:2302.03271}  (\bibinfo{year}{2023}).
\bibitem[{Bergna et~al.(2025)Bergna, Depeweg, Ordonez, Plenk, Cartea, and Hernandez-Lobato}]{bergna2025post}
\bibinfo{author}{R.~Bergna}, \bibinfo{author}{S.~Depeweg}, \bibinfo{author}{S.~C. Ordonez}, \bibinfo{author}{J.~Plenk}, \bibinfo{author}{A.~Cartea}, \bibinfo{author}{J.~M. Hernandez-Lobato},
\newblock \bibinfo{title}{Post-hoc uncertainty quantification in {P}re-{T}rained {N}eural {N}etworks via {A}ctivation-{L}evel {G}aussian {P}rocesses},
\newblock \bibinfo{journal}{arXiv preprint arXiv:2502.20966}  (\bibinfo{year}{2025}).
\bibitem[{Nair et~al.(2025)Nair, Jacob, Howard, Drgona, and Stinis}]{nair2025pinns}
\bibinfo{author}{A.~S. Nair}, \bibinfo{author}{B.~Jacob}, \bibinfo{author}{A.~A. Howard}, \bibinfo{author}{J.~Drgona}, \bibinfo{author}{P.~Stinis},
\newblock \bibinfo{title}{E-{PINN}s: {E}pistemic {P}hysics-{I}nformed {N}eural {N}etworks},
\newblock \bibinfo{journal}{arXiv preprint arXiv:2503.19333}  (\bibinfo{year}{2025}).
\bibitem[{Zou et~al.(2022)Zou, Meng, Psaros, and Karniadakis}]{zou2022neuraluq}
\bibinfo{author}{Z.~Zou}, \bibinfo{author}{X.~Meng}, \bibinfo{author}{A.~F. Psaros}, \bibinfo{author}{G.~E. Karniadakis},
\newblock \bibinfo{title}{{NeuralUQ}: A comprehensive library for uncertainty quantification in neural differential equations and operators},
\newblock \bibinfo{journal}{arXiv preprint arXiv:2208.11866}  (\bibinfo{year}{2022}).
\bibitem[{Raissi et~al.(2018)Raissi, Perdikaris, and Karniadakis}]{raissi2018numerical}
\bibinfo{author}{M.~Raissi}, \bibinfo{author}{P.~Perdikaris}, \bibinfo{author}{G.~E. Karniadakis},
\newblock \bibinfo{title}{Numerical gaussian processes for time-dependent and nonlinear partial differential equations},
\newblock \bibinfo{journal}{SIAM Journal on Scientific Computing} \bibinfo{volume}{40} (\bibinfo{year}{2018}) \bibinfo{pages}{A172--A198}.
\bibitem[{Long et~al.(2022)Long, Wang, Krishnapriyan, Kirby, Zhe, and Mahoney}]{long2022autoip}
\bibinfo{author}{D.~Long}, \bibinfo{author}{Z.~Wang}, \bibinfo{author}{A.~Krishnapriyan}, \bibinfo{author}{R.~Kirby}, \bibinfo{author}{S.~Zhe}, \bibinfo{author}{M.~Mahoney},
\newblock \bibinfo{title}{{AutoIP}: A united framework to integrate physics into {G}aussian processes},
\newblock in: \bibinfo{booktitle}{International Conference on Machine Learning}, \bibinfo{organization}{PMLR}, pp. \bibinfo{pages}{14210--14222}.
\bibitem[{Fang et~al.(2023)Fang, Cooley, Long, Li, Kirby, and Zhe}]{fang2023solving}
\bibinfo{author}{S.~Fang}, \bibinfo{author}{M.~Cooley}, \bibinfo{author}{D.~Long}, \bibinfo{author}{S.~Li}, \bibinfo{author}{R.~Kirby}, \bibinfo{author}{S.~Zhe},
\newblock \bibinfo{title}{Solving high frequency and multi-scale pdes with gaussian processes},
\newblock \bibinfo{journal}{arXiv preprint arXiv:2311.04465}  (\bibinfo{year}{2023}).
\bibitem[{Zhong and Meidani(2023)}]{zhong2023pi}
\bibinfo{author}{W.~Zhong}, \bibinfo{author}{H.~Meidani},
\newblock \bibinfo{title}{P{I}-{VAE}: Physics-informed variational auto-encoder for stochastic differential equations},
\newblock \bibinfo{journal}{Computer Methods in Applied Mechanics and Engineering} \bibinfo{volume}{403} (\bibinfo{year}{2023}) \bibinfo{pages}{115664}.
\bibitem[{Garnelo et~al.(2018)Garnelo, Schwarz, Rosenbaum, Viola, Rezende, Eslami, and Teh}]{garnelo2018neural}
\bibinfo{author}{M.~Garnelo}, \bibinfo{author}{J.~Schwarz}, \bibinfo{author}{D.~Rosenbaum}, \bibinfo{author}{F.~Viola}, \bibinfo{author}{D.~J. Rezende}, \bibinfo{author}{S.~Eslami}, \bibinfo{author}{Y.~W. Teh},
\newblock \bibinfo{title}{Neural processes},
\newblock \bibinfo{journal}{arXiv preprint arXiv:1807.01622}  (\bibinfo{year}{2018}).
\bibitem[{Cheng et~al.(2024)Cheng, Malik, De, Becker, and Doostan}]{cheng2024bi}
\bibinfo{author}{N.~Cheng}, \bibinfo{author}{O.~A. Malik}, \bibinfo{author}{S.~De}, \bibinfo{author}{S.~Becker}, \bibinfo{author}{A.~Doostan},
\newblock \bibinfo{title}{Bi-fidelity variational auto-encoder for uncertainty quantification},
\newblock \bibinfo{journal}{Computer Methods in Applied Mechanics and Engineering} \bibinfo{volume}{421} (\bibinfo{year}{2024}) \bibinfo{pages}{116793}.
\bibitem[{Yang et~al.(2020)Yang, Zhang, and Karniadakis}]{yang2020physics}
\bibinfo{author}{L.~Yang}, \bibinfo{author}{D.~Zhang}, \bibinfo{author}{G.~E. Karniadakis},
\newblock \bibinfo{title}{Physics-informed generative adversarial networks for stochastic differential equations},
\newblock \bibinfo{journal}{SIAM Journal on Scientific Computing} \bibinfo{volume}{42} (\bibinfo{year}{2020}) \bibinfo{pages}{A292--A317}.
\bibitem[{Yang and Perdikaris(2019)}]{yang2019adversarial}
\bibinfo{author}{Y.~Yang}, \bibinfo{author}{P.~Perdikaris},
\newblock \bibinfo{title}{Adversarial uncertainty quantification in physics-informed neural networks},
\newblock \bibinfo{journal}{Journal of Computational Physics} \bibinfo{volume}{394} (\bibinfo{year}{2019}) \bibinfo{pages}{136--152}.
\bibitem[{Chen and Wu(2024)}]{chen2024positional}
\bibinfo{author}{J.~Chen}, \bibinfo{author}{K.~Wu},
\newblock \bibinfo{title}{Positional knowledge is all you need: Position-induced transformer ({PiT}) for operator learning},
\newblock \bibinfo{journal}{arXiv preprint arXiv:2405.09285}  (\bibinfo{year}{2024}).
\bibitem[{Le et~al.(2005)Le, Smola, and Canu}]{le2005heteroscedastic}
\bibinfo{author}{Q.~V. Le}, \bibinfo{author}{A.~J. Smola}, \bibinfo{author}{S.~Canu},
\newblock \bibinfo{title}{Heteroscedastic {Gaussian} process regression},
\newblock in: \bibinfo{booktitle}{Proceedings of the 22nd international conference on Machine learning}, pp. \bibinfo{pages}{489--496}.
\bibitem[{Misra(2019)}]{misra2019mish}
\bibinfo{author}{D.~Misra},
\newblock \bibinfo{title}{Mish: A self regularized non-monotonic activation function},
\newblock \bibinfo{journal}{arXiv preprint arXiv:1908.08681}  (\bibinfo{year}{2019}).
\bibitem[{Nobile et~al.(2008)Nobile, Tempone, and Webster}]{nobile2008sparse}
\bibinfo{author}{F.~Nobile}, \bibinfo{author}{R.~Tempone}, \bibinfo{author}{C.~Webster},
\newblock \bibinfo{title}{A sparse grid stochastic collocation method for partial differential equations with random input data},
\newblock \bibinfo{journal}{SIAM Journal on Numerical Analysis} \bibinfo{volume}{46} (\bibinfo{year}{2008}) \bibinfo{pages}{2309--2345}.

\end{thebibliography}

\end{document}